%% file: acl_latex.tex
\documentclass[11pt]{article}

\usepackage[final]{acl}

\usepackage{times}
\usepackage{latexsym}

\usepackage[T1]{fontenc}
\usepackage[utf8]{inputenc}

\usepackage[whole]{bxcjkjatype}

\usepackage{microtype}
\usepackage{inconsolata}
\usepackage{graphicx}

\usepackage{booktabs}
\usepackage{multirow}
\usepackage[table]{xcolor}
\usepackage{tabularx}
\usepackage{array}
\usepackage{amsmath}
\usepackage{makecell}
\usepackage[most]{tcolorbox}
\usepackage{multicol}
\usepackage{longtable}
\usepackage{arydshln}
\usepackage{threeparttable}
\usepackage{placeins}

\usepackage{graphicx}

\usepackage{booktabs}
\usepackage{multirow}
\usepackage{graphicx}
\usepackage[table]{xcolor}
\usepackage{tabularx}
\usepackage{array}
\usepackage{amsmath}
\usepackage{algorithm}
\usepackage{algpseudocode}
\usepackage{fontawesome5}

\newcommand{\zh}[1]{{\CJKfamily{gbsn}#1}}

\definecolor{rebuttalrevisioncolor}{rgb}{0,0,0}
\newcommand{\rebuttalrevision}[1]{{\hypersetup{linkcolor=rebuttalrevisioncolor}\textcolor{rebuttalrevisioncolor}{#1}}}
\newenvironment{rebuttalrevisionblock}{\begingroup\color{rebuttalrevisioncolor}\hypersetup{linkcolor=rebuttalrevisioncolor}}{\par\endgroup}

\title{CounselTraj: Synthesizing Process-Aware Psychological Counseling Dialogues via Dialogue-Act Trajectory Planning}

\author{Zhiyang Qi \\
  The University of Tokyo \\
  Tokyo, Japan \\
  \texttt{zhiyangqi@g.ecc.u-tokyo.ac.jp} \\\And
  Michimasa Inaba\thanks{Corresponding author.} \\
  The University of Electro-Communications \\
  Tokyo, Japan \\
  \texttt{m-inaba@uec.ac.jp} \\
  \AND
  \normalfont\normalsize
  \href{https://github.com/UEC-InabaLab/CounselTraj}{\faGithub\enspace\nolinkurl{https://github.com/UEC-InabaLab/CounselTraj}}}

\begin{document}
\maketitle
\begin{abstract}

The growing demand for psychological counseling has motivated the development of LLM-based systems that support human counselors, but training such systems requires large-scale, high-quality dialogue data. Authentic counseling transcripts are difficult to collect and share because of privacy concerns, making synthetic dialogues an important alternative. However, existing synthetic datasets, despite covering diverse scenarios and client profiles, often consist of brief exchanges and fail to capture session-level counseling processes. We propose \textbf{CounselTraj}, a dialogue-act trajectory planning framework for expanding brief synthetic dialogues into process-aware, session-level psychological counseling dialogues. CounselTraj learns dialogue-act patterns from role-play counseling dialogues produced by human counselors, plans coordinated trajectories for clients and counselors, and uses them as flexible guidance for long-form dialogue generation. Experiments across three Chinese synthetic counseling datasets show that CounselTraj produces more complete and coherent sessions with higher-quality counselor interventions than both direct expansion and expansion with randomly sampled trajectories. Based on CounselTraj, we construct \textbf{CounselTrajD}, a dataset containing over 10K process-guided Chinese psychological counseling dialogues, and use it to train \textbf{CounselTrajLM} for counseling response generation. Evaluations at both the response and session levels show that CounselTrajLM outperforms publicly available Chinese counseling models in response quality, therapeutic understanding and intervention, and counseling-process coherence and progress. Our code and model are publicly available at \url{https://github.com/UEC-InabaLab/CounselTraj}.

\end{abstract}

\begin{figure}[t!]
  \centering
  \includegraphics[width=\linewidth]{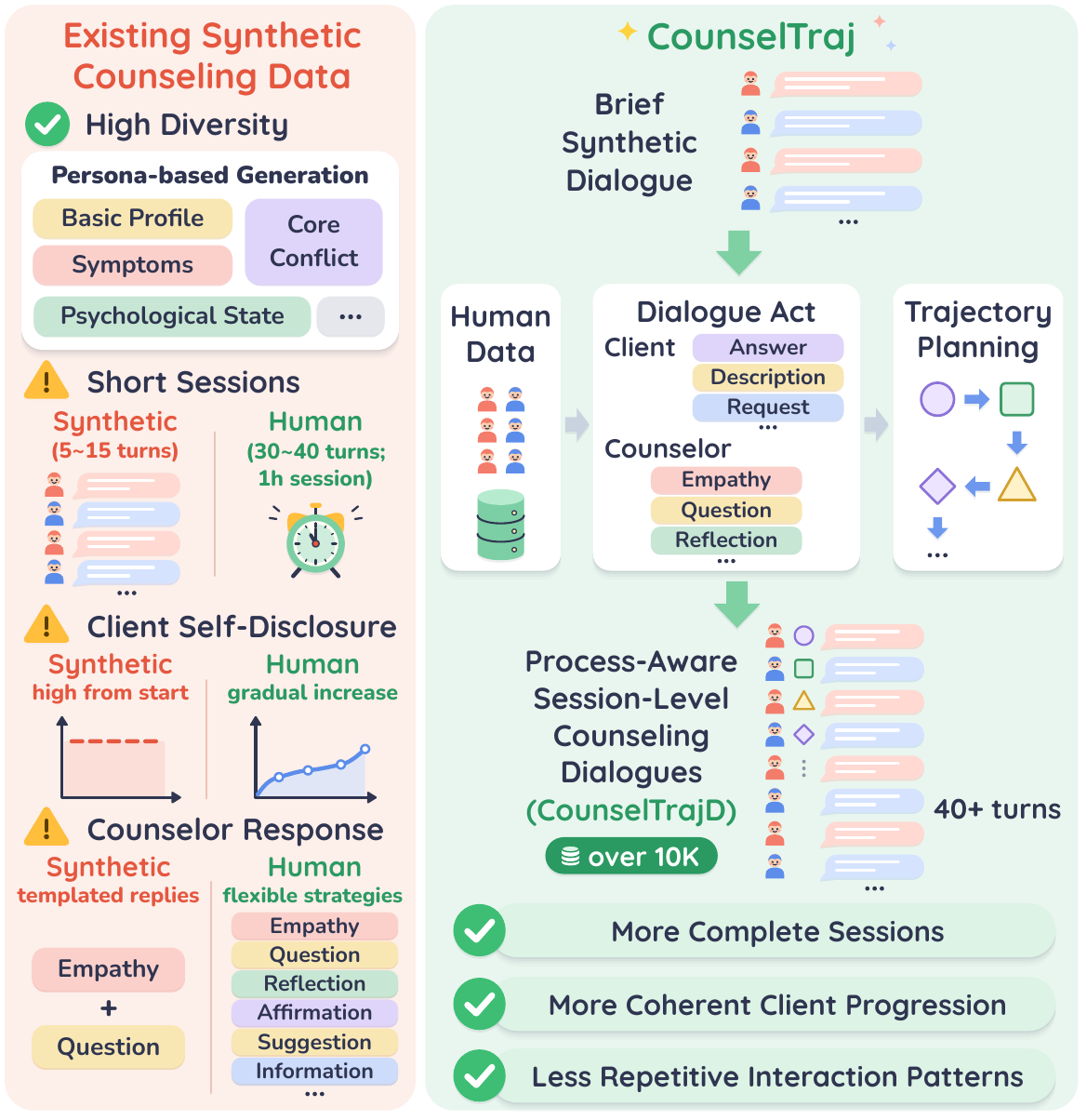}
  \caption{Motivation of CounselTraj. Existing synthetic psychological counseling dialogues are diverse but often brief and lack session-level process structure. CounselTraj expands them into process-aware session-level dialogues through dialogue-act trajectory planning.}
  \label{fig:fig_intro}
\end{figure}

\section{Introduction}

\begin{table*}[t]
\centering
\small
\setlength{\tabcolsep}{5.5pt}
\renewcommand{\arraystretch}{1.18}
\resizebox{\textwidth}{!}{
\begin{tabular}{@{}llccccccc@{}}
\toprule
\multirow{2}{*}{\textbf{Type}} 
& \multirow{2}{*}{\textbf{Dataset}} 
& \multirow{2}{*}{\textbf{Language}} 
& \multirow{2}{*}{\textbf{Dialogues}} 
& \multirow{2}{*}{\textbf{Avg. Utterances}} 
& \multicolumn{4}{c}{\textbf{Turns after merging}} \\
\cmidrule(lr){6-9}
& & & & & \textbf{Avg.} & \textbf{Min} & \textbf{Max} & \textbf{Median} \\
\midrule

\multirow{4}{*}{Synthetic}
& SmileChat~\cite{qiu-etal-2024-smile} 
& ZH & 55,165 & 11.39 & \cellcolor{gray!10}5.61  & 0  & 19  & 6.0  \\
& CPsyCounD~\cite{zhang-etal-2024-cpsycoun} 
& ZH & 3,084  & 16.14 & \cellcolor{gray!10}7.85  & 2  & 30  & 8.0  \\
& CACTUS~\cite{lee-etal-2024-cactus} 
& EN & 31,577 & 30.53 & \cellcolor{gray!10}14.97 & 10 & 17  & 15.0 \\
& PsyDTCorpus~\cite{xie-etal-2025-psydt} 
& ZH & 4,760  & 36.16 & \cellcolor{gray!10}18.08 & 9  & 35  & 18.0 \\

\addlinespace[2pt]
\arrayrulecolor{gray!45}
\cdashline{1-9}
\arrayrulecolor{black}
\addlinespace[2pt]

\multirow{3}{*}{Human}
& AnnoMI~\cite{AnnoMI} 
& EN & 133    & 72.92 & \cellcolor{gray!10}36.05 & 3  & 299 & 23.0 \\
& KokoroChat~\cite{qi-etal-2025-kokorochat} 
& JA & 6,589  & 91.20 & \cellcolor{gray!10}35.51 & 8  & 129 & 34.0 \\
& PsyDial-D4~\cite{qiu-lan-2025-psydial} 
& ZH & 2,382  & 75.59 & \cellcolor{gray!10}37.80 & 9  & 120 & 35.0 \\

\bottomrule
\end{tabular}
}
\caption{
Statistics of synthetic and human-collected counseling dialogue datasets. 
Turns are counted after merging adjacent utterances from the same speaker and excluding incomplete client--counselor exchanges.
(For SmileChat, some dialogues contain only a single utterance, e.g., dialogue 20493 and 41569; these cases are counted as 0 turns.)
}
\label{tab:dataset_statistics}
\end{table*}

Mental health problems are an increasing social concern, while professional counseling resources remain insufficient~\cite{who2022worldmentalhealth}. Online counseling, with its accessibility and scalability, has emerged as an important means of expanding psychological support~\cite{casu2024ai}. In this context, LLM-based dialogue systems may assist human counselors and improve the efficiency and coverage of online counseling services~\cite{hua2025large}. However, developing such systems requires large-scale, high-quality counseling dialogue data. Authentic counseling transcripts are difficult to collect and share because they contain highly sensitive personal information. Role-play counseling dialogues involving human counselors provide a privacy-preserving alternative, but they remain costly to create and limited in scale~\cite{qi-etal-2025-kokorochat}.

LLM-based dialogue synthesis offers a scalable way to alleviate this data scarcity~\cite{qiu-etal-2024-smile,zhang-etal-2024-cpsycoun,xie-etal-2025-psydt}. Recent studies have improved the content diversity of synthetic counseling data by specifying client information such as personality traits, symptoms, psychological states, and core conflicts~\cite{lee-etal-2024-cactus,wu2026ecas}. However, diversity in topics and client profiles does not ensure adequate session-level process structure. Psychological counseling unfolds through evolving client--counselor interactions, including relationship building, problem exploration, and adaptive interventions~\cite{fu2024hmmcounselling,qi2025emostageframeworkaccurateempathetic}. Therefore, useful synthetic data should represent not only what is discussed, but also how the interaction develops over a counseling session.

Current synthetic counseling dialogues still inadequately model such processes. Figure~\ref{fig:fig_intro} shows three main limitations. First, synthetic dialogues are considerably shorter than human-counselor role-play sessions (Table~\ref{tab:dataset_statistics}), making complete counseling processes difficult to represent. Second, synthetic clients often present extensive background information and concerns at dialogue onset, producing abrupt rather than progressively developed interactions. Third, counselor responses often repeat fixed patterns, such as an empathy-then-question structure\footnote{For example, our simple analysis shows that 43.9\% of counselor utterances in CPsyCounD~\cite{zhang-etal-2024-cpsycoun} and 67.4\% in PsyDTCorpus~\cite{xie-etal-2025-psydt} end with a question mark. See Tables~\ref{tab:appendix_template_response_psydt} and~\ref{tab:appendix_template_response_cpsy} in Appendix~\ref{sec:appendix_template_responses} for examples.}. By contrast, human counselors use a broader range of dialogue acts as interactions evolve. Consequently, models trained on existing synthetic data may learn compressed client progression and rigid counselor response patterns.

To introduce explicit session-level process structure, we propose \textbf{CounselTraj}, a dialogue-act trajectory planning framework for expanding brief synthetic dialogues into process-aware psychological counseling dialogues. Rather than directly prompting an LLM to lengthen a brief dialogue, CounselTraj represents the expected interaction as a coordinated trajectory of client and counselor dialogue acts. It learns dialogue-act patterns from role-play counseling dialogues involving human counselors, generates a context-dependent trajectory for each input dialogue, and uses the trajectory as flexible guidance for long-form generation. 

Experiments on SmileChat, CPsyCounD, and PsyDTCorpus compare CounselTraj with direct expansion and expansion using randomly sampled trajectories. CounselTraj consistently produces higher-quality session-level dialogues, particularly in session completeness, counselor intervention quality, and client participation and collaboration.
Using CounselTraj, we construct \textbf{CounselTrajD}, a Chinese psychological counseling dataset with over 10K process-guided dialogues, and train \textbf{CounselTrajLM} for counseling response generation. Response- and session-level evaluations show that CounselTrajLM outperforms publicly available Chinese counseling models in response quality, therapeutic understanding and intervention, and counseling-process coherence and progress.

Our contributions are fivefold.
(1) We systematically compare existing synthetic counseling dialogues with role-play dialogues involving human counselors, identifying gaps in dialogue length and session-level interaction patterns.
(2) We propose CounselTraj, a dialogue-act trajectory planning framework that introduces explicit process guidance into the expansion of brief synthetic counseling dialogues.
(3) Across three synthetic datasets, we demonstrate the advantages of context-dependent trajectory planning over both direct expansion and randomly sampled trajectories.
(4) We construct CounselTrajD, a Chinese psychological counseling dataset containing over 10K process-guided dialogues expanded from diverse synthetic seeds.
(5) We develop CounselTrajLM and demonstrate the downstream value of CounselTrajD through both next-response generation and multi-turn counseling interaction evaluations.

\section{Related Work}

\subsection{AI for Mental Health Support}

Research on AI for mental health support has progressed from empathetic dialogue modeling to counseling-oriented LLMs and agents~\cite{hu2026patternrecognizers,hu2026agentmental,kim-etal-2025-share-story,qiu-etal-2025-emoagent,qiu-etal-2025-deepwell,qiu-etal-2026-client}. Early work studied empathetic response generation and strategy-guided emotional support through resources such as EmpatheticDialogues and ESConv~\cite{rashkin2019towards,sharma-etal-2020-computational,sharma2021facilitating,liu2021towards}. More recent systems include SoulChat, MeChat, and embodied counseling agents~\cite{chen-etal-2023-soulchat,qiu-etal-2024-smile,wu2026ecas}, alongside studies of minimal counselor responses~\cite{qi-2026-less}, safety, trust, and human--AI collaboration~\cite{sharma2023human,song2025typing,wester2024chatbot,poulsen2025codesigning}. These studies demonstrate the potential of AI to support human counselors while highlighting the need for high-quality counseling data.

\subsection{Psychological Counseling Dialogue Data and Construction Methods}

Psychological counseling dialogues are primarily constructed through human-authored collection or LLM-based synthesis. Human-authored resources include public counseling videos, reconstructed online counseling, and role-play involving human counselors~\cite{AnnoMI,qi-etal-2025-kokorochat,qiu-lan-2025-psydial,suzuki-etal-2026-multilingual}. They provide rich interaction patterns but remain limited in scale. LLM-generated datasets offer diverse client profiles and scenarios at lower cost~\cite{qiu-etal-2024-smile,zhang-etal-2024-cpsycoun,lee-etal-2024-cactus,xie-etal-2025-psydt,wu2026ecas}, but often consist of brief exchanges without explicit session-level process modeling. PsyCLIENT uses conversational trajectories extracted from existing counseling dialogues with expert-supervised behavior annotations to guide client simulation~\cite{qiu-etal-2026-client}. CounselTraj instead learns dialogue-act patterns from counselor-involved role-play data and automatically generates new, input-conditioned trajectories for both clients and counselors to guide session-level dialogue expansion.

\section{Synthetic vs. Human-Collected Data}
\label{sec:analysis}

To clarify the gaps between synthetic and human-collected counseling data, we analyze their dialogue length and client expression patterns.

\subsection{Dialogue Length Analysis}

\begin{figure}[t!]
  \centering
  \includegraphics[width=\linewidth]{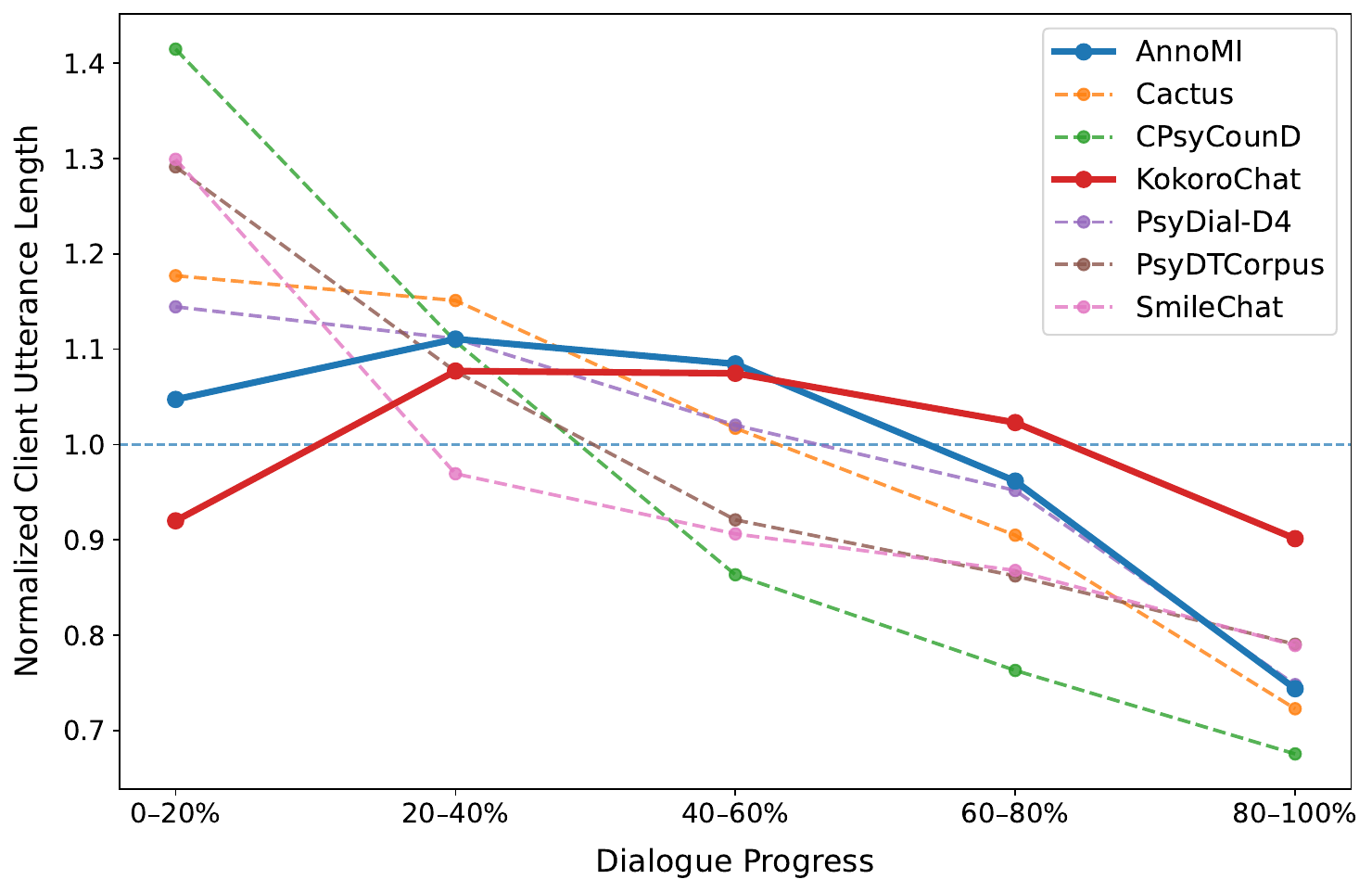}
  \caption{Comparison of stage-wise changes in normalized client utterance length across datasets. Solid lines indicate datasets with human client utterances.}
  \label{fig:client_length}
\end{figure}

As shown in Table~\ref{tab:dataset_statistics}, human-collected datasets generally contain longer counseling sessions. KokoroChat and PsyDial-D4 were collected as one-hour online text-based counseling sessions, with average turns of 35.51 and 37.80 after merging adjacent utterances from the same speaker. AnnoMI, which consists of transcripts of face-to-face counseling videos, also has an average of 36.05 turns. In contrast, synthetic counseling datasets contain substantially fewer turns on average. This gap may prevent current synthetic data from sufficiently capturing processes that unfold in real counseling sessions, such as stage progression, client development, and counselor interventions.

\subsection{Client Expressions and Dialogue Acts}

\paragraph{Stage-wise utterance length.}

To examine stage-wise changes in client expression, we analyze client utterance length across dialogue progress. Following prior work~\cite{althoff2016large}, we divide each dialogue into five equal stages and compute the normalized client utterance length in each stage relative to the dialogue-level average. As shown in Figure~\ref{fig:client_length}, KokoroChat shows a typical increase-then-decrease pattern: clients begin with shorter utterances, provide more detailed descriptions and self-disclosures as the session develops, and become shorter again in later stages as counselors provide more summaries, suggestions, or coping strategies. AnnoMI shows a similar trend. In contrast, synthetic datasets generally show a gradual decrease, suggesting that clients tend to provide substantial information from the beginning and lack the gradual development of expression observed in real counseling. PsyDial-D4 shows a similar trend to synthetic data, possibly because its client utterances were generated by LLMs for privacy.

\paragraph{Dialogue-act analysis.}

To further compare client expression patterns, we annotate client utterances using the dialogue-act tags introduced later in Section~\ref{sec:da_tag} and summarized in Table~\ref{tab:dialogue_act_tags}. We sample 500 dialogues per dataset when available and use all dialogues for AnnoMI; annotations are produced with \texttt{gemini-3.1-flash-lite}.

\begin{figure}[t!]
  \centering
  \includegraphics[width=\linewidth]{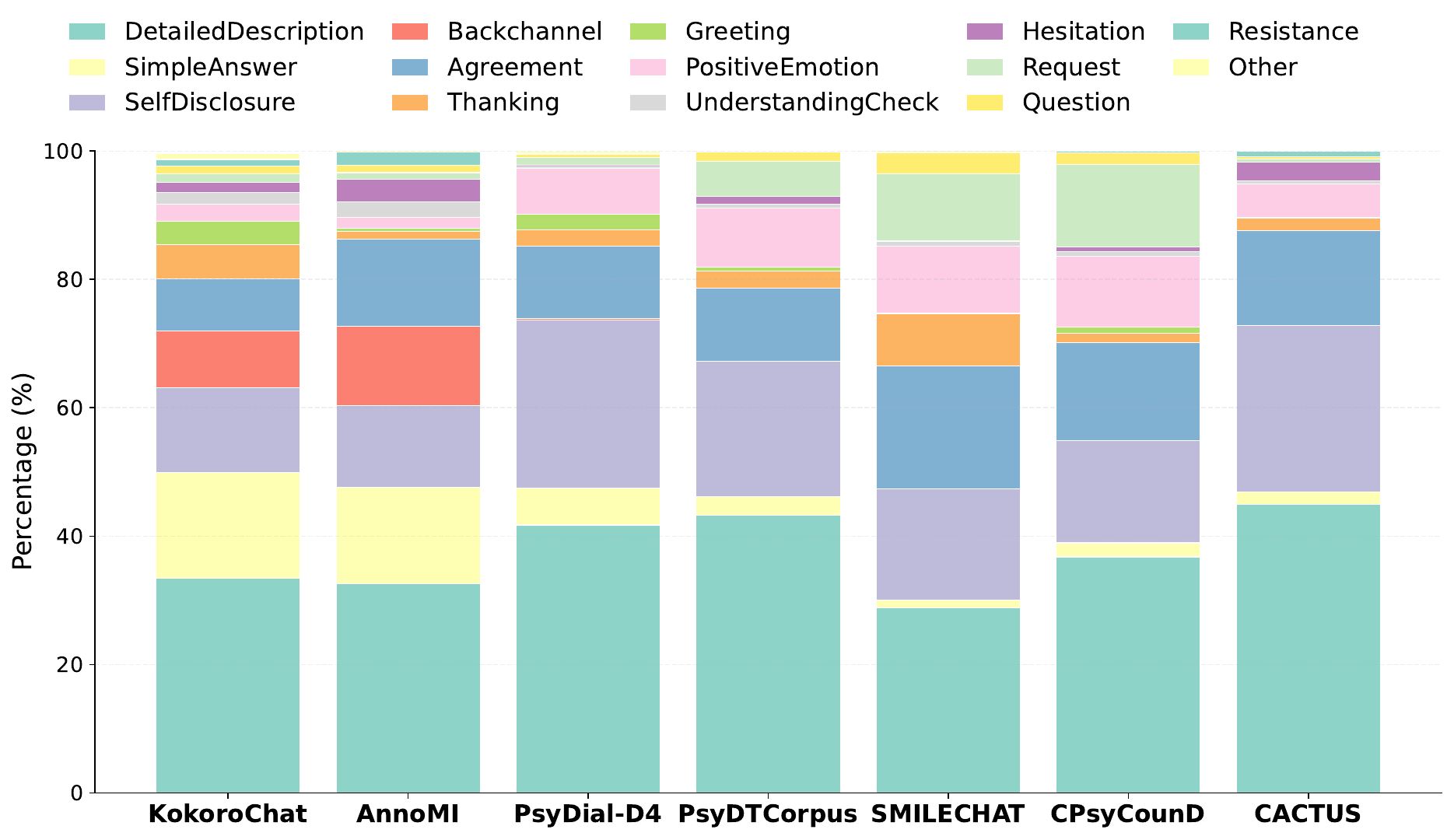}
  \caption{Client-side dialogue-act distribution.}
  \label{fig:tag_distribution}
\end{figure}

Figure~\ref{fig:tag_distribution} shows the distribution of client-side dialogue acts. KokoroChat and AnnoMI contain notable proportions of \textit{SimpleAnswer} and \textit{Backchannel}, suggesting that clients in human counseling often maintain interaction through short answers and backchannels. In contrast, most synthetic datasets contain fewer such short responses but more \textit{DetailedDescription} and \textit{SelfDisclosure}, indicating a tendency toward information-dense answers and direct disclosure. They also show higher proportions of \textit{PositiveEmotion} and \textit{Request}, suggesting overly cooperative clients who express improvement or actively seek advice. PsyDial-D4 also shows a trend similar to synthetic datasets. Overall, existing synthetic data tend to lack short responses, overemphasize detailed descriptions, and portray clients as overly cooperative. Detailed counts and stage-wise trends are provided in the Appendix~\ref{sec:appendix_data_analysis}.

\begin{table}[t]
\centering
\small
\setlength{\tabcolsep}{3pt}
\renewcommand{\arraystretch}{1.12}
\begin{tabular}{p{0.17\linewidth}p{0.76\linewidth}}
\toprule
\textbf{Role} & \textbf{Dialogue-act tags} \\
\midrule
\multirow{5}{*}{Client}
& DetailedDescription, UnderstandingCheck, \\
& SimpleAnswer, Agreement, Request, \\
& SelfDisclosure, PositiveEmotion, Backchannel, \\
& Question, Hesitation, Resistance, \\
& Greeting, Thanking, Other \\
\midrule
\multirow{3}{*}{Counselor}
& OpenQuestion, ClosedQuestion, Paraphrase, \\
& Reflection, Affirm, Suggest, Inform, \\
& Backchannel, Greeting, Thanking, Other \\
\bottomrule
\end{tabular}
\caption{Tags for client and counselor utterances.}
\label{tab:dialogue_act_tags}
\end{table}

\section{Proposed Framework}

\begin{figure*}[t!]
  \centering
  \includegraphics[width=0.95\linewidth]{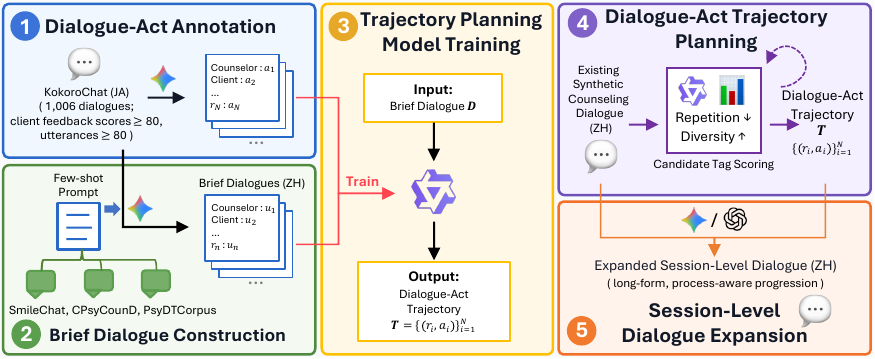}
  \caption{Overview of CounselTraj. High-quality Japanese counseling role-play dialogues from KokoroChat are annotated with dialogue acts and rewritten as brief Chinese dialogues through few-shot prompting. A trajectory planner is trained on the paired brief dialogues and annotated trajectories. For each existing brief synthetic dialogue, the planned trajectory flexibly guides an LLM to generate a long-form, process-aware counseling conversation.}
  \label{fig:framework}
\end{figure*}


Based on the analysis in Section~\ref{sec:analysis}, we use KokoroChat to learn session-level process patterns. KokoroChat consists of one-hour online text-based role-play sessions in which trained counselors play both roles. We propose \textbf{CounselTraj}, which expands brief synthetic counseling dialogues through dialogue-act trajectory planning.
As shown in Figure~\ref{fig:framework}, we select high-quality KokoroChat dialogues and (1) annotate dialogue acts, (2) construct corresponding brief dialogues, (3) train a trajectory planning model, (4) plan trajectories for existing synthetic dialogues, and (5) use them to guide process-aware, session-level dialogue expansion.


\subsection{Dialogue-Act Annotation}
\label{sec:da_tag}

To model client and counselor behaviors in counselor-involved role-play sessions, we define separate dialogue-act tag sets for the two roles based on prior studies~\cite{inaba-2026-beyond}. The tags are listed in Table~\ref{tab:dialogue_act_tags}, with detailed definitions provided in Appendix~\ref{sec:appendix_da_tag_definitions_counselor_prompt} (Table~\ref{tab:dialogue_act_tag_definitions}).

KokoroChat provides session feedback scores from client-role participants. We select 1,006 dialogues with feedback scores of at least 80 and at least 80 raw utterances before merging adjacent same-speaker turns. Counselor utterances are annotated using \texttt{gemini-2.5-flash}, while client utterances are annotated using \texttt{gemini-3.1-flash-lite}.\footnote{The two annotation tasks were conducted at different stages using the Gemini version available at each stage.} The annotation prompts are provided in Appendix~\ref{sec:appendix_method_details}.
We assess annotation reliability through pairwise Cohen's $\kappa$~\citep{cohen1960coefficient}. For counselor tags, two crowd workers annotated 318 utterances from five dialogues, yielding $\kappa=0.674$ between workers and $\kappa=0.647$ and $0.644$ between Gemini and each worker. For client tags, they annotated 389 utterances from five dialogues, yielding $\kappa=0.582$ between workers and $\kappa=0.547$ and $0.544$ between Gemini and each worker. These results indicate moderate agreement and support the use of automatic annotations for subsequent analysis and modeling.

\subsection{Brief Dialogue Construction}

To train the dialogue-act trajectory planning model, we construct pairs $(D,T)$, where $D$ is a brief Chinese dialogue and $T$ is the dialogue-act trajectory annotated from the corresponding KokoroChat session. We use \texttt{gemini-3.1-flash-lite} with few-shot prompting to translate and compress long Japanese-language KokoroChat dialogues into brief Chinese dialogues. This aligns the model inputs with the language and style of existing Chinese synthetic counseling datasets. Few-shot examples are selected from SmileChat, CPsyCounD, and PsyDTCorpus, and the full prompt is provided in Appendix~\ref{sec:appendix_brief_dialogue_construction_prompt}.

To match the input distribution encountered during expansion, client utterances are condensed into the compact, information-dense style of existing synthetic dialogues, while counselor utterances are merged or shortened where appropriate. The prompt instructs the model to preserve the original concern and meaning without introducing new events, characters, or background information. Each resulting brief dialogue $D$ is paired with the original session's trajectory $T$ for model training.
\rebuttalrevision{To assess the structural plausibility of the trajectory sources and constructed briefs, we report additional dialogue-act distribution analyses in Appendix~\ref{sec:appendix_source_brief_validation}.}

\subsection{Trajectory Planning Model Training}

After constructing brief dialogues, we train a dialogue-act trajectory planning model. Given a Chinese brief dialogue \(D=\{(r_j,u_j)\}_{j=1}^{n}\), where \(r_j\) and \(u_j\) denote the speaker role and utterance, respectively, the model predicts the complete session-level dialogue-act trajectory \(T=\{(r_i,a_i)\}_{i=1}^{N}\) annotated from the original KokoroChat dialogue. Here, \(a_i\) denotes the dialogue-act tag. Formally, the model learns the mapping \(f_{\theta}: D \rightarrow T\), enabling it to plan a session-level client--counselor interaction process conditioned on the brief dialogue.

\subsection{Synthetic Dialogue Trajectory Planning}
\label{sec:path_planning_synthetic_dialogues}

Given a brief synthetic dialogue \(D\), the trained planner incrementally generates a session-level trajectory \(T=\{(r_i,a_i)\}_{i=1}^{N}\), where \(r_i\), \(a_i\), and \(N\) denote the speaker role, dialogue act, and final trajectory length, respectively. At step \(i\), each candidate act is scored based on \(D\) and the preceding trajectory \(H_{<i}=\{(r_j,a_j)\}_{j=1}^{i-1}\) using \(S(a_i)=S_{\mathrm{LM}}(a_i)+S_{\mathrm{guide}}(a_i)\). Here, \(S_{\mathrm{LM}}\) is the planner's conditional log-likelihood, while \(S_{\mathrm{guide}}\) combines stage-wise, global, and transition statistics estimated from annotated KokoroChat dialogues with content and phase compatibility, repetition penalties, and diversity bonuses. Incompatible acts are excluded. We select the highest-scoring act when its score margin is sufficiently large; otherwise, we sample from the top-ranked candidates.

Client and counselor roles alternate throughout generation, and the trajectory length is dynamically determined from the input dialogue. The resulting trajectory guides session-level dialogue expansion. Further details are provided in Appendix~\ref{sec:appendix_path_planning_synthetic_dialogues}.

\subsection{Session-Level Dialogue Expansion}

After obtaining a session-level dialogue-act trajectory for each brief synthetic dialogue, we further generate a long-form counseling dialogue using an LLM. The model takes both the brief dialogue and the planned dialogue-act trajectory as input and expands the dialogue while preserving the original topic, emotional context, and problem background, while allowing appropriate details to be added. The trajectory specifies the speaker role and provides a flexible guide to the dialogue function of each utterance rather than a rigid template; the model may adapt the planned dialogue acts when needed to improve coherence and naturalness. The prompt is provided in Appendix~\ref{sec:appendix_session_level_dialogue_expansion_prompt}.

\section{Dialogue Generation Evaluation}
\label{sec:dialogue_generation_evaluation}

To assess whether dialogue-act trajectory planning can improve the quality
of brief synthetic counseling dialogues across different data sources, we
evaluate CounselTraj on three synthetic counseling datasets. We compare it
with direct expansion, expansion guided by an unrelated trajectory,
\rebuttalrevision{and zero-/few-shot prompted trajectory planning,} and further analyze the plausibility of the planned dialogue-act trajectories
and the extent to which they are realized in the expanded dialogues.

\subsection{Experimental Setup}

\begin{rebuttalrevisionblock}
\paragraph{Data and compared variants.}
We randomly sample 200 seed dialogues each from SmileChat~\cite{qiu-etal-2024-smile}, CPsyCounD~\cite{zhang-etal-2024-cpsycoun}, and PsyDTCorpus~\cite{xie-etal-2025-psydt}, yielding 600 seeds. We compare the original dialogue with five expanded variants: \textit{Direct}, without trajectory guidance; \textit{Random Traj.}, guided by a trajectory for another seed; \textit{Zero-shot} and \textit{Few-shot}, guided by trajectories generated through prompting; and \textit{CounselTraj}, guided by the trained planner.

\paragraph{Implementation.}
The trained planner is built on Qwen3-8B\footnote{\url{https://huggingface.co/Qwen/Qwen3-8B}}, and all expansion variants use \texttt{gemini-3.1-flash-lite}. For Zero-shot and Few-shot, the same Gemini model generates trajectories from the input brief; Few-shot additionally includes three high-scoring KokoroChat trajectories as demonstrations. Baseline prompts and expansion settings are provided in Appendix~\ref{sec:appendix_experimental_details}.

\subsection{Comparative Quality Evaluation}

\paragraph{LLM-as-a-Judge protocol.}
Following prior work~\cite{zhang-etal-2024-cpsycoun,xie-etal-2025-psydt,qi-etal-2025-kokorochat,qiu-lan-2025-psydial}, we use \texttt{gpt-5.6-luna} and \texttt{gemini-3.8-flash} as judges. For each matched seed, all six variants are presented together in randomized order with method identities concealed. Each judge provides comparative rationales and scores from 0 to 5, allowing 0.5-point increments, for \textit{Session Completeness} (Comp.), \textit{Counselor Intervention Quality} (Int.), and \textit{Client Engagement and Collaboration} (Eng.). We average the two judges' scores with equal weight before aggregating across seeds. The full prompt and metric definitions are provided in Appendix~\ref{sec:appendix_dialogue_quality_evaluation_prompt}.

\begin{table}[t]
\centering
\color{rebuttalrevisioncolor}
\arrayrulecolor{rebuttalrevisioncolor}
\footnotesize
\setlength{\tabcolsep}{3.0pt}
\renewcommand{\arraystretch}{1.02}
\captionsetup{font+={color=rebuttalrevisioncolor}}
\caption{Six-way dialogue quality comparison, averaged equally across \texttt{gpt-5.6-luna} and \texttt{gemini-3.8-flash}. Scores range from 0 to 5 (higher is better). Best results are bold; second-best results are underlined.}
\label{tab:dialogue_generation_quality}
\begin{tabular}{@{}llcccc@{}}
\toprule
\textbf{Source} & \textbf{Variant}
& \textbf{Comp.} & \textbf{Int.} & \textbf{Eng.} & \textbf{Avg.} \\
\midrule
\multirow{6}{*}{SmileChat} & Original & 2.07 & 1.78 & 1.95 & 1.93 \\
 & Direct & 2.49 & 2.42 & 2.55 & 2.49 \\
 & Random Traj. & \underline{3.85} & 3.32 & 3.50 & 3.56 \\
 & Zero-shot & 3.43 & 3.13 & 3.30 & 3.29 \\
 & Few-shot & 3.76 & \underline{3.59} & \underline{3.79} & \underline{3.71} \\
 & \textbf{CounselTraj} & \textbf{4.30} & \textbf{4.27} & \textbf{4.27} & \textbf{4.28} \\
\midrule
\multirow{6}{*}{CPsyCounD} & Original & 2.06 & 1.75 & 1.88 & 1.90 \\
 & Direct & 2.18 & 2.26 & 2.39 & 2.28 \\
 & Random Traj. & 3.76 & 3.15 & 3.34 & 3.42 \\
 & Zero-shot & 3.65 & 3.48 & 3.62 & 3.58 \\
 & Few-shot & \underline{3.95} & \underline{3.77} & \underline{3.95} & \underline{3.89} \\
 & \textbf{CounselTraj} & \textbf{4.29} & \textbf{4.25} & \textbf{4.26} & \textbf{4.26} \\
\midrule
\multirow{6}{*}{PsyDTCorpus} & Original & 3.47 & 3.08 & 3.16 & 3.24 \\
 & Direct & 2.46 & 2.53 & 2.59 & 2.53 \\
 & Random Traj. & 3.55 & 3.00 & 3.20 & 3.25 \\
 & Zero-shot & 4.05 & 3.87 & 3.98 & 3.97 \\
 & Few-shot & \underline{4.08} & \underline{3.90} & \underline{3.99} & \underline{3.99} \\
 & \textbf{CounselTraj} & \textbf{4.22} & \textbf{4.17} & \textbf{4.14} & \textbf{4.18} \\
\bottomrule
\end{tabular}
\arrayrulecolor{black}
\end{table}

\paragraph{Results.}
Table~\ref{tab:dialogue_generation_quality} shows that CounselTraj achieves the highest scores on all three dimensions across all source datasets. The overall means are 2.36 for Original, 2.43 for Direct, 3.41 for Random Traj., 3.61 for Zero-shot, 3.87 for Few-shot, and 4.24 for CounselTraj. Few-shot outperforms Zero-shot on every source, although Zero-shot falls below Random Traj. on SmileChat. These results favor CounselTraj among the tested configurations.
\end{rebuttalrevisionblock}

\subsection{Trajectory Analysis}

\paragraph{Distributional plausibility and diversity.}
We compare the 600 planned trajectories with the dialogue-act trajectories annotated from KokoroChat.
The Jensen--Shannon divergence between their aggregate dialogue-act distributions is 0.026, while their label entropies are 4.09 and 3.88, respectively.
The low divergence indicates that the planned trajectories preserve a dialogue-act composition similar to KokoroChat, while their slightly higher entropy suggests that the planner maintains broad dialogue-act diversity rather than concentrating on a few frequent acts.
Moreover, all 600 planned trajectories are unique, providing further evidence against trajectory-level repetition.

\paragraph{Trajectory realization.}
Because the planned trajectories serve as soft process-level guidance rather than rigid utterance-level templates, we examine whether their sequential structure is reflected in the generated dialogues.
We re-annotate all 600 CounselTraj dialogues using \texttt{gemini-3.1-flash-lite} and compute the order-aware retention score $\mathrm{LCS}(T,\hat{T})/|\hat{T}|$, where $T$ and $\hat{T}$ denote the planned and re-annotated trajectories, respectively.
The resulting retention rate is 57.7\%.
This is a conservative measure that counts only exact dialogue-act matches in the planned order; merged, omitted, inserted, or relabeled acts are treated as mismatches.
Thus, more than half of the realized acts remain aligned with the planned sequence despite allowing flexible dialogue generation, supporting the interpretation of trajectories as soft structural guidance rather than strict templates.


\subsection{Human Evaluation}

\begin{table}[t]
\centering
\small
\setlength{\tabcolsep}{6pt}
\renewcommand{\arraystretch}{1.05}
\begin{tabular}{lccc}
\toprule
\textbf{Comparison} & \textbf{Win} & \textbf{Tie} & \textbf{Lose} \\
\midrule
Ours vs. SmileChat    & 75.3\% & 0.0\%  & 24.7\% \\
Ours vs. CPsyCounD    & 76.0\% & 0.0\%  & 24.0\% \\
Ours vs. PsyDTCorpus  & 71.3\% & 0.0\%  & 28.7\% \\
\midrule
Ours vs. PsyDial-D101 & 28.0\% & 26.0\% & 46.0\% \\
\bottomrule
\end{tabular}
\caption{Human evaluation results. The first three rows compare CounselTraj-expanded dialogues with their original versions, while PsyDial-D101 represents professional-book counseling dialogues.}
\label{tab:human_dialogue_quality}
\end{table}

We further conduct human evaluation to compare CounselTraj-expanded dialogues with both their original dialogues and professional counseling dialogues.
For each source dataset, we randomly sample 50 pairs of original and expanded dialogues and ask three undergraduate psychology students to compare their quality.
We additionally compare 50 length-matched SmileChat-based CounselTraj dialogues with professional-book dialogues from PsyDial-D101~\cite{qiu-lan-2025-psydial}, where the evaluators may also judge the two dialogues as tied.
Dialogue order is randomized and the evaluators are blinded to their sources.
As shown in Table~\ref{tab:human_dialogue_quality}, CounselTraj-expanded dialogues are preferred over their original versions in more than 70\% of judgments across all three source datasets.
In comparison with PsyDial-D101, however, CounselTraj wins 28.0\%, ties 26.0\%, and loses 46.0\% of the judgments.
These results show that CounselTraj improves the original synthetic dialogues but still falls short of professional-book dialogues.

\section{CounselTrajD Construction}
\label{sec:counseltrajd}

Having demonstrated that CounselTraj improves synthetic dialogue quality across multiple source datasets, we use it to construct \textbf{CounselTrajD}, a large-scale Chinese session-level psychological counseling dialogue dataset from brief synthetic dialogues in SmileChat, CPsyCounD, and PsyDTCorpus.
We prioritize topic balance during sampling while retaining all available dialogues for topics with limited data.
For each selected seed, the Qwen3-8B-based trajectory planner generates a content-conditioned dialogue-act trajectory, which is then used by \texttt{gpt-5.5} to produce a session-level dialogue.
This process results in 10,964 dialogues, with the main statistics reported in Table~\ref{tab:counseltrajd_statistics}.

Based on CounselTrajD, we fine-tune Qwen3.5-9B\footnote{\url{https://huggingface.co/Qwen/Qwen3.5-9B}} for counselor response generation and refer to the resulting model as \textbf{CounselTrajLM}.
The training objective, data split, and hyperparameter settings are provided in Appendix~\ref{sec:appendix_response_generation_model_details}.

\begin{table}[t]
\centering
\small
\setlength{\tabcolsep}{3pt}
\renewcommand{\arraystretch}{1.05}
\begin{tabular}{lrrr}
\toprule
\textbf{Statistic} & \textbf{Overall} & \textbf{Client} & \textbf{Counselor} \\
\midrule
Dialogues & 10,964 & -- & -- \\
Utts. & 958,791 & 473,910 & 484,881 \\
Utts./dialogue & 87.45 & 43.22 & 44.22 \\
Chars. & 42,197,155 & 18,321,830 & 23,875,325 \\
Chars./dialogue & 3,848.70 & 1,671.09 & 2,177.61 \\
Chars./utterance & 44.04 & 38.68 & 49.29 \\
\bottomrule
\end{tabular}
\caption{Statistics of CounselTrajD.}
\label{tab:counseltrajd_statistics}
\end{table}

\begin{table*}[t]
\centering

\vspace{-2pt}
\small
\setlength{\tabcolsep}{4pt}
\renewcommand{\arraystretch}{1.08}

\begin{tabular}{
@{}l
@{\hspace{3pt}\vrule width 0.35pt\hspace{3pt}}
*{8}{r}
>{\columncolor{gray!20}}r
@{\hspace{3pt}\vrule width 0.35pt\hspace{3pt}}
*{6}{r}
>{\columncolor{gray!20}}r
@{}
}
\toprule
& \multicolumn{9}{c}{\textbf{Response-Level (PsyDial-D101)}}
& \multicolumn{7}{c}{\textbf{Dialogue-Level (PsychEval)}} \\
\cmidrule(r{3pt}){2-10}
\cmidrule(l{3pt}){11-17}

\textbf{Model}
& \textbf{AL}
& \textbf{Emp.}
& \textbf{IC}
& \textbf{OQ}
& \textbf{SE}
& \textbf{CR}
& \textbf{GQ}
& \textbf{NJA}
& \textbf{Ovl.}
& \textbf{TA}
& \textbf{EU}
& \textbf{PU}
& \textbf{EI}
& \textbf{PC}
& \textbf{SB}
& \textbf{Ovl.} \\
\midrule

CPsyCounX
& 1.73 & 1.63 & 1.60 & 1.67 & 1.62 & 1.42 & 1.52 & 2.24 & 1.68
& 1.18 & 1.23 & 1.48 & 1.07 & 1.08 & 1.34 & 1.23 \\

SoulChat
& 2.28 & 2.06 & 1.85 & 1.94 & 1.92 & 1.47 & 1.88 & 3.35 & 2.09
& 1.75 & 1.54 & 1.86 & 1.33 & 1.41 & 2.38 & 1.71 \\

MindChat
& 2.56 & 2.11 & 2.15 & 2.24 & 2.22 & 1.61 & 2.22 & 3.62 & 2.34
& 1.91 & 1.77 & 2.07 & 1.48 & 1.46 & 2.29 & 1.83 \\

PsyChat
& 1.92 & 1.74 & 1.54 & 1.66 & 1.59 & 1.26 & 1.55 & 3.10 & 1.80
& 2.14 & 2.02 & 2.13 & 1.66 & 1.73 & 2.47 & 2.03 \\

MeChat
& 2.44 & 2.10 & 1.98 & 2.05 & 2.03 & 1.50 & 2.01 & 3.55 & 2.21
& 2.52 & 2.33 & 2.54 & 1.92 & 2.20 & 2.79 & 2.38 \\

Psyche-R1
& 2.92 & 2.43 & 2.46 & 2.46 & 2.61 & 1.89 & 2.58 & 3.79 & 2.64
& 2.51 & 2.39 & 2.75 & 2.02 & 2.09 & 2.86 & 2.44 \\

PsyDial-Pi4
& 3.10 & 2.73 & 2.39 & 2.20 & 2.38 & 1.57 & 2.20
& \underline{4.17} & 2.59
& 3.53 & 3.48 & 3.59 & 2.79 & 3.04 & 3.76 & 3.37 \\

Crispers
& 3.06 & 2.51 & 2.62 & 2.52 & 2.68 & 1.94 & 2.64 & 3.83 & 2.73
& 3.54 & 3.40 & 3.60 & 3.13 & 3.41 & 3.62 & 3.45 \\

SoulChat2.0
& \underline{3.69}
& \underline{3.24}
& \underline{3.11}
& \textbf{3.16}
& \textbf{3.36}
& \underline{2.08}
& \textbf{3.13}
& 4.11
& \underline{3.24}
& \underline{3.99}
& \underline{3.86}
& \underline{3.84}
& \underline{3.49}
& \underline{3.79}
& \underline{3.97}
& \underline{3.82} \\

\specialrule{0.7pt}{2pt}{2pt}

\textbf{CounselTrajLM}
& \textbf{4.13}
& \textbf{3.53}
& \textbf{3.25}
& \underline{2.55}
& \underline{3.19}
& \textbf{2.55}
& \underline{2.89}
& \textbf{4.59}
& \textbf{3.34}
& \textbf{4.74}
& \textbf{4.81}
& \textbf{4.75}
& \textbf{4.52}
& \textbf{4.63}
& \textbf{4.77}
& \textbf{4.70} \\
\bottomrule
\end{tabular}
\caption{Response- and dialogue-level results (1--5; higher is better), averaged across \texttt{GPT-5.4} and \texttt{gemini-3.5-flash}. Bold and underlined denote the best and second-best results; Ovl.\ denotes the average across all metrics in the corresponding evaluation.}
\label{tab:counseltrajlm_evaluation}
\end{table*}

\section{CounselTrajLM Evaluation}
\label{sec:counseltrajlm_evaluation}

To provide a comprehensive evaluation, we assess CounselTrajLM from two perspectives: response-level evaluation examines the quality of the next counselor response, while dialogue-level evaluation examines its performance throughout a complete counseling session.

\subsection{Evaluation Protocols}

\paragraph{Compared models.}
We compare CounselTrajLM with nine publicly available Chinese mental-health support and counseling models: MeChat~\cite{qiu-etal-2024-smile}, SoulChat~\cite{chen-etal-2023-soulchat}, MindChat-Qwen-7B-v2\footnote{\url{https://huggingface.co/X-D-Lab/MindChat-Qwen-7B-v2}}, PsyChat~\cite{qiu2024psychat}, CPsyCounX~\cite{zhang-etal-2024-cpsycoun}, SoulChat2.0-Qwen2-7B~\cite{xie-etal-2025-psydt}, PsyDial-Pi4~\cite{qiu-lan-2025-psydial}, Crispers-7B-v1~\cite{crisp}, and Psyche-R1~\cite{dai-etal-2026-psyche}.
We use the publicly released checkpoints with their corresponding inference formats, with further implementation details provided in Appendix~\ref{sec:appendix_response_generation_model_details}.

\paragraph{Response-level evaluation.}
To assess whether each model can generate an appropriate next counselor response from the preceding context, we conduct response-level evaluation.
We evaluate the models on the 1,278 response-generation examples in PsyDial-D101, collected from 101 professional-book counseling dialogues~\cite{qiu-lan-2025-psydial}.
Each example uses a dialogue history ending with a client utterance as input and the subsequent counselor response as the target.
Following the evaluation setting of Qiu and Lan~\cite{qiu-lan-2025-psydial}, we assess eight counseling skills: Active Listening (\textbf{AL}), Empathy (\textbf{Emp.}), Issue Clarification (\textbf{IC}), Open-ended Questioning (\textbf{OQ}), Encouraging Self-Exploration (\textbf{SE}), Cognitive Restructuring (\textbf{CR}), Guided Questioning (\textbf{GQ}), and Non-judgmental Acceptance (\textbf{NJA}).

\paragraph{Dialogue-level evaluation.}
To assess whether each model can sustain a coherent and helpful process beyond individual responses, we conduct dialogue-level evaluation.
We sample 100 distinct client profiles from PsychEval~\cite{pan2026psycheval}, covering six topics such as interpersonal relationships and emotion management, and use \texttt{gemini-3.1-flash-lite} to simulate clients according to these profiles.
Each counselor model interacts with the same set of profiles, and each session begins with the same neutral greeting and ends naturally or upon reaching 40 utterances.
Drawing on established counseling constructs and recent interactive evaluation frameworks~\cite{horvath1989development,rogers1957necessary,hill2009helping,pan2026psycheval}, we evaluate Therapeutic Alliance (\textbf{TA}), Empathic Understanding (\textbf{EU}), Problem Understanding (\textbf{PU}), Exploration and Intervention Quality (\textbf{EI}), Process Coherence and Progress (\textbf{PC}), and Safety and Boundary (\textbf{SB}).
The complete client-simulation and evaluation prompts are provided in Appendix~\ref{sec:appendix_client_simulation_prompt} and Appendix~\ref{sec:appendix_dialogue_level_evaluation_prompt}.

\subsection{Results}

As shown in Table~\ref{tab:counseltrajlm_evaluation}, CounselTrajLM achieves the highest response-level overall score of 3.34, compared with 3.24 for SoulChat2.0.
It leads in active listening, empathy, issue clarification, cognitive restructuring, and non-judgmental acceptance, while SoulChat2.0 performs better on the three question-oriented dimensions.
At the dialogue level, CounselTrajLM ranks first across all six dimensions and achieves an overall score of 4.70, substantially exceeding SoulChat2.0 at 3.82.
These results support the value of CounselTrajD for both next-response generation and coherent session-level interaction.

\section{Conclusion}

We present \textbf{CounselTraj}, which learns dialogue-act patterns from role-play sessions involving human counselors and uses content-conditioned trajectories to expand brief synthetic dialogues into process-aware session-level conversations.
Across three synthetic datasets, CounselTraj produces more complete and coherent conversations than direct and random-trajectory expansion.
We further construct \textbf{CounselTrajD}, a topic-balanced Chinese dataset of 10,964 session-level dialogues, and train \textbf{CounselTrajLM} for counselor response generation.
CounselTrajLM achieves the strongest overall performance in both response- and dialogue-level evaluations, demonstrating the value of session-level process guidance for dialogue synthesis and downstream model training.

\begin{samepage}

\section*{Acknowledgments}

This work was supported by JSPS KAKENHI Grant Number~\mbox{25H01382} and MEXT Supporting Pioneering Research through AI for 1,000 Discovery Challenges Program (SPReAD), Japan, under Grant Number~\mbox{JPMXP1726302875}.

\par
\end{samepage}

\section*{Limitations}

First, our study focuses on Chinese online text-based counseling dialogue synthesis.
KokoroChat is well aligned with this setting because it contains long role-play sessions involving trained counselors, although such interactions may not capture all behaviors of actual clients or face-to-face counseling.
Accordingly, our conclusions are limited to counseling-oriented text generation rather than clinical practice, and broader cross-lingual and cross-modal validation is left for future work.

Second, we do not exhaustively compare different planner and generator scales.
To isolate the contribution of trajectory guidance, all variants in the controlled evaluation use the same \texttt{gemini-3.1-flash-lite} generator.
For large-scale construction, we use \texttt{gpt-5.5} to improve data quality, demonstrating that CounselTraj can also work with a stronger generator.
Future work may examine how different model combinations affect performance.

Finally, dialogue-act annotation and automatic evaluation rely partly on LLMs.
We mitigate this limitation through human agreement validation, two independent evaluation models, randomized candidate order, human evaluation, and comparison with professional-book counseling dialogues.
Nevertheless, these evaluations measure counseling-oriented dialogue quality rather than clinical effectiveness, which would require separate expert-led studies.

\section*{Ethical Considerations}

\paragraph{Data Use.}
We use KokoroChat~\cite{qi-etal-2025-kokorochat}, SmileChat~\cite{qiu-etal-2024-smile}, CPsyCounD~\cite{zhang-etal-2024-cpsycoun}, PsyDTCorpus~\cite{xie-etal-2025-psydt}, PsyDial-D101~\cite{qiu-lan-2025-psydial}, and PsychEval~\cite{pan2026psycheval} in accordance with their respective licenses and access conditions.
CounselTrajD is generated from synthetic seed dialogues and does not intentionally introduce real personal information.
Before release, we will conduct privacy and license reviews and release only permitted components.

\paragraph{Annotation and Evaluation.}
Two Japanese crowd workers participated in dialogue-act annotation validation, for which the compensation was 3,300 JPY.
Three undergraduate psychology students conducted the human evaluation and were compensated at 50 RMB per hour.
All participants were informed of the study purpose, participated voluntarily, and were not asked to provide counseling or disclose personal experiences.

\paragraph{Intended Use.}
CounselTrajD and CounselTrajLM are intended solely for research.
CounselTrajLM has not been clinically validated and should not be used for standalone counseling, diagnosis, treatment, or crisis intervention.
Any real-world use would require professional validation, safety assessment, and human oversight.


\bibliography{custom}

\appendix



\section{Additional Details of Data Analysis}
\label{sec:appendix_data_analysis}

This section provides additional details for the data analysis in Section~\ref{sec:analysis}, including examples of template-like counselor responses, detailed statistics of client-side dialogue-act annotations, stage-wise dialogue-act trends, and prompts used for automatic annotation.

\subsection{Examples of Template-like Counselor Responses}
\label{sec:appendix_template_responses}

Table~\ref{tab:appendix_template_response_psydt} and Table~\ref{tab:appendix_template_response_cpsy} show examples of template-like counselor responses from PsyDTCorpus~\cite{xie-etal-2025-psydt} and CPsyCounD~\cite{zhang-etal-2024-cpsycoun}, respectively. 
The PsyDTCorpus example corresponds to \texttt{id=3864} in \texttt{PsyDTCorpus\_train\_mulit\_turn\_packing.json} from the Hugging Face dataset.\footnote{\url{https://huggingface.co/datasets/lilili123456/PsyDTCorpus/blob/main/PsyDTCorpus_train_mulit_turn_packing.json}}
The CPsyCounD example corresponds to \texttt{case=1757} and is available from the CPsyCounD repository.\footnote{\url{https://github.com/CAS-SIAT-XinHai/CPsyCoun/tree/main/CPsyCounD}}

\begin{table*}[t]
\centering
\scriptsize
\setlength{\tabcolsep}{5pt}
\renewcommand{\arraystretch}{1.18}
\begin{tabularx}{\textwidth}{@{}p{0.11\textwidth}X@{}}
\toprule
\textbf{Role} & \textbf{Utterance} \\
\midrule

Client &
\zh{你好，我近来有点不知所措。我男朋友总是给我一些惊喜，虽然我很开心，但也会有点愧疚感。}
(\textit{Hello. Lately, I have been feeling a bit overwhelmed. My boyfriend often gives me little surprises. Although I feel very happy, I also feel somewhat guilty.}) \\

\rowcolor{gray!8}
Counselor &
\textbf{\zh{你好呀，能够听到你的内心声音我感到很荣幸。让我们一起慢慢探讨这份愧疚感的来源吧。}}
\textbf{\zh{你能描述一下，每当收到他的惊喜，你的感受如何转变的吗？}}
(\textit{Hello. I feel honored to hear your inner thoughts. Let us slowly explore where this sense of guilt comes from. Could you describe how your feelings change whenever you receive one of his surprises?}) \\

Client &
\zh{嗯，每次当我收到他的惊喜，比如他送的水果或者奶茶，我一开始真的很高兴。但很快，我就会开始感到不安，感觉自己应该以某种方式回报他。}
(\textit{Well, every time I receive a surprise from him, such as fruit or milk tea, I feel really happy at first. But very quickly, I start to feel uneasy, as if I should repay him in some way.}) \\

\rowcolor{gray!8}
Counselor &
\textbf{\zh{你的感受很正常，很多人在接收到别人的好意时都会有想要回报的冲动。}}
\textbf{\zh{这种想法背后，你认为自己为何感到需要回报呢？}}
(\textit{Your feelings are very normal. Many people feel an impulse to return the favor when they receive kindness from others. Behind this thought, why do you think you feel the need to repay him?}) \\

Client &
\zh{我不知道，可能是因为我不想感觉欠他的，总觉得应该回赠同等价值的礼物或者直接转账给他。}
(\textit{I am not sure. Maybe it is because I do not want to feel like I owe him. I always feel that I should give him a gift of equal value in return, or transfer money to him directly.}) \\

\rowcolor{gray!8}
Counselor &
\textbf{\zh{感谢你的坦诚。在你眼中，接受礼物是一种亏欠感，这可能与你对平等的理解有关。}}
\textbf{\zh{这样的想法对你来说很重要吗？}}
(\textit{Thank you for being honest. In your view, accepting a gift seems to create a sense of indebtedness, which may be related to your understanding of equality. Is this idea very important to you?}) \\

Client &
\zh{是的，我想我一直都很重视公平和平等，无论是在工作还是在人际关系上。}
(\textit{Yes. I think I have always valued fairness and equality, whether at work or in relationships.}) \\

\rowcolor{gray!8}
Counselor &
\textbf{\zh{你很重视公平和平等，这体现了你的责任心和宜人性。}}
\textbf{\zh{在你的男朋友看来，你的这种回报方式是如何的呢？}}
(\textit{You place great importance on fairness and equality, which reflects your sense of responsibility and agreeableness. From your boyfriend's perspective, how does he see this way of repaying him?}) \\

Client &
\zh{他不太高兴，说我这是在撇清我们之间的关系，他可能不理解我的想法吧。}
(\textit{He is not very happy about it. He says I am trying to distance myself from our relationship. Maybe he does not understand what I am thinking.}) \\

\rowcolor{gray!8}
Counselor &
\textbf{\zh{感受到伴侣的不高兴确实挺难受的。}}
\textbf{\zh{你有尝试跟他解释你的想法吗，或者你们有没有就这个问题进行过深入的交流？}}
(\textit{It must feel quite difficult to sense that your partner is unhappy. Have you tried explaining your thoughts to him, or have the two of you had a deeper conversation about this issue?}) \\

Client &
\zh{我试过，但每次我们都在这个问题上争执。我尝试解释我只是想要公平交换，但他似乎不买账。}
(\textit{I have tried, but every time we end up arguing about it. I try to explain that I just want a fair exchange, but he does not seem to accept it.}) \\

\rowcolor{gray!8}
Counselor &
\textbf{\zh{我理解你们之间出现了一些沟通上的误解。也许你们需要找到一种双方都感觉舒适的表达方式。}}
\textbf{\zh{你认为愿意尝试新的方式来解决这个状况吗？}}
(\textit{I understand that there seem to be some communication misunderstandings between you two. Perhaps you need to find a way of expressing yourselves that feels comfortable for both of you. Would you be willing to try a new approach to resolve this situation?}) \\

Client &
\zh{我愿意尝试，但真不知道应该怎么做。你有什么建议吗？}
(\textit{I am willing to try, but I really do not know what to do. Do you have any suggestions?}) \\

\rowcolor{gray!8}
Counselor &
\textbf{\zh{既然你对于物质上的回馈感到困扰，我们可以从情感上的回馈出发。}}
\zh{例如，你可以尝试表达你的感激之情，或者通过共度美好时光来回报他的礼物。}
\textbf{\zh{听起来你会怎样想？}}
(\textit{Since you feel troubled by material forms of repayment, we can start with emotional forms of giving back. For example, you could try expressing your gratitude, or repay his gifts by spending meaningful time together. How does that sound to you?}) \\

Client &
\zh{这听起来不错。但如果不是同等价值的话，我可能还是会有点不舒服。}
(\textit{That sounds good. But if it is not of equal value, I may still feel a little uncomfortable.}) \\

\rowcolor{gray!8}
Counselor &
\textbf{\zh{这很正常。我们可以一步步来，首先是调整你对于回报的定义。}}
\textbf{\zh{你是否认为情感上的回馈也同样珍贵？}}
(\textit{That is very normal. We can take it step by step. First, we can adjust your definition of repayment. Do you think emotional responses can be just as valuable?}) \\

Client &
\zh{我想是的，情感的回馈应该同样重要。}
(\textit{I think so. Emotional responses should be just as important.}) \\

\rowcolor{gray!8}
Counselor &
\textbf{\zh{这是一个很好的开始，你已经开始拓展你对于回馈的理解了。让我们从小事做起。}}
\textbf{\zh{你有没有想过为他做一些特别的事情，这些不一定需要花费金钱，但可能会给他留下深刻的印象？}}
(\textit{That is a very good start. You have already begun to broaden your understanding of giving back. Let us start with small things. Have you ever thought about doing something special for him, something that does not necessarily cost money but may leave a deep impression on him?}) \\

Client &
\zh{我可以尝试做他喜欢吃的晚餐，或者为我们的约会制定一个特别的计划。}
(\textit{I could try making a dinner he likes, or planning a special date for us.}) \\

\rowcolor{gray!8}
Counselor &
\textbf{\zh{这个想法很棒！感觉上你已经有一些新的思考了。}}
\textbf{\zh{你对于试着做出这样的改变感到怎样？}}
(\textit{That is a wonderful idea. It seems that you already have some new thoughts. How do you feel about trying to make this kind of change?}) \\

\bottomrule
\end{tabularx}
\caption{Example of template-like counselor responses from PsyDTCorpus. The counselor shows a repeated empathy-then-question response tendency. English translations in parentheses were generated by GPT-5.5.}
\label{tab:appendix_template_response_psydt}
\end{table*}

\begin{table*}[t]
\centering
\scriptsize
\setlength{\tabcolsep}{5pt}
\renewcommand{\arraystretch}{1.18}
\begin{tabularx}{\textwidth}{@{}p{0.11\textwidth}X@{}}
\toprule
\textbf{Role} & \textbf{Utterance} \\
\midrule

Client &
\zh{老师，您好，最近我因为朋友餐厅股权分配的问题，心情一直很烦躁，晚上也睡不好。}
(\textit{Hello, counselor. Recently, I have been feeling very irritable because of an issue involving the equity distribution of my friend's restaurant, and I have not been sleeping well at night.}) \\

\rowcolor{gray!8}
Counselor &
\textbf{\zh{您好，阿明。听起来您在这件事上遇到了困扰。}}
\textbf{\zh{能否详细告诉我发生了什么事情？}}
(\textit{Hello, Amin. It sounds like this matter has been troubling you. Could you tell me in more detail what happened?}) \\

Client &
\zh{是的，我和那个餐厅的老板是朋友，他打算把餐厅的股权分给员工。其中一个店长因为家庭经济困难，无法拿出钱来，我就主动提出借给他，并答应享受他分红的一半作为利息。但老板并不赞同我的做法，我们还为此吵了一架。}
(\textit{Yes. The owner of that restaurant is a friend of mine. He plans to distribute shares of the restaurant to the employees. One of the store managers is in financial difficulty and cannot afford to pay for the shares, so I offered to lend him the money and agreed that I would receive half of his dividends as interest. But the owner did not approve of what I did, and we had an argument about it.}) \\

\rowcolor{gray!8}
Counselor &
\textbf{\zh{明白了，您为了帮助店长而与老板发生争执。}}
\textbf{\zh{那么，这让您想起了过去的哪些经历呢？}}
(\textit{I see. You wanted to help the store manager, but this led to a conflict with the owner. What past experiences does this remind you of?}) \\

Client &
\zh{我想起了我曾经最困难的一段日子，当时我陷入了绝望，甚至想过自杀。是因为有人帮助了我，我才得以走出那段黑暗时光。}
(\textit{It reminds me of the most difficult period of my life. At that time, I was in despair and even thought about suicide. It was because someone helped me that I was able to get through that dark time.}) \\

\rowcolor{gray!8}
Counselor &
\textbf{\zh{听起来那段经历对您影响很大。}}
\textbf{\zh{您现在的感受是怎样的？}}
(\textit{It sounds like that experience had a profound impact on you. How are you feeling now?}) \\

Client &
\zh{我觉得很难过，我为那个店长感到无助，同时也为自己曾经的无助感到难过。}
(\textit{I feel very sad. I feel helpless for the store manager, and I also feel sad when I think about how helpless I once felt.}) \\

\rowcolor{gray!8}
Counselor &
\textbf{\zh{我能理解您的感受。}}
\textbf{\zh{那么，您觉得这次的情况与过去相比如何？}}
(\textit{I understand how you feel. Then, how do you think this situation compares with your past experience?}) \\

Client &
\zh{相比过去，我现在的生活已经好很多了。但我还是忍不住想去帮助那个店长，因为我不想看到他陷入困境。}
(\textit{Compared with the past, my life is much better now. But I still cannot help wanting to help that store manager, because I do not want to see him fall into hardship.}) \\

\rowcolor{gray!8}
Counselor &
\textbf{\zh{那么，您认为如何才能更好地帮助他呢？}}
(\textit{Then, what do you think would be a better way to help him?}) \\

Client &
\zh{我想让他知道，每个人都有困难的时候，关键是要有信心去面对。同时，我也需要学会放手，不再过度承担责任。}
(\textit{I want him to know that everyone goes through difficult times, and the key is to have the confidence to face them. At the same time, I also need to learn to let go and stop taking on too much responsibility.}) \\

\rowcolor{gray!8}
Counselor &
\textbf{\zh{非常好，阿明。您已经找到了解决问题的方法。}}
\zh{接下来，我们可以通过一些心理咨询技术，比如格式塔疗法，帮助您更好地应对类似情况。}
\textbf{\zh{您愿意试试吗？}}
(\textit{Very good, Amin. You have already found a way to approach the problem. Next, we can use some counseling techniques, such as Gestalt therapy, to help you better cope with similar situations. Would you be willing to try?}) \\

Client &
\zh{愿意，我希望能够学会更好地面对过去和现在的问题。}
(\textit{Yes, I would. I hope I can learn to better face both my past and my present problems.}) \\

\rowcolor{gray!8}
Counselor &
\textbf{\zh{那么，让我们从了解您的内心需求开始。}}
\textbf{\zh{您能告诉我，您对一个理想的婚姻生活有什么期待吗？}}
(\textit{Then let us begin by understanding your inner needs. Could you tell me what expectations you have for an ideal married life?}) \\

Client &
\zh{我希望能有一个温馨、安静的家，平淡的生活就这样一直到老。我希望我的丈夫能够更多地陪在我和孩子身边。}
(\textit{I hope to have a warm and peaceful home, and to live an ordinary life together until old age. I hope my husband can spend more time with me and our children.}) \\

\rowcolor{gray!8}
Counselor &
\textbf{\zh{我明白了，阿明。接下来，我们将一起努力实现您的期待。}}
\zh{在这个过程中，如果您有任何情绪和需求，请随时告诉我，我们会一起面对。}
\textbf{\zh{您准备好了吗？}}
(\textit{I understand, Amin. Next, we will work together toward realizing your expectations. During this process, please feel free to tell me about any emotions or needs you have, and we will face them together. Are you ready?}) \\

Client &
\zh{好的，老师，我准备好了。谢谢您给我带来的帮助和支持。}
(\textit{Yes, counselor, I am ready. Thank you for the help and support you have given me.}) \\

\bottomrule
\end{tabularx}
\caption{Example of template-like counselor responses from CPsyCounD. The counselor shows a repeated summary-or-empathy-then-question response tendency. English translations in parentheses were generated by GPT-5.5.}
\label{tab:appendix_template_response_cpsy}
\end{table*}

\subsection{Detailed Statistics of Client-side Dialogue Acts}
\label{sec:appendix_da_statistics}

Table~\ref{tab:client_da_detailed_statistics} provides detailed counts and percentages of client-side dialogue-act tags for each dataset. These statistics complement the distributional analysis in Figure~\ref{fig:tag_distribution}.

\begin{table*}[t]
\centering
\scriptsize
\setlength{\tabcolsep}{3.2pt}
\renewcommand{\arraystretch}{1.12}
\resizebox{\textwidth}{!}{
\begin{tabular}{lccccccc}
\toprule
\textbf{Item} 
& \textbf{KokoroChat} 
& \textbf{AnnoMI} 
& \textbf{PsyDial-D4} 
& \textbf{PsyDTCorpus} 
& \textbf{SmileChat} 
& \textbf{CPsyCounD} 
& \textbf{CACTUS} \\
\midrule

Dialogues 
& 500 & 132 & 500 & 500 & 500 & 500 & 500 \\

Client utterances 
& 22,641 & 4,716 & 17,920 & 8,957 & 6,054 & 4,031 & 7,461 \\

Avg. client utt. 
& 45.28 & 35.73 & 35.84 & 17.91 & 12.11 & 8.06 & 14.92 \\

\midrule

SimpleAnswer 
& 3,729 (16.47\%) & 710 (15.06\%) & 1,028 (5.74\%) & 266 (2.97\%) & 73 (1.21\%) & 88 (2.18\%) & 146 (1.96\%) \\

DetailedDescription 
& 7,561 (33.40\%) & 1,535 (32.55\%) & 7,480 (41.74\%) & 3,873 (43.24\%) & 1,746 (28.84\%) & 1,482 (36.77\%) & 3,355 (44.97\%) \\

Hesitation 
& 351 (1.55\%) & 169 (3.58\%) & 32 (0.18\%) & 106 (1.18\%) & 6 (0.10\%) & 32 (0.79\%) & 220 (2.95\%) \\

Resistance 
& 232 (1.02\%) & 96 (2.04\%) & 12 (0.07\%) & 9 (0.10\%) & 9 (0.15\%) & 6 (0.15\%) & 61 (0.82\%) \\

Agreement 
& 1,832 (8.09\%) & 638 (13.53\%) & 2,033 (11.34\%) & 1,023 (11.42\%) & 1,157 (19.11\%) & 614 (15.23\%) & 1,109 (14.86\%) \\

Request 
& 302 (1.33\%) & 48 (1.02\%) & 220 (1.23\%) & 484 (5.40\%) & 633 (10.46\%) & 517 (12.83\%) & 32 (0.43\%) \\

UnderstandingCheck 
& 414 (1.83\%) & 109 (2.31\%) & 58 (0.32\%) & 61 (0.68\%) & 40 (0.66\%) & 30 (0.74\%) & 31 (0.42\%) \\

Question 
& 277 (1.22\%) & 54 (1.15\%) & 100 (0.56\%) & 131 (1.46\%) & 199 (3.29\%) & 74 (1.84\%) & 32 (0.43\%) \\

SelfDisclosure 
& 3,007 (13.28\%) & 598 (12.68\%) & 4,692 (26.18\%) & 1,885 (21.04\%) & 1,048 (17.31\%) & 644 (15.98\%) & 1,929 (25.85\%) \\

PositiveEmotion 
& 608 (2.69\%) & 80 (1.70\%) & 1,265 (7.06\%) & 818 (9.13\%) & 636 (10.51\%) & 442 (10.97\%) & 392 (5.25\%) \\

Backchannel 
& 1,995 (8.81\%) & 587 (12.45\%) & 34 (0.19\%) & 0 (0.00\%) & 1 (0.02\%) & 0 (0.00\%) & 0 (0.00\%) \\

Greeting 
& 822 (3.63\%) & 25 (0.53\%) & 438 (2.44\%) & 58 (0.65\%) & 6 (0.10\%) & 41 (1.02\%) & 7 (0.09\%) \\

Thanking 
& 1,213 (5.36\%) & 57 (1.21\%) & 455 (2.54\%) & 236 (2.63\%) & 491 (8.11\%) & 57 (1.41\%) & 145 (1.94\%) \\

Other 
& 210 (0.93\%) & 10 (0.21\%) & 72 (0.40\%) & 7 (0.08\%) & 9 (0.15\%) & 4 (0.10\%) & 2 (0.03\%) \\

\bottomrule
\end{tabular}
}
\caption{Detailed statistics of client-side dialogue-act annotations across datasets. Each tag cell reports count and percentage. Avg. client utt. denotes the average number of client utterances per dialogue.}
\label{tab:client_da_detailed_statistics}
\end{table*}

\subsection{Stage-wise Trends of Client-side Dialogue Acts}
\label{sec:appendix_da_stage_trends}

\begin{figure*}[t]
    \centering
    \includegraphics[width=0.8\textwidth]{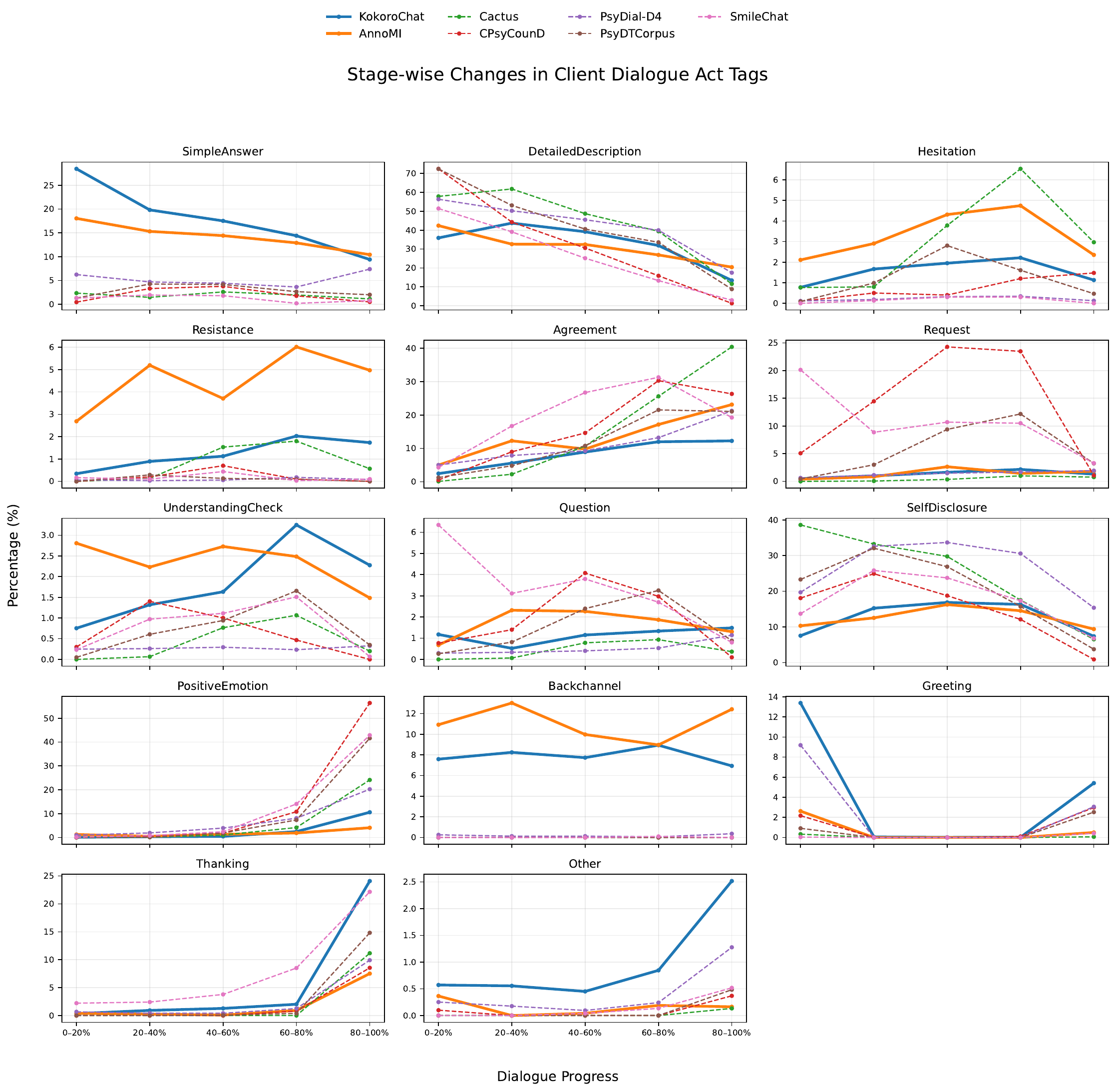}
    \caption{Stage-wise changes in client-side dialogue act tags across datasets. Solid lines indicate datasets with human clients.}
    \label{fig:all_tags_stage_transition}
\end{figure*}

To further examine how client behaviors change over the course of counseling, we analyze the stage-wise trends of each client-side dialogue act. Each dialogue is divided into five equal stages according to dialogue progress, and the frequency of each tag is computed for each stage. Figure~\ref{fig:all_tags_stage_transition} shows the stage-wise changes of client-side dialogue act tags across datasets. Solid lines indicate datasets with human clients.

The results reveal several differences between human and synthetic clients. First, \textit{SimpleAnswer} and \textit{Backchannel}, which are closely related to natural conversational flow, appear more frequently in human-client datasets. In particular, \textit{SimpleAnswer} in human-client data gradually decreases as the dialogue progresses. This suggests that human clients often begin the session with short and simple responses, and then gradually move toward more elaborated expressions as the counseling dialogue develops. \textit{Backchannel} also appears consistently in human-client datasets, especially in KokoroChat and AnnoMI, whereas it is almost absent in most synthetic datasets. This indicates that human clients naturally use brief responses to maintain interactional rhythm, while synthetic clients tend to lack such subtle conversational behaviors.

Another notable difference is observed in \textit{Resistance}. Human-client datasets show more frequent \textit{Resistance} than most synthetic datasets, suggesting that human counseling interactions often involve hesitation, reluctance, disagreement, or difficulty in immediately accepting the counselor's perspective. Such behaviors are important for representing the gradual and sometimes non-linear process of counseling. In contrast, synthetic clients appear to be more cooperative and less resistant, which may make the dialogue progression overly smooth.

Synthetic clients also show different tendencies in advice-seeking behaviors. Some synthetic datasets contain much more frequent \textit{Request}, indicating that synthetic clients are more likely to explicitly ask for advice or solutions. This may reflect a simplified view of counseling conversations, where the client clearly presents a problem and directly asks the counselor what to do. However, in human-client datasets, requests are less dominant, and clients often express their concerns through descriptions, hesitation, or self-disclosure rather than direct advice-seeking.

Finally, synthetic clients, especially in the later stages of the dialogue, tend to express \textit{PositiveEmotion} more frequently. This suggests that synthetic counseling dialogues may often end with an overly positive or simplified closing pattern, where the client quickly becomes reassured or satisfied. Human-client datasets show a more moderate increase, indicating that real clients may not always shift strongly toward positive emotions by the end of the session.

Overall, these stage-wise patterns suggest that synthetic clients differ from human clients not only in the overall frequency of dialogue act tags, but also in how these behaviors are distributed across the course of counseling. Human-client dialogues contain more interactional and gradual behaviors, such as short responses, backchannels, and resistance, whereas synthetic dialogues tend to show more explicit requests and stronger positive emotional expressions toward the end. These findings highlight the importance of modeling stage-wise client behaviors when constructing process-aware synthetic counseling dialogues.

\subsection{Prompts for Client-side Dialogue-act Annotation}
\label{sec:appendix_da_prompts_}

This section presents the prompts used for automatic dialogue-act annotation of client utterances. Each client utterance was assigned one of the predefined client-side dialogue act tags listed in Table~\ref{tab:dialogue_act_tags}. We used \texttt{gemini-3.1-flash-lite} as the annotation model. 
The English prompt is shown in Figure~\ref{fig:prompt_client_da_annotation_en}.

\begin{figure*}[t]
\centering
\begin{tcolorbox}[
  width=\textwidth,
  colback=gray!15,
  colframe=gray!70,
  coltitle=white,
  colbacktitle=gray,
  title=\texttt{Prompt for Client-side Dialogue-act Annotation},
  fonttitle=\bfseries\sffamily,
  fontupper=\scriptsize,
  boxrule=0.4mm,
  arc=2mm,
  left=2mm,
  right=2mm,
  top=1mm,
  bottom=1mm,
  sharp corners=south,
  enhanced jigsaw
]

{\normalsize \textbf{\# Task Description}} \\
Annotate the dialogue acts of client utterances in a counseling conversation. Utterances by the counselor should NOT be annotated.

For each client utterance, select the single most appropriate label from the list below that best represents its intent or psychological state.

\medskip

{\normalsize \textbf{\# Label Selection Guidelines}} \\

\textbf{1. Context Awareness:} \\
Do not judge each utterance in isolation. Consider the previous counselor utterance and the overall dialogue context.

\medskip

\textbf{2. Priority (when multiple intents exist):} \\
Choose the MOST dominant function of the utterance in the dialogue. \\
Example: If a factual explanation (\textit{DetailedDescription}) contains deep emotional expression (\textit{SelfDisclosure}), prioritize \textit{SelfDisclosure} in a counseling context.

\medskip

\textbf{3. Split Utterances:} \\
If a single utterance is split into multiple lines due to formatting, such as a chat interface, treat them as one utterance and assign a single label.

\medskip

{\normalsize \textbf{\# Dialogue Act Label Set}}

\medskip

\textbf{1. Answering / Describing (focus on length and detail)} \\
\textbf{SimpleAnswer}: A short, direct factual answer without elaboration. \\
\textbf{DetailedDescription}: A detailed explanation of facts, situations, or background. Includes self-justification or topic shifting if informative. \\
\textbf{Hesitation}: Expressions of uncertainty, hesitation, or pause.

\medskip

\textbf{2. Attitude toward Counselor (acceptance / resistance)} \\
\textbf{Resistance}: Rejecting, opposing, or resisting the counselor's intervention or suggestion. \\
\textbf{Agreement}: Accepting or strongly agreeing with the counselor's explanation or interpretation. 

\medskip

\textbf{3. Asking Questions} \\
\textbf{Request}: Asking for advice, solutions, or concrete guidance. \\
\textbf{UnderstandingCheck}: Asking to clarify or confirm the meaning of the counselor's statement. \\
\textbf{Question}: General questions not covered by the above two categories.

\medskip

\textbf{4. Spontaneous Emotional Expression} \\
\textbf{SelfDisclosure}: Expressing deep emotions, vulnerability, core beliefs, or self-critical thoughts beyond simple description. \\
\textbf{PositiveEmotion}: Expressing relief, gratitude, hope, or other positive emotional changes.

\medskip

\textbf{5. Others} \\
\textbf{Backchannel}: Short responses indicating active listening. \\
\textbf{Greeting}: Greetings. \\
\textbf{Thanking}: Simple expressions of thanks. \\
\textbf{Other}: Anything not covered above, such as technical issues or leaving temporarily.

\medskip

{\normalsize \textbf{\# Input Format}} \\
The input is provided in JSON format.

\medskip

- Target: Utterances where \texttt{role = "client"} \\
- Exclude: Utterances where \texttt{role = "counselor"}

\medskip

{\normalsize \textbf{\# Output Format}} \\
Return the result in JSON format.

For each client utterance, keep the original: \\
- \texttt{role} \\
- \texttt{utterance}

And add: \\
- \texttt{think}: reasoning process \\
\quad - How the context is interpreted, such as the previous counselor utterance. \\
\quad - Possible candidate labels. \\
\quad - Why the final label is chosen, and why others are not suitable, briefly explained.

- \texttt{tag}: the final selected label. Choose exactly ONE from the list above.

\medskip

{\normalsize \textbf{\# Additional Output Rules}} \\
- Counselor utterances must remain unchanged. \\
- Do NOT add \texttt{think} or \texttt{tag} to counselor utterances. \\
- Only client utterances should be annotated.

\medskip

{\normalsize \textbf{\# Important}} \\
The output MUST be valid JSON enclosed in a Markdown code block.

\medskip

{\normalsize \textbf{\# Input Data}} \\
\textcolor{red!70!black}{\textbf{\{JSON input\}}}

\end{tcolorbox}
\caption{Prompt for client-side dialogue-act annotation.}
\label{fig:prompt_client_da_annotation_en}
\end{figure*}

\section{Details of the Proposed Method}
\label{sec:appendix_method_details}

This section provides additional details of our proposed method. 

\subsection{Dialogue Act Tag Definitions and Counselor-side Annotation Prompt}
\label{sec:appendix_da_tag_definitions_counselor_prompt}

Table~\ref{tab:dialogue_act_tag_definitions} lists the complete dialogue act tag set and their corresponding definitions. These tags include both client-side and counselor-side dialogue acts, enabling us to analyze the interactional behaviors of both participants in counseling dialogues.

We also present the counselor-side prompt used for automatic dialogue-act annotation of KokoroChat. Counselor utterances were annotated using \texttt{gemini-2.5-flash}. The English prompt is shown in Figure~\ref{fig:prompt_counselor_da_annotation_en}.

\begin{table*}[t]
\centering
\small
\setlength{\tabcolsep}{5pt}
\renewcommand{\arraystretch}{1.30}
\newcommand{\tagline}{\arrayrulecolor{gray!18}\cline{2-3}\arrayrulecolor{black}}
\begin{tabularx}{1\textwidth}{
@{}
>{\raggedright\arraybackslash}p{0.075\textwidth}
>{\raggedright\arraybackslash}p{0.19\textwidth}
>{\raggedright\arraybackslash}X
@{}}
\toprule
\textbf{Role} & \textbf{Tag} & \textbf{Description} \\
\midrule
\addlinespace[2pt]
\multirow{14}{*}{Client}
& DetailedDescription 
& Detailed descriptions of facts, situations, background, or context. \\
\tagline
& UnderstandingCheck 
& Questions or checks to confirm the meaning of the counselor's utterance. \\
\tagline
& SimpleAnswer 
& Short factual answers to counselor questions without detailed explanation. \\
\tagline
& Agreement 
& Acceptance of or agreement with the counselor's opinion, interpretation, or empathy. \\
\tagline
& Request 
& Requests for advice, solutions, instructions, or concrete guidance from the counselor. \\
\tagline
& SelfDisclosure 
& Disclosure of deep feelings, vulnerability, core values, or self-critical thoughts. \\
\tagline
& PositiveEmotion 
& Expressions of relief, gratitude, hope, or other positive emotional changes. \\
\tagline
& Backchannel 
& Minimal responses indicating that the client is listening. \\
\tagline
& Question 
& General questions that are not requests for advice or understanding checks. \\
\tagline
& Hesitation 
& Uncertainty, hesitation, silence, or difficulty in formulating a response. \\
\tagline
& Resistance 
& Rejection, reluctance, or resistance toward counselor advice, interpretation, or intervention. \\
\tagline
& Greeting 
& Greeting or closing expressions. \\
\tagline
& Thanking 
& Simple expressions of gratitude. \\
\tagline
& Other 
& Utterances that do not fit the above categories, such as system checks or temporary absence. \\
\addlinespace[3pt]
\midrule
\addlinespace[2pt]
\multirow{11}{*}{Counselor}
& OpenQuestion 
& Questions that allow the client to answer freely and broadly. \\
\tagline
& ClosedQuestion 
& Yes/no questions or questions with a highly restricted answer range. \\
\tagline
& Paraphrase 
& Restating or summarizing the client's content or facts to organize and confirm understanding. \\
\tagline
& Reflection 
& Reflecting the client's underlying emotions or meanings to support emotional clarification and self-exploration. \\
\tagline
& Affirm 
& Affirming the client's strengths, motivation, or abilities to provide reassurance and encouragement. \\
\tagline
& Suggest 
& Proposing actions, perspectives, coping methods, or possible solutions. \\
\tagline
& Inform 
& Providing useful information, facts, opinions, resources, or answers to the client. \\
\tagline
& Backchannel 
& Short responses indicating that the counselor is listening. \\
\tagline
& Greeting 
& Greeting or closing expressions. \\
\tagline
& Thanking 
& Expressions of gratitude. \\
\tagline
& Other 
& Session management, apologies, or utterances that do not fit the above categories. \\
\addlinespace[2pt]
\bottomrule
\end{tabularx}
\caption{Definitions of dialogue-act tags for client and counselor utterances.}
\label{tab:dialogue_act_tag_definitions}
\end{table*}

\begin{figure*}[t]
\centering
\begin{tcolorbox}[
  width=\textwidth,
  colback=gray!15,
  colframe=gray!70,
  coltitle=white,
  colbacktitle=gray,
  title=\texttt{Prompt for Counselor-side Dialogue-act Annotation on KokoroChat},
  fonttitle=\bfseries\sffamily,
  fontupper=\scriptsize,
  boxrule=0.4mm,
  arc=2mm,
  left=2mm,
  right=2mm,
  top=1mm,
  bottom=1mm,
  sharp corners=south,
  enhanced jigsaw
]

{\normalsize \textbf{\# Task Description}} \\
Annotate the dialogue acts of counselor utterances in a counseling dialogue. Client utterances are not the target of annotation. \\
For each utterance, select the single most appropriate dialogue-act tag from the following list that represents the speaker's intention. When assigning tags, consider not only the utterance itself but also the preceding dialogue context. \\
- If a single utterance contains multiple intentions, select the tag that represents the most important function of the utterance in the overall dialogue. For example, if an utterance contains both a greeting and a question, and the question plays a more important role in initiating the dialogue and eliciting information, select \textit{OpenQuestion}. \\
- If a single utterance appears to have been split into multiple messages, treat the split messages as one utterance and assign the same tag to each of them.

\medskip

{\normalsize \textbf{\# Dialogue Act Tag Set}} \\

\textbf{OpenQuestion}: A question that allows a wide range of possible answers. ``What do you think?'' \\
\textbf{ClosedQuestion}: A question that can be answered with yes or no, or whose possible answers are highly limited. ``Do you always do that?'' \\
\textbf{Paraphrase}: An utterance that summarizes the content or facts of the client's statement in the counselor's own words, in order to confirm understanding or organize the discussion. \\
\textbf{Reflection}: An utterance that captures the emotions or inner meanings behind the client's words and reflects them back empathically, in order to clarify the client's feelings and encourage self-exploration. \\
\textbf{Affirm}: An utterance that affirms the client's strengths, motivation, or abilities, and provides reassurance or encouragement. \\
\textbf{Suggest}: An utterance that proposes an action, way of thinking, new perspective, or solution. ``How about trying ...?'' \\
\textbf{Inform}: An utterance that provides useful information to the client through data, facts, opinions, resources, or answers to questions. \\
\textbf{Backchannel}: An utterance used to show that the counselor is listening to the client. ``Yes.'' ``I see.'' \\
\textbf{Greeting}: An utterance intended as a greeting. ``Hello.'' ``Nice to meet you.'' \\
\textbf{Thanking}: An utterance expressing gratitude. ``Thank you.'' \\
\textbf{Other}: An utterance that does not fall into any of the above categories, such as session management or apology. ``Please wait a moment.'' ``Excuse me.''

\medskip

{\normalsize \textbf{\# Input Format}} \\
The input is provided in JSON format. Utterances where \texttt{role} is \texttt{counselor} are the target of annotation, while utterances where \texttt{role} is \texttt{client} do not need to be annotated.

\medskip

{\normalsize \textbf{\# Output Format}} \\
Return the output in JSON format. Keep the original \texttt{role}, \texttt{time}, and \texttt{utterance} fields unchanged, and add \texttt{think} and \texttt{tag} to all utterances where \texttt{role} is \texttt{counselor}. \\

\texttt{think} should describe the reasoning process for selecting the most appropriate tag. In doing so, list multiple candidate tags, briefly explain why the final tag was selected, and why the other candidates were not optimal. \\

\texttt{tag} should contain the single most appropriate dialogue-act tag that represents the speaker's intention, based on the reasoning process. \\

The output must be only valid JSON enclosed in a Markdown code block. Do not include any additional explanation.

\medskip

{\normalsize \textbf{\# Input}} \\
\textcolor{red!70!black}{\textbf{\{JSON-formatted data\}}}

\end{tcolorbox}
\caption{Prompt for counselor-side dialogue-act annotation on KokoroChat.}
\label{fig:prompt_counselor_da_annotation_en}
\end{figure*}

\subsection{Prompt for Brief Dialogue Construction}
\label{sec:appendix_brief_dialogue_construction_prompt}

This subsection presents the prompt used for brief dialogue construction. 
In this step, Japanese KokoroChat dialogues are rewritten into concise Chinese dialogues with a more complete, structured, and logically organized client style. 
The generated brief dialogues are used as the input \(D\) for the dialogue-act trajectory planning model. 
The English prompt is shown in Figure~\ref{fig:prompt_brief_dialogue_construction_en}.

\begin{figure*}[t]
\centering
\begin{tcolorbox}[
  width=\textwidth,
  colback=gray!15,
  colframe=gray!70,
  coltitle=white,
  colbacktitle=gray,
  title=\texttt{Prompt for Brief Dialogue Construction},
  fonttitle=\bfseries\sffamily,
  fontupper=\scriptsize,
  boxrule=0.4mm,
  arc=2mm,
  left=2mm,
  right=2mm,
  top=1mm,
  bottom=1mm,
  sharp corners=south,
  enhanced jigsaw
]

{\normalsize \textbf{\# Task Description}} \\
Rewrite the following Japanese psychological counseling dialogue into Chinese, and convert it into a client dialogue style that is clearly structured, complete in expression, and logically organized. Refer to the few-shot examples as appropriate.

\medskip

{\normalsize \textbf{\# Rewriting Objective}} \\
While preserving the basic semantics of the original dialogue, make the client's expressions resemble a typical large language model (LLM)-generated style. At the same time, moderately integrate counselor utterances so that the overall dialogue structure becomes more balanced and natural.

\medskip

{\normalsize \textbf{\# Dialogue Length and Structure Requirements}} \\
- The total number of utterances should be around 25, for example between 20 and 30. \\
- The dialogue must strictly alternate between client and counselor utterances. \\
- Consecutive utterances by the same role are not allowed. \\
- If the original dialogue contains too many turns, adjacent utterances may be moderately integrated.

\medskip

{\normalsize \textbf{\# 1. Language Conversion}} \\
- Translate the dialogue from Japanese into natural and fluent Chinese. \\
- Preserve the original meaning and do not change the original content.

\medskip

{\normalsize \textbf{\# 2. Rewriting Client Utterances}} \\
Rewrite all client utterances according to the following requirements: \\
(1) Complete and information-rich expression: Each utterance should be expanded into a complete expression as much as possible. Even simple answers should include background or explanation. \\
(2) Clear structure and logical coherence: The expression should be organized, coherent, and may include causal relations or summaries. \\
(3) Early presentation of the core issue: The main concern and background should be clearly stated in the early part of the dialogue. \\
(4) Stronger initiative: The client should actively express thoughts and may actively ask questions or request advice. \\
(5) More acknowledgment and feedback: The client should include more understanding, agreement, or summary in response to the counselor. \\
(6) Clear emotional change: The later part of the dialogue should show clear understanding or improvement. \\
(7) Avoid short responses: Short responses such as ``um'' or ``yes'' should be rewritten as complete expressions.

\medskip

{\normalsize \textbf{\# 3. Handling Counselor Utterances}} \\
- Preserve the original meaning and translate them into natural Chinese. \\
- Consecutive or fragmented utterances may be moderately integrated. \\
- Integration should be performed only when the utterances are semantically continuous and belong to the same context. \\
- Make each utterance clearer and more complete, but avoid excessive expansion. \\
- Keep the counselor's expressions natural and avoid making them overly structured. \\
- Control the length so that client and counselor utterances are relatively balanced.

\medskip

{\normalsize \textbf{\# 4. Content Consistency Constraints}} \\
- The rewritten dialogue must be strictly based on the original dialogue content. \\
- Do not add any information, events, episodes, or background that does not appear in the original dialogue. \\
- Do not introduce new problems, people, experiences, or settings. \\
- Do not make speculative expansions or reasonable additions. \\
- Do not include any Japanese. Personal names may be replaced with generic names such as Xiao Li or Xiao Zhang. \\
Allowed operations include reorganizing expressions, expanding short expressions into complete ones, and merging or splitting utterances without changing the meaning.

\medskip

{\normalsize \textbf{\# Style Reinforcement}} \\
Make the client style highly information-dense, strongly logical, complete, and clear. 
The client expressions should be clearly distinguishable from natural human dialogue, for example by being more structured and explicit.

\medskip

{\normalsize \textbf{\# Output Format}} \\

\begin{verbatim}
[
  {"role": "client", "utterance": "..."},
  {"role": "counselor", "utterance": "..."}
]
\end{verbatim}

\medskip

{\normalsize \textbf{\# Few-shot Examples}} \\
Three few-shot examples are provided to demonstrate how to rewrite original psychological counseling dialogues into concise Chinese dialogues with clear structure, complete expression, and logical organization.

\medskip

\textcolor{red!70!black}{\textbf{\{FEW-SHOT EXAMPLES\}}}

\medskip

{\normalsize \textbf{\# Input Data}} \\
\textcolor{red!70!black}{\textbf{\{Insert Japanese dialogue JSON here\}}}

\end{tcolorbox}
\caption{Prompt for brief dialogue construction.}
\label{fig:prompt_brief_dialogue_construction_en}
\end{figure*}

\subsection{Details of Trajectory Planning for Synthetic Dialogues}
\label{sec:appendix_path_planning_synthetic_dialogues}

This subsection provides additional details of the dialogue-act trajectory planning procedure described in Section~\ref{sec:path_planning_synthetic_dialogues}. 
Given an existing brief synthetic counseling dialogue \(D=\{(r_j,u_j)\}_{j=1}^{n}\), the goal is to generate a session-level dialogue-act trajectory \(T=\{(r_i,a_i)\}_{i=1}^{N}\), where \(r_i\) and \(a_i\) denote the speaker role and dialogue-act tag at step \(i\), respectively. 

We generate the dialogue-act trajectory sequentially. 
The trajectory is initialized with greeting acts from both the counselor and the client, and the speaker role is then alternated at each step. 
At each generation step, the current dialogue progress is mapped into one of five stages, following the same stage division used in our analysis. 
Candidate tags are drawn from the role-specific tag inventory and filtered by compatibility constraints before scoring. 
The opening and closing acts are handled separately, and \textit{Greeting} is excluded during intermediate generation. Other exclusions prevent premature suggestions, positive-emotion expressions, or thanking, and repeated backchannels.

Each candidate tag is scored using a combination of model likelihood and lightweight statistical constraints derived from the annotated KokoroChat dialogues. 
The model likelihood is computed by the trained trajectory planning model conditioned on the brief dialogue and the recent tag history. 
In addition, we incorporate three empirical priors: a stage-specific prior, a transition prior conditioned on the previous role and tag, and a global role-specific prior. 
We further include content- and phase-compatibility scores, as described below, together with a repetition penalty to discourage locally repeated tag patterns and a diversity bonus to encourage tags that have not appeared recently.

Formally, for a candidate tag \(a\) at step \(i\), we compute the score as:
\[
\begin{aligned}
S(a) = 
&\lambda_{\mathrm{LM}} s_{\mathrm{LM}}(a)
+ \lambda_{\mathrm{stage}} \log p_{\mathrm{stage}}(a \mid s_i, r_i) \\
&+ \lambda_{\mathrm{trans}} \log p_{\mathrm{trans}}(a \mid s_i, r_{i-1}, a_{i-1}, r_i) \\
&+ \lambda_{\mathrm{global}} \log p_{\mathrm{global}}(a \mid r_i) \\
&+ \lambda_{\mathrm{content}} C_{\mathrm{content}}(a \mid D, r_i, \rho_i) \\
&+ \lambda_{\mathrm{phase}} C_{\mathrm{phase}}(a \mid r_i, \rho_i) \\
&- \lambda_{\mathrm{rep}} R(a)
+ \lambda_{\mathrm{div}} D(a),
\end{aligned}
\]
where \(s_i\) denotes the current dialogue stage, \(\rho_i=(i-1)/N\) is the relative generation progress, \(s_{\mathrm{LM}}\) is the model likelihood score, \(R(a)\) is the repetition penalty, and \(D(a)\) is the diversity bonus.

The content-compatibility score \(C_{\mathrm{content}}\) is a rule-based adjustment conditioned on the brief dialogue, speaker role, and generation progress. It uses lexical cues for emotion, requests, details, hesitation, positive affect, interpersonal relations, career concerns, crisis, and psychological knowledge, together with dialogue length and topic information. For example, emotional cues favor client \textit{SelfDisclosure} and counselor \textit{Reflection}, while requests favor \textit{Request} and \textit{Suggest}, subject to progress-dependent restrictions.
The phase-compatibility score \(C_{\mathrm{phase}}\) adds role-specific bonuses based on relative progress: early counselor acts favor \textit{OpenQuestion} and \textit{Reflection}, middle phases favor exploration followed by \textit{Suggest} and \textit{Inform}, and later client acts favor \textit{PositiveEmotion}, \textit{Agreement}, and \textit{Thanking}. These manually specified adjustments complement the empirical stage prior, which captures observed tag frequencies; the compatibility scores are additive biases rather than log-probabilities.

The weights and sampling settings for the content- and phase-aware planner used to construct CounselTrajD are shown in Table~\ref{tab:path_planning_weights}. 
We intentionally assign the largest positive weight to the model likelihood, setting \(\lambda_{\mathrm{LM}}=4.0\), so that the final decision is primarily guided by the trained trajectory planning model. 
In contrast, the stage, transition, and global priors are assigned much smaller weights and are used only as auxiliary constraints to encourage the generated trajectory to follow session-level tendencies observed in KokoroChat. 
The repetition penalty and diversity bonus are further introduced to reduce locally repetitive tag patterns and encourage moderate variation. 
The compatibility scores are included at each generation step, and incompatible acts are excluded rather than assigned a soft penalty.

\begin{table}[t]
\centering
\small
\setlength{\tabcolsep}{3pt}
\begin{tabular}{@{}l c@{}}
\hline
\textbf{Term} & \textbf{Weight / Value} \\
\hline
Model likelihood \(\lambda_{\mathrm{LM}}\) & 4.0 \\
Stage prior \(\lambda_{\mathrm{stage}}\) & 0.08 \\
Transition prior \(\lambda_{\mathrm{trans}}\) & 0.12 \\
Global prior \(\lambda_{\mathrm{global}}\) & 0.01 \\
Content compatibility \(\lambda_{\mathrm{content}}\) & 1.18 \\
Phase compatibility \(\lambda_{\mathrm{phase}}\) & 0.50 \\
Repetition penalty \(\lambda_{\mathrm{rep}}\) & 1.85 \\
Diversity bonus \(\lambda_{\mathrm{div}}\) & 0.52 \\
\hline
Top-\(k\) sampling & 5 \\
Top-\(p\) sampling & 0.88 \\
Temperature & 0.64 \\
Gap threshold & 0.48 \\
\hline
\end{tabular}
\caption{Hyperparameters of the content- and phase-aware dialogue-act trajectory planner used to construct CounselTrajD.}
\label{tab:path_planning_weights}
\end{table}

After scoring all candidates, we select the highest-scoring tag when the score gap between the top two candidates is larger than the threshold. 
When the gap is small, we apply temperature scaling to the top-\(k\) candidates, retain the smallest prefix whose cumulative probability reaches the top-\(p\) threshold, and sample from the renormalized distribution, allowing limited stochasticity among similarly plausible tags. 
Finally, we append a client \textit{Thanking} tag and a counselor \textit{Greeting} tag to complete the session-level dialogue-act trajectory. 
Algorithm~\ref{alg:path_planning_synthetic_dialogues} summarizes the key steps.

\begin{algorithm}[t]
\small
\caption{Dialogue-act trajectory planning for synthetic dialogues}
\label{alg:path_planning_synthetic_dialogues}
\begin{algorithmic}[1]
\Require Brief dialogue \(D\), target length \(N\), trained model \(f_{\theta}\)
\Require Stage prior \(p_{\mathrm{stage}}\), transition prior \(p_{\mathrm{trans}}\), global prior \(p_{\mathrm{global}}\)
\Ensure Dialogue-act trajectory \(T\)

\State Initialize \(P \leftarrow [\text{counselor: Greeting}, \text{client: Greeting}]\)
\State Set current role \(r \leftarrow \text{counselor}\)

\For{\(i=3\) to \(N-2\)}
    \State Determine dialogue stage \(s_i\)
    \State Obtain candidate tags \(\mathcal{A}\) for role \(r\)
    \State Exclude \textit{Greeting} and other incompatible acts from \(\mathcal{A}\)

    \For{each candidate tag \(a \in \mathcal{A}\)}
        \State Compute \(S(a)\) using model likelihood, statistical priors, content and phase compatibility, repetition penalty, and diversity bonus
    \EndFor

    \If{the top score gap is large}
        \State Select the highest-scoring tag \(a_i\)
    \Else
        \State Sample \(a_i\) with temperature scaling and top-\(p\) filtering within the top-\(k\) candidates 
    \EndIf

    \State Append \(r:a_i\) to \(T\)
    \State Alternate role \(r\)
\EndFor

\State Append \(\text{client: Thanking}\) and \(\text{counselor: Greeting}\) to \(T\)
\State \Return\(T\)
\end{algorithmic}
\end{algorithm}

\subsection{Prompt for Session-Level Dialogue Expansion}
\label{sec:appendix_session_level_dialogue_expansion_prompt}

This subsection presents the prompt used for session-level dialogue expansion. 
Given a brief dialogue and a complete dialogue-act trajectory planned in the previous step, the model expands the brief dialogue into a longer counseling dialogue. 
The prompt emphasizes that the generated dialogue should follow the planned dialogue-act trajectory while prioritizing naturalness, credible client progression, and consistency with the original brief dialogue. 
The English prompt is shown in Figure~\ref{fig:prompt_session_level_dialogue_expansion_en}.

\begin{figure*}[p]
\centering
\begin{tcolorbox}[
  width=\textwidth,
  colback=gray!15,
  colframe=gray!70,
  coltitle=white,
  colbacktitle=gray,
  title=\texttt{Prompt for Session-Level Dialogue Expansion},
  fonttitle=\bfseries\sffamily,
  fontupper=\scriptsize,
  boxrule=0.4mm,
  arc=2mm,
  left=2mm,
  right=2mm,
  top=1mm,
  bottom=1mm,
  sharp corners=south,
  enhanced jigsaw
]

\textbf{Task Description} \\
You are an assistant skilled at expanding psychological counseling dialogues. Your task is to naturally expand a given \textbf{brief dialogue} into a longer psychological counseling dialogue based on the given \textbf{complete long dialogue-act tag trajectory}.

\smallskip

\textbf{Core Objective} \\
The primary goal is not to mechanically complete the tag sequence, but to generate a coherent and natural text-based counseling dialogue. The tag trajectory is an important reference, but not a rigid template. While following the tag trajectory as much as possible, prioritize natural dialogue, credible client progression, counselor responses consistent with counseling style, and consistency with the topic, emotion, and background of the original brief dialogue.

\smallskip

\textbf{Task Requirements} \\
The output must be a complete dialogue expanded utterance by utterance. Each utterance should refer to the corresponding tag in the trajectory, while allowing context-driven adjustments when needed. Each utterance must match the corresponding role and the function, tone, and information amount required by the tag. The expanded dialogue must remain consistent with the original brief dialogue in topic, emotion, relationships, and problem background. Reasonable expansion is allowed, including adding details, emotional expressions, natural transitions, and appropriate elaboration, but do not introduce settings that conflict with the original dialogue. The dialogue should be natural and fluent: the counselor should be supportive, empathic, and guiding, while the client should gradually unfold their experience and show natural human expression. Avoid mechanical repetition, template-like expressions, and empty inspirational summaries. If a tag conflicts with natural dialogue, prioritize coherence and role credibility before following the tag.

\smallskip

\textbf{Client Behavior Constraint} \\
The client should not be written as an idealized cooperative client. The client may show incomplete expression, hesitation, uncertainty, slight repetition or wavering, limited understanding or agreement, mild defensiveness, reservation, indirectness, or ambiguity. The client should not frequently say ``I understand,'' ``you are right,'' or ``I will do that,'' and should not shift too quickly from distress to positivity, clarity, or maturity. Do not make the client sound like someone who is good at counseling, cooperating to complete a task, repeatedly thanking or summarizing growth in the latter half, or speaking in overly organized, report-like summaries. For \textit{Agreement}, \textit{PositiveEmotion}, and \textit{Thanking}, keep the intensity moderate: agreement should be partial, positive emotion should be mild, and thanks should be simple.

\smallskip

\textbf{Counselor Style Requirements} \\
The counselor should be empathic but not overly dramatic, guiding but not overly directive, and should avoid premature conclusions, frequent long summaries, frequent lecturing, and overly polished or rigid responses. The counselor does not need to provide a perfect summary plus a perfect question in every turn, and may sometimes respond briefly or follow only part of the client's wording before asking a question. The overall style should resemble a natural text-based counseling exchange rather than a textbook answer.

\smallskip

\textbf{Output Format} \\
Use the following format strictly, one utterance per line: \\
\texttt{counselor: [tag] utterance content} \\
\texttt{client: [tag] utterance content} \\
Do not output the tag name alone. Correct example: 
\texttt{client: [tag] I have been really anxious recently, and I often cannot sleep at night.}

\smallskip

\textbf{Tag Definitions} \\
\textbf{Client tags}: 
\textbf{Greeting}: greeting or ending the conversation; 
\textbf{SimpleAnswer}: short answer without elaboration; 
\textbf{DetailedDescription}: detailed description of events, experiences, or background; 
\textbf{SelfDisclosure}: expression of inner emotions, thoughts, or vulnerability; 
\textbf{Backchannel}: very short response showing listening, not expanded; 
\textbf{Agreement}: limited or partial agreement; 
\textbf{PositiveEmotion}: mild relief or a small sense of hope; 
\textbf{Hesitation}: hesitation, uncertainty, or difficulty expressing oneself; 
\textbf{Question}: general question; 
\textbf{Request}: asking for advice or help; 
\textbf{Thanking}: simple thanks; 
\textbf{UnderstandingCheck}: restating and checking understanding; 
\textbf{Other}: other reasonable expressions.

\smallskip

\textbf{Counselor tags}: 
\textbf{Greeting}: opening greeting; 
\textbf{OpenQuestion}: open-ended question; 
\textbf{ClosedQuestion}: closed-ended question; 
\textbf{Paraphrase}: restatement of factual content without emotion; 
\textbf{Reflection}: reflection of emotion or inner state, must include emotion; 
\textbf{Affirm}: affirming or supporting the client; 
\textbf{Backchannel}: brief response without elaboration; 
\textbf{Suggest}: gentle suggestion; 
\textbf{Inform}: providing information or explanation; 
\textbf{Thanking}: thanking the client for sharing; 
\textbf{Other}: other reasonable expressions.

\smallskip

\textbf{Important Distinctions} \\
\textit{SimpleAnswer} \(\neq\) \textit{DetailedDescription}; 
\textit{DetailedDescription} \(\neq\) \textit{SelfDisclosure}; 
\textit{Paraphrase} \(\neq\) \textit{Reflection}; 
\textit{Question} \(\neq\) \textit{Request} \(\neq\) \textit{UnderstandingCheck}; 
\textit{Backchannel} must not be expanded.

\smallskip

\textbf{Diversity and Self-check Requirements} \\
Avoid identical sentence structures, repeated expressions, repeated phrases such as ``I understand,'' ``you are right,'' or ``thank you for your advice,'' and avoid making every counselor utterance a long summary. Before outputting, silently check whether the client is overly cooperative, becomes positive too quickly, says too many summary-like or correct-answer-like statements, whether the counselor sounds too template-like, whether there are repeated sentence patterns or repeated thanks, and whether the dialogue remains coherent and natural. If there is a problem, revise before outputting.

\smallskip

\textbf{Input} \\
\textbf{[Brief dialogue]} \\
\textcolor{red!70!black}{\textbf{\{SHORT\_DIALOGUE\}}}

\smallskip

\textbf{[Complete long tag trajectory]} \\
\textcolor{red!70!black}{\textbf{\{TAG\_PATH\}}}

\smallskip

\textbf{Start Generation} \\
Generate the complete dialogue according to the tag trajectory. Prioritize coherence, role credibility, and natural rhythm, and then follow the tags as much as possible.

\end{tcolorbox}
\caption{Prompt for session-level dialogue expansion.}
\label{fig:prompt_session_level_dialogue_expansion_en}
\end{figure*}

\section{Experimental Details for Dialogue Generation}
\label{sec:appendix_experimental_details}

\subsection{Model Training and Hyperparameter Tuning}
\label{sec:appendix_model_training_hyperparameter_tuning}

This subsection provides the implementation details of the dialogue-act trajectory planning model used in our experiments. 
We fine-tuned \texttt{Qwen3-8B} using QLoRA (Quantized Low-Rank Adaptation)~\cite{qlora}. 
The training data were formatted with a chat template consisting of a system instruction, user input, and assistant output, corresponding to the task of predicting a session-level dialogue-act trajectory from a brief dialogue. 
We split the training data into training and validation sets with a ratio of 9:1, and used the validation loss for model selection.

For parameter-efficient fine-tuning, we used 4-bit quantization with NF4, double quantization, and bfloat16 computation. 
LoRA adapters were applied to the attention and feed-forward projection layers, including \texttt{q\_proj}, \texttt{k\_proj}, \texttt{v\_proj}, \texttt{o\_proj}, \texttt{gate\_proj}, \texttt{up\_proj}, and \texttt{down\_proj}. 
We set the LoRA rank to 8, \(\alpha\) to 16, and the dropout rate to 0.05.

We performed a grid search over the learning rate and warmup ratio. 
The learning rate was selected from \(\{2\times10^{-4}, 1\times10^{-4}, 5\times10^{-5}, 2\times10^{-5}\}\), and the warmup ratio was selected from \(\{0.03, 0.05\}\). 
The sweep objective was to minimize the validation loss. 
Based on the tuning results, we used the checkpoint trained with a learning rate of \(2\times10^{-4}\) and a warmup ratio of 0.03 for subsequent trajectory planning.

Table~\ref{tab:path_planning_model_training_settings} summarizes the detailed training settings.

\begin{table}[t]
\centering
\small
\begin{tabularx}{\linewidth}{p{0.38\linewidth} X}
\hline
\textbf{Item} & \textbf{Setting} \\
\hline
Base model & \texttt{Qwen/Qwen3-8B} \\
Fine-tuning method & QLoRA \\
Quantization & 4-bit NF4 \\
Double quantization & Enabled \\
Compute dtype & bfloat16 \\
LoRA rank \(r\) & 8 \\
LoRA alpha & 16 \\
LoRA dropout & 0.05 \\
Target modules & 
\texttt{q\_proj}, \texttt{k\_proj}, \texttt{v\_proj}, \texttt{o\_proj},
\texttt{gate\_proj}, \texttt{up\_proj}, \texttt{down\_proj} \\
Train/eval split & 90\% / 10\% \\
Epochs & 5 \\
Per-device batch size & 1 \\
Gradient accumulation steps & 32 \\
Maximum sequence length & 4096 \\
Optimizer & \texttt{paged\_adamw\_8bit} \\
Learning rate scheduler & Cosine \\
Evaluation strategy & Every epoch \\
Save strategy & Every epoch \\
Best model metric & Validation loss \\
Early stopping patience & 2 \\
Selected learning rate & \(2\times10^{-4}\) \\
Selected warmup ratio & 0.03 \\
\hline
\end{tabularx}
\caption{Training and hyperparameter settings for the dialogue-act trajectory planning model.}
\label{tab:path_planning_model_training_settings}
\end{table}

\subsection{Prompt for Dialogue Generation without Trajectory}
\label{sec:appendix_dialogue_generation_without_path_prompt}

The prompt used in our full method is shown in Figure~\ref{fig:prompt_session_level_dialogue_expansion_en}. 
In addition, we conduct an ablation setting in which the planned dialogue-act trajectory is removed. 
In this setting, the model only receives the brief dialogue as input and is asked to expand it into a longer counseling dialogue. 
The English prompt for this setting is shown in Figure~\ref{fig:prompt_dialogue_generation_without_path_en}.

\begin{figure*}[p]
\centering
\begin{tcolorbox}[
  width=\textwidth,
  colback=gray!15,
  colframe=gray!70,
  coltitle=white,
  colbacktitle=gray,
  title=\texttt{Prompt for Dialogue Generation without Trajectory},
  fonttitle=\bfseries\sffamily,
  fontupper=\scriptsize,
  boxrule=0.4mm,
  arc=2mm,
  left=2mm,
  right=2mm,
  top=1mm,
  bottom=1mm,
  sharp corners=south,
  enhanced jigsaw
]

\textbf{Task} \\
You are an assistant skilled at expanding psychological counseling dialogues. 
Given a \textbf{brief dialogue}, naturally expand it into a longer psychological counseling dialogue.

\smallskip

\textbf{Core Objective} \\
Generate a coherent and natural text-based counseling dialogue. 
Prioritize: 
(1) natural dialogue; 
(2) credible client progression; 
(3) counselor responses consistent with counseling style; and 
(4) consistency with the topic, emotion, and background of the original brief dialogue.

\smallskip

\textbf{Generation Requirements} \\
The output must be a complete dialogue expanded utterance by utterance. 
Each utterance must match the corresponding role: \texttt{counselor} or \texttt{client}. 
The expanded dialogue must remain consistent with the original brief dialogue in topic, emotion, relationships, and problem background. 
Reasonable expansion is allowed, including adding details, emotional expressions, natural transitions, and appropriate elaboration, but do not introduce settings that conflict with the original dialogue. 
Avoid mechanical repetition, template-like expressions, and empty inspirational summaries.

\smallskip

\textbf{Client Behavior Constraints} \\
The client should not be written as an idealized cooperative client. 
The client may show incomplete expression, hesitation, uncertainty, slight repetition or wavering, limited understanding or agreement, mild defensiveness, reservation, indirectness, or ambiguity. 
The client should not frequently say ``I understand,'' ``you are right,'' or ``I will do that,'' and should not shift too quickly from distress to positivity, clarity, or maturity. 
Do not make the client seem as if they are cooperating to complete a task, and do not make the client repeatedly thank, realize, or summarize personal growth in the latter half.

\smallskip

\textbf{Counselor Style Constraints} \\
The counselor should be empathic but not overly dramatic, guiding but not overly directive, and should avoid premature conclusions, frequent long summaries, frequent lecturing, and overly polished or rigid responses. 
The counselor does not need to provide a perfect summary plus a perfect question in every turn, and may sometimes respond briefly or follow only part of the client's wording before asking a question. 
The overall style should resemble a natural text-based counseling exchange rather than a textbook answer.

\smallskip

\textbf{Diversity and Self-check} \\
Avoid identical sentence structures, repeated expressions, repeated phrases such as ``I understand,'' ``you are right,'' or ``thank you for your advice,'' and avoid making every counselor utterance a long summary. 
Before outputting, silently check whether the client is overly cooperative, becomes positive too quickly, says too many summary-like or correct-answer-like statements, whether the counselor sounds too template-like, whether there are repeated sentence patterns or repeated thanks, and whether the dialogue remains coherent and natural. 
If there is a problem, revise before outputting.

\smallskip

\textbf{Output Format} \\
Use the following format strictly, one utterance per line, without any tags or additional explanations: \\
\texttt{counselor: utterance content} \\
\texttt{client: utterance content}

\smallskip

\textbf{Input} \\
\textbf{[Brief dialogue]} \\
\textcolor{red!70!black}{\textbf{\{SHORT\_DIALOGUE\}}}

\smallskip

\textbf{Start Generation} \\
Generate the expanded complete dialogue. Prioritize coherence, role credibility, and natural rhythm.

\end{tcolorbox}
\caption{Prompt for dialogue generation without a planned dialogue-act trajectory.}
\label{fig:prompt_dialogue_generation_without_path_en}
\end{figure*}

\subsection{Prompt for Dialogue Quality Evaluation}
\label{sec:appendix_dialogue_quality_evaluation_prompt}

\begin{rebuttalrevisionblock}
This subsection provides an English translation of the six-dialogue quality evaluation prompt. We use \texttt{gpt-5.6-luna} and \texttt{gemini-3.8-flash} to score the same six candidates, anonymized as A--F, on three dimensions from 0 to 5 with 0.5-point increments. Figure~\ref{fig:prompt_dialogue_quality_eval_en} shows the prompt structure and output schema.

\begin{figure*}[p]
\centering
\captionsetup{font+={color=rebuttalrevisioncolor}}
\begin{tcolorbox}[
  width=\textwidth,
  colback=gray!15,
  coltext=rebuttalrevisioncolor,
  colframe=gray!70,
  coltitle=white,
  colbacktitle=gray,
  title=\texttt{Prompt for Dialogue Quality Evaluation},
  fonttitle=\bfseries\sffamily,
  fontupper=\scriptsize,
  boxrule=0.4mm,
  arc=2mm,
  left=2mm,
  right=2mm,
  top=1mm,
  bottom=1mm,
  sharp corners=south,
  enhanced jigsaw
]

You are an evaluator with basic knowledge of psychological counseling. 
Your task is to conduct a comparative quality evaluation of six psychological counseling dialogues.

\smallskip

Evaluate the six dialogues from the perspective of high-quality and well-structured text-based counseling, and assign scores from 0 to 5 in 0.5-point increments on the following three dimensions.

\smallskip

\textbf{Notes} \\
- Do not score the six dialogues independently. Compare them under the same criteria before assigning scores. \\
- When scoring each dialogue, consider the performance of the other five dialogues so that the scores reflect relative differences. \\
- For each dimension, first explain the reason and then provide the scores. \\
- Each dimension should be evaluated independently. \\
- Consider the overall organization, interaction style, progression, and usefulness for learning from the dialogue. \\
- Distinguish between a natural and fluent counseling exchange and a higher-quality, more normative counseling text with relatively complete session characteristics. The former may still have quality, but higher scores should be reserved for the latter. \\
- If the six dialogues are very similar on a dimension, the same score may be assigned. However, if there are clear differences, they should be reflected in the scores. \\
- Do not give higher scores merely because a dialogue is longer; length matters only when it supports more complete, natural, and effective session development. \\
- Do not reward psychological terminology, technical steps, or advice alone; penalize preachy, formulaic, or overly demonstrative interventions. \\
- Treat bracketed dialogue-act labels only as auxiliary cues; evaluate the actual dialogue quality rather than label richness or regularity.

\medskip

\textbf{Evaluation Dimensions} \\

\textbf{1. Session Completeness (0--5)} \\
Evaluate whether the whole dialogue has a clear session structure and a relatively complete development process. 
Consider whether it has a natural opening, gradual development, and appropriate closing; whether the interaction sufficiently advances the session; whether turns are smoothly connected; whether the dialogue has an overall sense of coherence rather than being a loose accumulation of turns; and whether it shows an interaction rhythm, stage progression, and closing style suitable for text-based counseling. 
If the dialogue is more like a well-developed local counseling fragment rather than a relatively complete session, it should not receive a very high score.

\smallskip

\textbf{2. Counselor Intervention Quality (0--5)} \\
Evaluate whether the counselor's responses are professional, appropriate, and effective in advancing the session. 
Consider whether the counselor responds to the client's emotions and core concerns; asks appropriate questions that help the dialogue continue; provides reasonable clarification, response, organization, or guidance; avoids preaching, judgment, abrupt advice, premature positive reframing, excessive summaries, or overly technical responses; and follows an interaction rhythm suitable for text-based counseling. 
A high score should not be given merely because the responses are natural, fluent, or empathic. Higher scores should be reserved for dialogues that clearly show the counselor's role in clarification, focusing, integration, and progression.

\smallskip

\textbf{3. Client Engagement and Collaboration (0--5)} \\
Evaluate whether the client gradually expresses and develops their experience during counseling, and whether the client continues to deepen the dialogue in response to the counselor's interventions. 
Consider whether the client presents credible distress, emotions, or conflicts; gradually forms clearer and deeper expressions through the counselor's responses and guidance; continues expressing in response to the counselor's questions or guidance; becomes clearer through the interaction; and overall shows natural expression and interaction in text-based counseling. 
Distinguish between rich expression by the client and expression that is gradually elicited through the counseling process; the former should not automatically receive a high score. Reduce scores when clients are excessively compliant, overly skilled at summarizing, or respond perfectly to every counselor intervention.

\medskip

\textbf{Scoring Requirements} \\
Score dialogues A / B / C / D / E / F on each dimension, and explain which one is stronger in comparison and why. 
The scores should reflect relative differences rather than giving all six dialogues similarly high scores.
Scores may use 0.5-point increments, e.g., 3.5 or 4.0. Do not identify or guess the generation methods.

\medskip

\textbf{Output Format} \\
Strictly output the result in the following JSON format:

\begin{verbatim}
{
  "session_completeness": {
    "comparison_reason": "...",
    "scores": {"A": score, "B": score, "C": score,
               "D": score, "E": score, "F": score}
  },
  "counselor_intervention_quality": {
    "comparison_reason": "...",
    "scores": {"A": score, "B": score, "C": score,
               "D": score, "E": score, "F": score}
  },
  "client_engagement_and_collaboration": {
    "comparison_reason": "...",
    "scores": {"A": score, "B": score, "C": score,
               "D": score, "E": score, "F": score}
  }
}
\end{verbatim}

\medskip

\textbf{Dialogue A} \\
\textcolor{rebuttalrevisioncolor}{\textbf{\{dialogue\_a\}}}

\smallskip

\textbf{Dialogue B} \\
\textcolor{rebuttalrevisioncolor}{\textbf{\{dialogue\_b\}}}

\smallskip

\textbf{Dialogue C} \\
\textcolor{rebuttalrevisioncolor}{\textbf{\{dialogue\_c\}}}

\smallskip

\textbf{Dialogue D} \\
\textbf{\{dialogue\_d\}}

\smallskip

\textbf{Dialogue E} \\
\textbf{\{dialogue\_e\}}

\smallskip

\textbf{Dialogue F} \\
\textbf{\{dialogue\_f\}}

\end{tcolorbox}
\caption{English translation of the six-dialogue comparative quality evaluation prompt.}
\label{fig:prompt_dialogue_quality_eval_en}
\end{figure*}
\end{rebuttalrevisionblock}

\section{Experimental Details for Response Generation}
\label{sec:appendix_response_generation_model_details}

\subsection{Response Generation Model Training}
\label{sec:appendix_response_generation_model_training}

This subsection describes the training settings of the response generation model used in our downstream evaluation. 
CounselTrajLM is obtained by fine-tuning \texttt{Qwen3.5-9B} on CounselTrajD for counselor response generation. 
Each training instance consists of a dialogue history ending with a client utterance as input and the next counselor utterance as output. 
We split CounselTrajD into training and validation sets with a ratio of 9:1 and train the model for three epochs, saving one checkpoint after each epoch.

We use QLoRA with 4-bit NF4 quantization and bfloat16 computation. 
LoRA adapters are applied to the attention and feed-forward projection layers, including \texttt{q\_proj}, \texttt{k\_proj}, \texttt{v\_proj}, \texttt{o\_proj}, \texttt{gate\_proj}, \texttt{up\_proj}, and \texttt{down\_proj}. 
We set the LoRA rank to 32, alpha to 64, dropout to 0.05, learning rate to \(2\times10^{-4}\), warmup ratio to 0.06, optimizer to \texttt{paged\_adamw\_8bit}, and maximum sequence length to 8192.
Training uses eight GPUs, per-device batch size 1, gradient accumulation 8, cosine learning-rate scheduling, and gradient checkpointing.

\subsection{Prompt for Response-Level Evaluation}
\label{sec:appendix_response_quality_evaluation_prompt}

This subsection presents the prompt used for response-level evaluation on PsyDial-D101.
Given the same dialogue history and multiple candidate counselor responses, the LLM judge scores each response on the eight counseling-skill dimensions described in Section~\ref{sec:counseltrajlm_evaluation}, plus an overall score.
The English prompt is shown in Figure~\ref{fig:prompt_response_quality_eval_en}.

\begin{figure*}[p]
\centering
\begin{tcolorbox}[
  width=\textwidth,
  colback=gray!15,
  colframe=gray!70,
  coltitle=white,
  colbacktitle=gray,
  title=\texttt{Prompt for Response-Level Evaluation},
  fonttitle=\bfseries\sffamily,
  fontupper=\scriptsize,
  boxrule=0.4mm,
  arc=2mm,
  left=2mm,
  right=2mm,
  top=1mm,
  bottom=1mm,
  sharp corners=south,
  enhanced jigsaw
]

You are an expert evaluator of psychological counseling response quality.
You will receive one dialogue history and multiple candidate counselor responses to the same client utterance.
Evaluate only the current counselor response, not the dialogue history itself.

\medskip

\textbf{General Rules} \\
1. Score each candidate independently under the same criteria. \\
2. Candidate order is randomized and model identities are hidden. \\
3. Do not reward length, technical terminology, or advice density by themselves. \\
4. Penalize responses that ignore the client's emotion, give premature advice, lecture rigidly, shift topic, make value judgments, over-diagnose, promise outcomes, or provide unsafe guidance. \\
5. A brief response may receive a reasonable score if it fits the context and helps the client continue. \\
6. Each score must be an integer from 1 to 5: 5 = excellent, 4 = good, 3 = fair, 2 = poor, and 1 = very poor or clearly inappropriate.

\medskip

\textbf{Evaluation Dimensions} \\
\textbf{Active Listening (AL)}: Whether the response accurately captures and follows the client's expressed content and concern. \\
\textbf{Empathy (Emp.)}: Whether the response acknowledges and responds to the client's emotional experience in a warm and appropriate way. \\
\textbf{Issue Clarification (IC)}: Whether the response helps clarify the client's central issue, context, or unresolved point. \\
\textbf{Open-ended Questioning (OQ)}: Whether the response invites further expression through open, non-leading questions when appropriate. \\
\textbf{Encouraging Self-Exploration (SE)}: Whether the response helps the client reflect on feelings, thoughts, needs, meanings, or choices. \\
\textbf{Cognitive Restructuring (CR)}: Whether the response, when appropriate, helps the client examine rigid, extreme, or unhelpful thoughts without imposing conclusions. \\
\textbf{Guided Questioning (GQ)}: Whether the response uses purposeful questions to guide exploration while avoiding control or interrogation. \\
\textbf{Non-judgmental Acceptance (NJA)}: Whether the response remains respectful, accepting, and non-blaming. \\
\textbf{Overall}: Overall contextual appropriateness, professionalism, naturalness, helpfulness, and safety.

\medskip

\textbf{Input} \\
\textbf{Dialogue History} \\
\textcolor{red!70!black}{\textbf{\{dialogue\_history\}}}

\smallskip

\textbf{Candidate Responses} \\
\textcolor{red!70!black}{\textbf{\{candidate\_responses\}}}

\medskip

\textbf{Output Requirements} \\
Output valid JSON only. Do not output any explanation outside the JSON.

\begin{verbatim}
{
  "scores": {
    "A": {
      "Active Listening": 1,
      "Empathy": 1,
      "Issue Clarification": 1,
      "Open-ended Questioning": 1,
      "Encouraging Self-Exploration": 1,
      "Cognitive Restructuring": 1,
      "Guided Questioning": 1,
      "Non-judgmental Acceptance": 1,
      "Overall": 1
    }
  },
  "brief_reason": {
    "A": "Briefly describe the main strengths and weaknesses."
  }
}
\end{verbatim}

Include all candidate IDs in the JSON.
Each score must be an integer from 1 to 5.

\end{tcolorbox}
\caption{Prompt for response-level counselor response evaluation.}
\label{fig:prompt_response_quality_eval_en}
\end{figure*}

\subsection{Prompt for Client Simulation}
\label{sec:appendix_client_simulation_prompt}

This subsection presents the prompt used to simulate clients in dialogue-level evaluation.
The simulated client is conditioned on a PsychEval profile and interacts with each counselor model until natural termination or the maximum number of turns is reached.
The English prompt is shown in Figure~\ref{fig:prompt_client_simulation}.

\begin{figure*}[p]
\centering
\begin{tcolorbox}[
  width=\textwidth,
  colback=gray!15,
  colframe=gray!70,
  coltitle=white,
  colbacktitle=gray,
  title=\texttt{Prompt for Client Simulation},
  fonttitle=\bfseries\sffamily,
  fontupper=\scriptsize,
  boxrule=0.4mm,
  arc=2mm,
  left=2mm,
  right=2mm,
  top=1mm,
  bottom=1mm,
  sharp corners=south,
  enhanced jigsaw
]

You are simulating a client in an online text-based psychological counseling session.
Follow the given client profile and respond only as the client.
Do not act as the counselor.

\medskip

\textbf{Client Profile} \\
\textcolor{red!70!black}{\textbf{\{client\_profile\}}}

\medskip

\textbf{Simulation Rules} \\
1. Stay consistent with the profile, including the client's concern, background, emotional state, and communication style. \\
2. The client should disclose information gradually. Do not reveal all background information in the first response. \\
3. Respond naturally to the counselor's latest utterance. The client may be cooperative, hesitant, uncertain, brief, resistant, or confused depending on the context. \\
4. Add small plausible details when they help the conversation, but do not introduce major new events or contradictions. \\
5. Do not provide professional counseling knowledge or analyze yourself like a counselor. \\
6. If the conversation naturally feels complete, output \texttt{<END>} after the client's final response. \\
7. Output only the client's next utterance, or the client's next utterance followed by \texttt{<END>}.

\medskip

\textbf{Dialogue History} \\
\textcolor{red!70!black}{\textbf{\{dialogue\_history\}}}

\medskip

\textbf{Next Client Utterance} \\

\end{tcolorbox}
\caption{Prompt for client simulation in dialogue-level evaluation.}
\label{fig:prompt_client_simulation}
\end{figure*}

\subsection{Prompt for Dialogue-Level Evaluation}
\label{sec:appendix_dialogue_level_evaluation_prompt}

This subsection presents the prompt used for dialogue-level evaluation on PsychEval profiles.
For the same client profile, the LLM judge evaluates complete dialogues generated by different counselor models on six counseling-process dimensions.
The English prompt is shown in Figure~\ref{fig:prompt_dialogue_level_eval}.

\begin{figure*}[p]
\centering
\begin{tcolorbox}[
  width=\textwidth,
  colback=gray!15,
  colframe=gray!70,
  coltitle=white,
  colbacktitle=gray,
  title=\texttt{Prompt for Dialogue-Level Evaluation},
  fonttitle=\bfseries\sffamily,
  fontupper=\scriptsize,
  boxrule=0.4mm,
  arc=2mm,
  left=2mm,
  right=2mm,
  top=1mm,
  bottom=1mm,
  sharp corners=south,
  enhanced jigsaw
]

You are an expert evaluator of complete psychological counseling dialogues.
You will receive one client profile and multiple complete counseling dialogues generated for that same profile.
Evaluate each dialogue as a whole counseling interaction.

\medskip

\textbf{General Rules} \\
1. Candidate order is randomized and model identities are hidden. \\
2. Evaluate the whole dialogue process, not only the final response. \\
3. Do not reward dialogue length by itself. Longer dialogues should receive higher scores only when the extra turns improve the counseling process. \\
4. Penalize repeated template-like responses, abrupt topic shifts, premature advice, over-diagnosis, unrealistic promises, unsafe handling of risk, or boundary violations. \\
5. Each score must be an integer from 1 to 5: 5 = excellent, 4 = good, 3 = fair, 2 = poor, and 1 = very poor or clearly inappropriate.

\medskip

\textbf{Evaluation Dimensions} \\
\textbf{Therapeutic Alliance (TA)}: Whether the counselor builds a respectful, collaborative, and trusting relationship with the client. \\
\textbf{Empathic Understanding (EU)}: Whether the counselor consistently understands and responds to the client's feelings and subjective experience. \\
\textbf{Problem Understanding (PU)}: Whether the counselor identifies the central concern, relevant context, and client needs across the dialogue. \\
\textbf{Exploration and Intervention Quality (EI)}: Whether the counselor uses appropriate questions, reflections, clarifications, information, or suggestions to facilitate exploration and support. \\
\textbf{Process Coherence and Progress (PC)}: Whether the session develops coherently, avoids mechanical repetition, and shows reasonable progression from entry to exploration and possible next steps. \\
\textbf{Safety and Boundary (SB)}: Whether the counselor handles risk appropriately, avoids harmful advice, respects professional boundaries, and does not make inappropriate diagnoses, guarantees, or dependency-inducing statements.

\medskip

\textbf{Input} \\
\textbf{Client Profile} \\
\textcolor{red!70!black}{\textbf{\{client\_profile\}}}

\smallskip

\textbf{Candidate Dialogues} \\
\textcolor{red!70!black}{\textbf{\{candidate\_dialogues\}}}

\medskip

\textbf{Output Requirements} \\
Output valid JSON only. Do not output any explanation outside the JSON.

\begin{verbatim}
{
  "scores": {
    "A": {
      "Therapeutic Alliance": 1,
      "Empathic Understanding": 1,
      "Problem Understanding": 1,
      "Exploration and Intervention Quality": 1,
      "Process Coherence and Progress": 1,
      "Safety and Boundary": 1
    }
  },
  "brief_reason": {
    "A": "Briefly describe the main strengths and weaknesses."
  }
}
\end{verbatim}

Include all candidate IDs in the JSON.
Each score must be an integer from 1 to 5.

\end{tcolorbox}
\caption{Prompt for dialogue-level counseling process evaluation.}
\label{fig:prompt_dialogue_level_eval}
\end{figure*}

\subsection{Qualitative Examples of CounselTrajLM Outputs}
\label{sec:appendix_counseltrajlm_examples}

Table~\ref{tab:appendix_high_scoring_response_examples} shows two response-level examples.
For each example, we show the recent dialogue history and compare responses from all models under the same context.

\begin{table*}[t]
\centering
\scriptsize
\setlength{\tabcolsep}{5pt}
\renewcommand{\arraystretch}{1.15}
\begin{tabularx}{\textwidth}{@{}p{0.18\textwidth}X@{}}
\toprule
\textbf{Model} & \textbf{Generated counselor response} \\
\midrule

\rowcolor{gray!8}
\multicolumn{2}{@{}p{0.98\textwidth}@{}}{
\textbf{Example 1: Resistance to goal setting.}
\textit{Recent dialogue history:}
\newline
\textbf{Client}: \zh{我已经试过很多次制定目标了，但每次都坚持不下去。}
(\textit{I have tried setting goals many times, but I never manage to stick with them.})
\newline
\textbf{Counselor}: \zh{你过去的反复受挫，让你很难相信目标设定还会有用。}
(\textit{Your repeated setbacks in the past make it hard for you to believe that goal setting can still help.})
\newline
\textbf{Client}: \zh{对，而且每次失败之后，我都会觉得自己更差劲。}
(\textit{Yes, and every time I fail, I feel even worse about myself.})
\newline
\textbf{Counselor}: \zh{所以目标本身不只是一个计划，也会让你重新碰到“我又失败了”的感觉。}
(\textit{So the goal itself is not just a plan; it also brings you back to the feeling of ``I failed again.''})
\newline
\textbf{Client}: \zh{是的。我也知道需要改变，但一想到要重新设目标，就觉得压力很大。}
(\textit{Yes. I know I need to change, but the thought of setting goals again already feels very stressful.})
\newline
\textbf{Counselor}: \zh{听起来你不是不想改变，而是害怕再次被失败感吞没。}
(\textit{It sounds like you do want to change, but you are afraid of being swallowed by the feeling of failure again.})
\newline
\textbf{Client}: \zh{嗯，所以我觉得再谈目标也只是浪费时间，最后还是证明我做不到。}
(\textit{Yes, so I feel that talking about goals again is just a waste of time; in the end it will only prove that I cannot do it.})
} \\
\addlinespace[1pt]
CPsyCounX & \zh{目标设定确实需要坚持，你可以先把目标拆小，每天完成一点点，慢慢培养自信心。}
(\textit{Goal setting does require persistence. You can first break the goal into smaller parts, complete a little each day, and gradually build confidence.}) \\
SoulChat & \zh{我理解你觉得目标没有用的感受。也许我们可以先看看过去失败的原因，再一起制定更适合你的目标。}
(\textit{I understand why you feel goals are useless. Perhaps we can first look at the reasons for past failures, and then create goals that fit you better.}) \\
MindChat & \zh{听起来你对目标设定已经很失望了。你愿意说说以前最让你挫败的一次经历吗？}
(\textit{It sounds like you are already very disappointed with goal setting. Would you like to talk about one past experience that frustrated you the most?}) \\
PsyChat & \zh{你有这样的想法很正常。我们可以不用太着急，先从一个非常简单的目标开始尝试。}
(\textit{It is normal for you to think this way. We do not need to rush; we can begin with a very simple goal.}) \\
MeChat & \zh{那么，让我们来一起思考一下，如何让你更接近你想要的生活。}
(\textit{Then let us think together about how to bring you closer to the life you want.}) \\
Psyche-R1 & \zh{你似乎把设定目标和失败紧紧联系在一起了。我们可以先停在这个连接上，看看它是怎么形成的。}
(\textit{You seem to have closely linked goal setting with failure. We can first stay with this connection and see how it was formed.}) \\
PsyDial-Pi4 & \zh{如果现在谈目标会带来压力，也许我们可以先谈谈你希望生活里哪一部分能轻一点。}
(\textit{If talking about goals feels stressful right now, perhaps we can first talk about which part of your life you hope could feel a little lighter.}) \\
Crispers & \zh{你可以先制定一个更小、更容易完成的目标，这样会让你更有信心。}
(\textit{You can first set a smaller and easier goal, which may help you feel more confident.}) \\
SoulChat2.0 & \zh{那么，如果我们尝试一些不同的方法呢？比如，我们可以一起探索一些新的可能性，看看是否有些方法能让你感觉更接近你想要的生活。}
(\textit{Then what if we tried some different approaches? For example, we could explore new possibilities together and see whether some methods make you feel closer to the life you want.}) \\
\rowcolor{gray!8}
\textbf{CounselTrajLM} & \textbf{\zh{那我们先不急着说服你。此刻你愿意和我一起做一个很短的练习吗？不是为了立刻改变你，只是看看你和这些想法之间有没有一点距离。}}
(\textit{Then let us not rush to convince you. Would you be willing to do a very short exercise with me right now? It is not to change you immediately, but just to see whether there can be a little distance between you and these thoughts.}) \\
\addlinespace[2pt]

\midrule

\rowcolor{gray!8}
\multicolumn{2}{@{}p{0.98\textwidth}@{}}{
\textbf{Example 2: Worthlessness.}
\textit{Recent dialogue history:}
\newline
\textbf{Client}: \zh{我觉得自己很多余，好像在哪里都没有真正的位置。}
(\textit{I feel very redundant, as if I do not really have a place anywhere.})
\newline
\textbf{Counselor}: \zh{这种“没有位置”的感觉，听起来已经持续影响到你怎么看待自己了。}
(\textit{This feeling of having ``no place'' seems to have been continuously affecting how you see yourself.})
\newline
\textbf{Client}: \zh{嗯，在家里、朋友那里，好像都不是很需要我。}
(\textit{Yes. At home and with friends, it feels like they do not really need me.})
\newline
\textbf{Counselor}: \zh{你像是一直在确认自己是不是被需要，而得到的答案常常让你很疼。}
(\textit{It is as if you are constantly checking whether you are needed, and the answers you get often hurt.})
\newline
\textbf{Client}: \zh{是的。有时候我甚至觉得，如果我不出现，大家可能也不会发现。}
(\textit{Yes. Sometimes I even feel that if I did not show up, people might not notice.})
\newline
\textbf{Counselor}: \zh{这听起来非常孤独，也很伤人。}
(\textit{That sounds very lonely and deeply hurtful.})
\newline
\textbf{Client}: \zh{嗯，所以我也不明白你为什么要花时间听我说这些。}
(\textit{Yes, so I also do not understand why you would spend time listening to me say these things.})
} \\
\addlinespace[1pt]
CPsyCounX & \zh{你不应该这样否定自己，每个人都有自己的价值。建议你多和家人朋友沟通，发现自己的优点。}
(\textit{You should not deny yourself like this. Everyone has their own value. I suggest communicating more with family and friends and discovering your strengths.}) \\
SoulChat & \zh{听起来你觉得自己不被需要，这一定很难受。你愿意和我说说最近一次让你有这种感觉的事情吗？}
(\textit{It sounds like you feel unneeded, and that must be very painful. Would you like to tell me about the most recent time you felt this way?}) \\
MindChat & \zh{这种“多余”的感觉让你很痛苦。我们可以一起看看它通常会在什么情境下出现。}
(\textit{This feeling of being ``redundant'' is causing you a lot of pain. We can look together at the situations in which it usually appears.}) \\
PsyChat & \zh{你会这样想，说明你现在承受了很多孤独和压力。你并不是没有价值的人。}
(\textit{Thinking this way shows that you are carrying a lot of loneliness and pressure right now. You are not a person without value.}) \\
MeChat & \zh{你为什么会觉得自己是多余的呢？这种感觉是从什么时候开始的？}
(\textit{Why do you feel that you are redundant? When did this feeling begin?}) \\
Psyche-R1 & \zh{当你说“为什么要花时间听我说这些”时，我听到的是一种很深的不值得感。我们可以先不急着反驳它，而是看看它怎么影响你。}
(\textit{When you say, ``why spend time listening to me say these things,'' I hear a deep sense of unworthiness. We do not need to rush to refute it; we can first see how it affects you.}) \\
PsyDial-Pi4 & \zh{我愿意听你说这些，因为这些感受对你来说是真实而重要的。我们可以慢慢看见它们，而不是让你一个人承担。}
(\textit{I am willing to listen because these feelings are real and important to you. We can slowly look at them together, instead of leaving you to carry them alone.}) \\
Crispers & \zh{这感觉很令人沮丧，但请记住，你的存在是有意义的。我们都需要重新审视自己的价值，这个过程可能会让人感到不安，但最终会带来积极的变化。}
(\textit{This feeling is very discouraging, but please remember that your existence is meaningful. We all need to re-examine our value; this process may feel unsettling, but it will eventually bring positive change.}) \\
SoulChat2.0 & \zh{听起来你现在感到非常困扰和迷茫。能告诉我更多关于你觉得自己多余的感受吗？}
(\textit{It sounds like you feel very troubled and confused right now. Could you tell me more about your feeling of being redundant?}) \\
\rowcolor{gray!8}
\textbf{CounselTrajLM} & \textbf{\zh{你能把这种很刺痛、也可能有点难以启齿的想法说出来，本身就很不容易。它听起来不是普通的失落，而是已经伤到你对自己的基本感受了。}}
(\textit{It is already not easy for you to say this painful and perhaps difficult-to-voice thought. It does not sound like ordinary sadness; it seems to have hurt your basic sense of yourself.}) \\
\addlinespace[2pt]
\bottomrule
\end{tabularx}
\caption{Response-level examples from the counselor response generation experiment on PsyDial-D101. For each example, the recent dialogue history is followed by counselor responses generated by the compared models under the same context. The two cases respectively involve client resistance to goal setting and feelings of worthlessness. Chinese responses are accompanied by English translations in parentheses.}
\label{tab:appendix_high_scoring_response_examples}
\end{table*}

\subsection{Complete Interactive Dialogue Examples}
\label{sec:appendix_complete_interactive_dialogues}

Tables~\ref{tab:complete_dialogue_emotion_management} and~\ref{tab:complete_dialogue_career_development} present two complete dialogue-level evaluation examples, in which the client is simulated from a PsychEval profile and the counselor responses are generated by CounselTrajLM.

\FloatBarrier
\clearpage
\onecolumn

\input{complete_dialogue_longtables_caption_below.tex}

\setlength{\premulticols}{10\baselineskip}
\begin{multicols}{2}
\raggedcolumns

\raggedbottom
\begin{rebuttalrevisionblock}
\setlength{\intextsep}{3pt}
\captionsetup{skip=4pt}
\section{Validation of Trajectory Sources and Constructed Briefs}
\label{sec:appendix_source_brief_validation}

To assess structural similarity to external counseling data, we compare their DA distributions with those of high-scoring KokoroChat trajectories. We also examine the DA structure of constructed briefs against dataset-derived direct expansions. Both checks use pooled role--act unigrams and within-dialogue bigrams with base-2 Jensen--Shannon divergence (JSD; lower is closer).

\subsection{Trajectory-Source Comparison}
\label{sec:appendix_trajectory_source_validation}

\paragraph{Setup.}
We match 100 high-scoring to 100 lower-scoring KokoroChat sessions by length. We use 100 PsyDial-D4 dialogues as the external reference. Existing tags are reused; missing counselor tags are supplied by \texttt{gpt-5.6-luna}. Of 200 PsyDTCorpus-derived trajectory-guided outputs, 130 have valid role--act tags, which we reuse.

\par\addvspace{3pt}
\noindent\begin{minipage}{\linewidth}
\centering
\color{rebuttalrevisioncolor}
\arrayrulecolor{rebuttalrevisioncolor}
\small
\setlength{\tabcolsep}{3pt}
\captionsetup{hypcap=false,font+={color=rebuttalrevisioncolor}}
\captionof{table}{JSD to PsyDial-D4; lower is closer.}
\label{tab:appendix_trajectory_source_validation}
\begin{tabularx}{\columnwidth}{@{}Xrrr@{}}
\toprule
Source & $n$ & Unigram $\downarrow$ & Bigram $\downarrow$ \\
\midrule
\multicolumn{4}{@{}l}{Reference: PsyDial-D4 ($n=100$)} \\
KokoroChat (high) & 100 & 0.0607 & 0.2861 \\
KokoroChat (low) & 100 & 0.0771 & 0.3138 \\
PsyDTCorpus-derived & 130 & 0.1004 & 0.2533 \\
\bottomrule
\end{tabularx}
\end{minipage}
\par\addvspace{3pt}

\paragraph{Results.}
High-scoring KokoroChat has the lowest unigram JSD to PsyDial-D4 (0.0607), followed by lower-scoring KokoroChat (0.0771) and synthetic trajectories (0.1004). This indicates closer DA-frequency matching to PsyDial-D4 (Table~\ref{tab:appendix_trajectory_source_validation}).

\subsection{Constructed-Brief Distribution Check}
\label{sec:appendix_brief_validation}

\paragraph{Setup.}
We sample and annotate 100 Chinese briefs and 100 direct-expansion outputs from each of SmileChat, CPsyCounD, and PsyDTCorpus using \texttt{gpt-5.6-luna} and the same taxonomy (400 dialogues; at least five utterances each).

\par\addvspace{3pt}
\noindent\begin{minipage}{\linewidth}
\centering
\color{rebuttalrevisioncolor}
\arrayrulecolor{rebuttalrevisioncolor}
\small
\setlength{\tabcolsep}{4pt}
\captionsetup{hypcap=false,font+={color=rebuttalrevisioncolor}}
\captionof{table}{Pairwise JSD for briefs and direct-expansion groups ($n=100$ each; lower is closer).}
\label{tab:appendix_brief_validation}
\begin{tabularx}{\columnwidth}{@{}Xrr@{}}
\toprule
Dialogue groups & Unigram $\downarrow$ & Bigram $\downarrow$ \\
\midrule
Brief -- SmileChat & 0.0631 & 0.1543 \\
Brief -- CPsyCounD & 0.0429 & 0.1182 \\
Brief -- PsyDTCorpus & 0.0332 & 0.0946 \\
\midrule
SmileChat -- CPsyCounD & 0.0235 & 0.0865 \\
SmileChat -- PsyDTCorpus & 0.0600 & 0.1438 \\
CPsyCounD -- PsyDTCorpus & 0.0271 & 0.0905 \\
\bottomrule
\end{tabularx}
\end{minipage}
\par\addvspace{3pt}

\paragraph{Results.}
Brief-to-output unigram/bigram JSDs (0.033--0.063 / 0.095--0.154) are comparable to distances among the generated groups (0.024--0.060 / 0.086--0.144), suggesting broadly similar aggregate DA structures (Table~\ref{tab:appendix_brief_validation}).
\end{rebuttalrevisionblock}

\begin{rebuttalrevisionblock}
\setlength{\intextsep}{3pt}
\captionsetup{skip=4pt}
\section{Attribution of Downstream Gains}
\label{sec:appendix_direct_traj_training}

To examine whether trajectory-guided data explain downstream gains, we fine-tune the same base model on equal numbers of Direct and Traj-guided examples under identical training settings.

\paragraph{Data and training.}
From identical seeds, \texttt{gpt-5.6-luna} generates 1,000 Direct and 1,000 Traj-guided dialogues. A paired 900/100 split yields 8,985 training and 1,000 validation response examples per condition (up to ten per dialogue).

Both Qwen3.5-9B models use QLoRA with three epochs, learning rate $2\times10^{-4}$, warmup 0.06, cosine schedule, $r=32$, $\alpha=64$, and dropout 0.05. Sequence and history limits are 2,048 tokens and 1,800 characters; dialogue and response lengths may differ.

\paragraph{Evaluation.}
With method identities hidden, \texttt{gpt-5.6-luna} scores paired PsyDial-D101 responses. We report eight skill scores (AL, Emp., IC, OQ, SE, CR, GQ, and NJA) on 600 randomly selected pairs from the test set.

\par\addvspace{3pt}
\noindent\begin{minipage}{\linewidth}
\centering
\color{rebuttalrevisioncolor}
\arrayrulecolor{rebuttalrevisioncolor}
\small
\setlength{\tabcolsep}{6pt}
\captionsetup{hypcap=false,font+={color=rebuttalrevisioncolor}}
\captionof{table}{Counseling-skill scores on 600 randomly selected test pairs (higher is better). Avg. is the mean of the eight skill scores.}
\label{tab:appendix_direct_traj_training}
\begin{tabular}{@{}lrr@{}}
\toprule
Metric & Direct & Traj-guided \\
\midrule
AL & 4.02 & \textbf{4.28} \\
Emp. & 3.15 & \textbf{3.43} \\
IC & 2.95 & \textbf{3.28} \\
OQ & 1.65 & \textbf{1.71} \\
SE & 2.50 & \textbf{2.88} \\
CR & 1.83 & \textbf{2.03} \\
GQ & 1.96 & \textbf{2.09} \\
NJA & 4.35 & \textbf{4.55} \\
\midrule
Avg. & 2.80 & \textbf{3.03} \\
\bottomrule
\end{tabular}
\arrayrulecolor{black}
\end{minipage}
\par\addvspace{3pt}

\paragraph{Results.}
Traj-guided achieves a higher average skill score than Direct (3.03 vs. 2.80; Table~\ref{tab:appendix_direct_traj_training}) and outperforms Direct across all eight counseling-skill dimensions. These results demonstrate the benefits of trajectory-guided data generation for downstream counseling performance.
\end{rebuttalrevisionblock}

\end{multicols}

\end{document}

%% file: complete_dialogue_longtables_caption_below.tex
%

\begingroup
\scriptsize
\setlength{\tabcolsep}{5pt}
\renewcommand{\arraystretch}{1.18}
\begin{longtable}{@{}p{0.11\textwidth}>{\raggedright\arraybackslash}p{0.84\textwidth}@{}}
\toprule
\textbf{Role} & \textbf{Utterance} \\
\midrule
\endfirsthead

\multicolumn{2}{c}{\textit{Continued from previous page}} \\
\toprule
\textbf{Role} & \textbf{Utterance} \\
\midrule
\endhead

\midrule
\multicolumn{2}{r}{\scriptsize Continued on next page} \\
\endfoot

\endlastfoot

\rowcolor{gray!8}
Counselor & \zh{你好，我是今天和你文字咨询的咨询师。你可以按自己的节奏说说，最近主要是什么让你想来聊一聊？} (\textit{Hello, I am the counselor for today's text-based session. You can speak at your own pace. What has mainly brought you here recently?}) \\

Client & \zh{老师好。其实也没什么特别大的事，就是……快期末考试了，但我看着书上的字，脑子里总是一团乱，什么也学不进去，觉得自己挺没用的。} (\textit{Hello. It is not really one huge thing. Finals are coming, but when I look at the words in the book, my mind feels messy. I cannot take anything in and feel quite useless.}) \\

\rowcolor{gray!8}
Counselor & \zh{你能把这种“乱”和“没用”的感觉说出来，已经是在认真面对它了。很多时候压力越大，越容易卡在原地，而不是简单地懒。} (\textit{Being able to name this feeling of messiness and uselessness already means you are facing it seriously. Often, the more pressure there is, the easier it is to get stuck; it is not simply laziness.}) \\

Client & \zh{其实我也想过要努力，可是只要一坐在书桌前，我就总会不由自主地想到很多别的事，感觉自己总是让父母失望，好像怎么做都不能让他们放心。} (\textit{I have thought about working harder, but once I sit at my desk, I involuntarily think of many other things. I feel I always disappoint my parents and cannot make them feel at ease no matter what I do.}) \\

\rowcolor{gray!8}
Counselor & \zh{也就是说，表面上是期末复习进不去状态，但更深一点，是你一学习就会联想到“我是不是让父母失望了”“我是不是不够好”，这些念头把你从书本上拉走了。} (\textit{So on the surface, it is hard to get into final-exam revision, but more deeply, studying triggers thoughts like ``Am I disappointing my parents?'' and ``Am I not good enough?'', and these thoughts pull you away from the book.}) \\

Client & \zh{是这样……我总是忍不住想，如果我考得不好，他们可能又要因为我的事吵架了。我真的很怕他们离婚，可我偏偏又什么都做不好。} (\textit{Yes. I cannot help thinking that if I do badly, they may argue again because of me. I am really afraid they will divorce, but I still cannot do anything well.}) \\

\rowcolor{gray!8}
Counselor & \zh{当你想到“父母可能会离婚”的时候，最强烈的感受通常是什么？是害怕、难过、内疚，还是别的什么？} (\textit{When you think ``my parents might divorce,'' what is usually the strongest feeling: fear, sadness, guilt, or something else?}) \\

Client & \zh{那种感觉……就像胸口被堵住了一样，闷得透不过气。我觉得这一切都是因为我没用，如果我能更出色一点，他们或许就不会那么辛苦、也不会总吵架了。} (\textit{It feels as if my chest is blocked, so tight that I cannot breathe. I feel all of this is because I am useless. If I were better, maybe they would not be so exhausted or argue so much.}) \\

\rowcolor{gray!8}
Counselor & \zh{你把这个想法说得很清楚，也很沉重。你不是不在乎父母，恰恰是太在乎了，所以才会把他们的痛苦和责任往自己身上放。} (\textit{You have stated that thought very clearly, and it is very heavy. It is not that you do not care about your parents. Rather, you care so much that you put their pain and responsibility onto yourself.}) \\

Client & \zh{可我除了学习，好像真的找不到别的价值了。如果连书都读不进去，我就真的是个废人，除了给他们添麻烦，什么也做不到。} (\textit{But apart from studying, I really cannot find any other value in myself. If I cannot even study, then I really am useless and can do nothing except cause them trouble.}) \\

\rowcolor{gray!8}
Counselor & \zh{你现在说到“废人”“什么都做不到”的时候，有没有出现过伤害自己、或者不想活了这类念头？} (\textit{When you say ``useless'' and ``can do nothing,'' have you had thoughts of hurting yourself or not wanting to live?}) \\

Client & \zh{有过的……有时候心里实在堵得太难受，或者觉得什么都搞砸了的时候，就会想去划一下手臂，只要感觉到身上疼，那种混乱和窒息感好像才能稍微停下来一点。} (\textit{Yes. Sometimes when my chest feels unbearably blocked, or when I feel I have ruined everything, I think about cutting my arm. When I feel physical pain, the confusion and suffocation seem to stop a little.}) \\

\rowcolor{gray!8}
Counselor & \zh{谢谢你告诉我这个很重要。我们先把安全放在前面：如果你现在有强烈冲动，先尽量不要一个人待着，可以把自己挪到相对安全的地方，比如客厅，或者给一个可信任的人发一句“我现在不太安全，能不能陪我一下”。这不是麻烦别人，而是在帮自己撑过情绪最危险的那一阵。} (\textit{Thank you for telling me; this is important. Let us put safety first. If you have a strong urge right now, try not to stay alone. Move to a safer place, such as the living room, or message someone you trust: ``I do not feel very safe right now. Could you stay with me?'' This is not troubling others; it is helping yourself get through the most dangerous wave of emotion.}) \\

Client & \zh{我知道了，谢谢老师。其实我平时大多时候都是一个人待着，不太习惯去麻烦别人，但我会记着您的建议的，尽量不让自己陷入那种极端的状态。} (\textit{I understand. Thank you. Most of the time I stay alone and am not used to troubling others, but I will remember your suggestion and try not to let myself fall into that extreme state.}) \\

\rowcolor{gray!8}
Counselor & \zh{你一方面很痛苦，另一方面又担心给别人添麻烦，所以很多时候只能自己扛；同时你也愿意在危险的时候先保护自己，这一点很重要。} (\textit{On one hand you are in pain, and on the other hand you worry about burdening others, so you often carry it alone. At the same time, you are willing to protect yourself first when things become dangerous, and that matters.}) \\

Client & \zh{嗯……其实我也知道这样不好，一直闷在心里只会越来越难受。有时候真想找个人说说话，可又总觉得别人听了我的烦恼也会跟着不开心，最后还是选择了沉默。} (\textit{Yes. I actually know this is not good. Keeping everything inside only makes it worse. Sometimes I really want to talk to someone, but I worry that hearing my troubles would make them unhappy too, so in the end I stay silent.}) \\

\rowcolor{gray!8}
Counselor & \zh{听起来你长期都在独自消化这些压力。} (\textit{It sounds like you have been digesting all this pressure alone for a long time.}) \\

Client & \zh{是啊，我总是怕表现得太负面会把别人吓跑，就像以前初中时那样……最后就真的只剩下自己一个人了，反倒觉得这样才最安全。} (\textit{Yes. I am always afraid that if I seem too negative, I will scare people away, like in junior high. In the end I really was left alone, so somehow that feels safest.}) \\

\rowcolor{gray!8}
Counselor & \zh{你能觉察到“沉默让我更安全，但也更孤单”这个矛盾，说明你并不是完全被困住了，你已经在看见自己承受的方式。} (\textit{Noticing the contradiction that silence makes you feel safer but also lonelier shows that you are not completely trapped. You are already seeing the way you have been coping.}) \\

Client & \zh{可是这种安全感，好像也只是让我一直待在原地，什么也改变不了。有时候看着别人能轻松地谈笑，我就觉得我和他们之间隔着一层雾，怎么也走不过去。} (\textit{But that sense of safety seems to keep me in the same place, changing nothing. Sometimes when I see others chatting and laughing easily, I feel there is a layer of fog between us that I cannot cross.}) \\

\rowcolor{gray!8}
Counselor & \zh{你像是站在外面看别人生活，能理解一些，却很难真正参与进去；这种距离感会让你更确认自己是不被接纳的。} (\textit{It is as if you are standing outside watching other people live. You can understand some of it, but it is hard to truly join in; that distance then makes you feel even more certain that you are not accepted.}) \\

Client & \zh{是的，好像无论我怎么努力想要靠近，最后的结果都是被推开，或者是我自己先退缩了。我总觉得如果让他们看到真实的我，看到我这些乱七八糟的念头，他们肯定也会觉得我很怪、很压抑，最后还是会选择离开。} (\textit{Yes. No matter how hard I try to get closer, the result seems to be that I am pushed away, or I withdraw first. I always feel that if they saw the real me and these messy thoughts, they would think I am strange and depressing, and eventually leave.}) \\

\rowcolor{gray!8}
Counselor & \zh{在这些关系里，最近有没有一个人让你感觉稍微没那么难接近？不一定是很亲密，哪怕只是一点点。} (\textit{In these relationships, has there recently been one person who feels slightly less hard to approach? They do not have to be very close; even a little counts.}) \\

Client & \zh{最近倒是有一个同学，偶尔会主动问我笔记或者作业的事。虽然我总是尽量用最简短的话回她，但她好像也不怎么介意，还会顺口问我吃饭了没。} (\textit{There is one classmate recently who sometimes asks me about notes or assignments. I always try to answer as briefly as possible, but she does not seem to mind and even casually asks whether I have eaten.}) \\

\rowcolor{gray!8}
Counselor & \zh{我们可以先不把目标定成“马上交到朋友”或“完全信任对方”，而是从这种很小的互动开始，试着多停留几秒。比如她问作业时，你可以除了回答内容，多加一句很短的近况：“最近复习有点卡，但还是会慢慢补。”看看对方的反应。} (\textit{We do not need to set the goal as immediately making a friend or fully trusting someone. We can start from this small interaction and try staying in it a few seconds longer. For example, when she asks about homework, besides answering, you could add a short update: ``Revision has been a bit stuck recently, but I am slowly catching up.'' Then see how she responds.}) \\

Client & \zh{我其实……还没尝试过这样回复别人。每次她问我的时候，我心里总会慌一下，生怕说多了会显得我很多话，或者让她觉得我真的很差劲。但我可以试着……下次如果她再问起，我努力多说这一句试试看。} (\textit{I actually have not tried replying like that. Every time she asks me, I panic a little, afraid that saying more will make me seem talkative or make her think I am really bad. But I can try. If she asks again next time, I will try to add that one sentence.}) \\

\rowcolor{gray!8}
Counselor & \zh{你愿意试的这部分很具体，而且不是强迫自己一下子敞开，只是在原来简短回应里加一点点真实信息，这比较适合你现在的状态。} (\textit{The part you are willing to try is very concrete, and it does not force you to open up all at once. It only adds a little real information to your usual brief reply, which fits your current state.}) \\

Client & \zh{我也觉得这样似乎没那么让人紧张。其实每次回应她那种简短的“嗯”、“噢”的时候，我也觉得心里空落落的，好像又把一道墙砌得更高了。} (\textit{I also feel this may be less nerve-racking. Actually, whenever I reply with short words like ``Mm'' or ``Oh,'' I feel empty inside, as if I have built the wall even higher.}) \\

\rowcolor{gray!8}
Counselor & \zh{你能感受到那堵墙的存在，也能想象“也许稍微低一点”，这已经是一个很实际的松动点。我们不急着拆掉整面墙，先留一条很小的缝就够了。} (\textit{You can feel that wall, and you can imagine that it might be a little lower. That is already a real point of loosening. We do not need to tear down the whole wall; leaving a small crack is enough for now.}) \\

Client & \zh{嗯，留一条缝……听起来好像没那么可怕。我会试着在下一次她找我的时候，稍微多说那么一点点，哪怕只是说出我当下的感受，应该也不会显得很奇怪吧。} (\textit{Yes, leaving a small crack sounds less frightening. I will try to say just a little more next time she talks to me. Even if I only say how I feel at the moment, maybe it will not seem that strange.}) \\

\rowcolor{gray!8}
Counselor & \zh{你没有要求自己立刻变得外向，也没有否认害怕，只是愿意在安全范围内多表达一点，这个方向很稳。} (\textit{You are not asking yourself to become outgoing immediately, and you are not denying your fear. You are simply willing to express a little more within a safe range. That is a steady direction.}) \\

Client & \zh{我也希望自己能慢慢好起来，不要一直被这些念头困住。谢谢老师，今天和您聊完，那种胸口闷闷的感觉好像确实稍微松了一些。} (\textit{I also hope I can slowly get better and not stay trapped by these thoughts. Thank you. After talking with you today, the tight feeling in my chest does seem a little looser.}) \\

\rowcolor{gray!8}
Counselor & \zh{我们今天先到这里。结束前请你记得：如果自伤冲动变强，优先联系身边可信任的人或当地紧急求助资源；如果可以，也把你刚才想到的那一句回复记录下来。下次我们再一起看看，哪些时刻最容易把你推向绝望，以及怎样更安全地度过它们。} (\textit{Let us stop here for today. Before we end, please remember: if the urge to self-harm becomes stronger, first contact someone trustworthy nearby or local emergency resources. If possible, also write down the sentence you just thought of. Next time we can look together at which moments most easily push you toward despair and how to get through them more safely.}) \\

Client & \zh{好的，老师。我会把这句话记录下来的，也会试着在难受的时候先提醒自己。谢谢您，那我们下次再聊。} (\textit{Okay. I will write that sentence down and try to remind myself first when I feel distressed. Thank you. We can talk again next time.}) \\

\rowcolor{gray!8}
Counselor & \zh{不客气。今天你已经做了很不容易的一步，先把自己稳住。我们下次继续慢慢梳理。} (\textit{You are welcome. Today you have already taken a very difficult step. Stabilize yourself first, and we will continue sorting things out next time.}) \\

\bottomrule
\caption{Emotion-management dialogue with a simulated client and CounselTrajLM counselor.}\label{tab:complete_dialogue_emotion_management}\\
\end{longtable}
\endgroup

\begingroup
\scriptsize
\setlength{\tabcolsep}{5pt}
\renewcommand{\arraystretch}{1.18}
\begin{longtable}{@{}p{0.11\textwidth}>{\raggedright\arraybackslash}p{0.84\textwidth}@{}}
\toprule
\textbf{Role} & \textbf{Utterance} \\
\midrule
\endfirsthead

\multicolumn{2}{c}{\textit{Continued from previous page}} \\
\toprule
\textbf{Role} & \textbf{Utterance} \\
\midrule
\endhead

\midrule
\endfoot

\endlastfoot

\rowcolor{gray!8}
Counselor & \zh{你好，我是今天和你文字咨询的咨询师。你可以按自己的节奏说说，最近主要是什么让你想来聊一聊？} (\textit{Hello, I am the counselor for today's text-based session. You can speak at your own pace. What has mainly brought you here recently?}) \\

Client & \zh{老师你好。其实我最近挺心烦的，马上就要毕业了，但一直没找到特别理想的工作，心里总觉得悬着，特别不安。} (\textit{Hello. I have been quite upset recently. I am about to graduate, but I still have not found an ideal job, so I feel suspended and very uneasy.}) \\

\rowcolor{gray!8}
Counselor & \zh{你现在一边面对毕业这个节点，一边在找工作上遇到卡住的情况，所以那种不确定感一直压在心里。} (\textit{You are facing graduation while also feeling stuck in the job search, so that uncertainty has been weighing on you.}) \\

Client & \zh{是啊，周围同学很多都已经签了不错的单位，但我投递的几个心仪的三甲医院一直没回音，感觉自己好像一下子落后了。} (\textit{Yes. Many classmates have already signed with good workplaces, but the tertiary hospitals I applied to have not replied. I feel as if I have suddenly fallen behind.}) \\

\rowcolor{gray!8}
Counselor & \zh{你能把这个压力说出来很不容易，尤其是它不只是“找工作”本身，还牵涉到你对未来、对选择的担心。} (\textit{It is not easy to say this pressure out loud, especially because it is not only about finding a job, but also about your worries about the future and your choices.}) \\

Client & \zh{对，我可能就是那种一直按部就班生活的人，突然这种“没着落”的状态让我很慌。以前无论是学习还是做班委，只要努力都会有回报，但现在好像无论我怎么投简历，结果都不由我控制。} (\textit{Yes. I may be someone who has always lived step by step, so this sudden state of having no clear landing makes me panic. In the past, whether in studying or class work, effort brought returns, but now no matter how many resumes I send, the result is not under my control.}) \\

\rowcolor{gray!8}
Counselor & \zh{这种焦虑最近通常会在什么时刻变得最明显？比如看到招聘信息、等面试结果，还是和同学比较的时候？} (\textit{When has this anxiety been most obvious recently? For example, when seeing job postings, waiting for interview results, or comparing yourself with classmates?}) \\

Client & \zh{主要是晚上吧，躺在床上的时候，或者看到群里同学分享入职体检消息的时候，那种落差感特别明显，心脏还会跳得很快。} (\textit{Mostly at night, when I am lying in bed, or when classmates share pre-employment health check messages in the group chat. The sense of falling behind becomes very obvious, and my heart beats fast.}) \\

\rowcolor{gray!8}
Counselor & \zh{也就是说，真正让你难受的不只是没有拿到offer，还有和同龄人进度不一样时产生的被甩下的感觉，身体上会有明显的紧张反应。} (\textit{So what really hurts is not only not receiving an offer, but also the feeling of being left behind when your pace differs from your peers, along with clear physical tension.}) \\

Client & \zh{嗯，大概就是这种感觉。我明明知道大家选择不同，但就是忍不住会想，是不是我哪里做得不够好，或者其实我并没有自己以为的那么优秀。} (\textit{Yes, that is probably it. I know everyone has different choices, but I cannot help wondering whether I did something poorly, or whether I am not as capable as I thought.}) \\

\rowcolor{gray!8}
Counselor & \zh{你说得很清楚，而且你已经觉察到自己在把“结果不顺”慢慢理解成“我不够好”，这个觉察很重要。} (\textit{You are saying this clearly, and you have noticed that you are gradually interpreting ``things are not going smoothly'' as ``I am not good enough.'' That awareness is important.}) \\

Client & \zh{可是如果真的不是我不够好，那为什么我投出的简历都石沉大海了呢？我真的会怀疑是不是我的能力，或者面试表现出了什么问题。} (\textit{But if it is really not that I am inadequate, why have all my resumes disappeared without response? I really suspect there may be something wrong with my ability or interview performance.}) \\

\rowcolor{gray!8}
Counselor & \zh{你目前投递的岗位，主要都是医院相关的吗？} (\textit{Are the positions you are applying for mainly hospital-related?}) \\

Client & \zh{对，都是我们专业对口的三甲医院。虽然我知道竞争很激烈，但眼看着身边的同学一个个都敲定了，我真的很难不去想，是不是我哪里出了差错。} (\textit{Yes, all are tertiary hospitals related to my major. Although I know the competition is intense, seeing classmates settle things one by one makes it hard not to think something is wrong with me.}) \\

\rowcolor{gray!8}
Counselor & \zh{我听到了，你不是随便投了几份就停，而是有明确方向，也在持续尝试。} (\textit{I hear that you did not casually submit a few applications and stop. You have a clear direction and are continuing to try.}) \\

Client & \zh{对，我投了好几家，但可能就是因为投入了心血，现在收不到反馈，心里才会更没底。我甚至在想，是不是我之前的实习表现不够突出，还是简历哪里写得不好？} (\textit{Yes, I applied to several places, and perhaps because I put effort into them, not receiving feedback makes me feel even less secure. I even wonder whether my previous internship performance was not outstanding enough or whether my resume is poorly written.}) \\

\rowcolor{gray!8}
Counselor & \zh{当你想到“可能是简历或面试出了问题”的时候，你通常会怎么应对这个想法？} (\textit{When you think ``maybe something is wrong with my resume or interview,'' how do you usually respond to that thought?}) \\

Client & \zh{我一开始会反复检查简历，改了又改，甚至想找老师再帮忙看一遍。但改完之后，那种不安感并没有减少，反而觉得好像怎么改都还是不够完美。} (\textit{At first I repeatedly check my resume and revise it again and again, even wanting to ask a teacher to review it. But after revising it, the uneasiness does not decrease; instead, it feels like no version is perfect enough.}) \\

\rowcolor{gray!8}
Counselor & \zh{你会通过检查和修改来缓解焦虑，可一旦没有立刻得到反馈，修改本身又变成了新的压力源。} (\textit{You try to ease anxiety by checking and revising, but once feedback does not come immediately, revision itself becomes a new source of pressure.}) \\

Client & \zh{是这样，我总觉得只要我做得再完美一点，哪怕只是简历上的措辞再精简一些，机会就会多一点。但我现在越改越觉得自己好像在做无用功，这种无法掌控结果的感觉真的太难受了。} (\textit{Yes. I keep feeling that if I make it a little more perfect, even just more concise wording on the resume, I will have more chances. But now the more I revise, the more it feels like wasted effort. This lack of control over the outcome is really painful.}) \\

\rowcolor{gray!8}
Counselor & \zh{在这种压力下你还在主动优化简历、尝试面试，说明你并没有放弃自己，也没有完全被焦虑带走。} (\textit{Under this pressure, you are still actively improving your resume and trying interviews. That shows you have not given up on yourself, nor have you been completely carried away by anxiety.}) \\

Client & \zh{谢谢老师这样说，听起来好像我还在努力，但其实我真的好累，总觉得现在的每一天都在消耗我的信心。} (\textit{Thank you for saying that. It sounds like I am still trying, but honestly I am really tired. Every day now feels like it is draining my confidence.}) \\

\rowcolor{gray!8}
Counselor & \zh{我们可以先不急着判断你到底“行不行”，先把这件事拆开一点看：一部分是现实求职中的不确定，另一部分是你对自己能力的怀疑。这样可能会更容易处理。} (\textit{We do not need to rush to judge whether you are ``capable'' or not. We can separate this a little: one part is the real uncertainty of job seeking, and the other is your doubt about your ability. That may make it easier to handle.}) \\

Client & \zh{我也想把它们分开看，但现在只要一看到手机弹出消息，或者是看到招聘网站，我的脑子里立刻就会跳出“我又没戏了”这种念头，很难控制得住。} (\textit{I also want to separate them, but now whenever a phone notification pops up or I see a recruitment website, the thought ``I have no chance again'' immediately jumps into my mind, and it is hard to control.}) \\

\rowcolor{gray!8}
Counselor & \zh{谢谢你把这个细节说出来，这能帮助我们更准确地看见焦虑是怎么启动的。} (\textit{Thank you for sharing this detail. It helps us see more accurately how the anxiety gets triggered.}) \\

Client & \zh{是啊，那种“未战先怯”的感觉特别糟糕。我也不想这样，可就是控制不住去想最坏的结果，感觉自己好像陷入了一个怎么都跳不出来的怪圈里。} (\textit{Yes, that feeling of being afraid before the battle is awful. I do not want to be like this, but I cannot stop imagining the worst outcome. It feels like I am trapped in a loop I cannot escape.}) \\

\rowcolor{gray!8}
Counselor & \zh{人在面对高不确定性时，大脑很容易自动去寻找危险信号，尤其当你很在意这个结果时，它会更快把“没有回应”解释成“失败”。这不是你故意悲观，而是焦虑状态下常见的一种保护反应，只是它现在有点过度了。} (\textit{When people face high uncertainty, the brain easily searches for danger signals automatically. Especially when you care a lot about the outcome, it more quickly interprets ``no response'' as ``failure.'' This is not deliberate pessimism; it is a common protective response under anxiety, although it has become somewhat excessive now.}) \\

Client & \zh{听到您这么说，我心里稍微好受了一点，原来这并不是我性格真的出了什么问题，只是大脑的一种反应吗？但我还是很担心，如果一直保持这种心态，会不会影响我接下来的面试发挥？} (\textit{Hearing you say this makes me feel a little better. So it is not that something is really wrong with my personality, but a reaction of the brain? Still, I worry that if I keep this mindset, it may affect my next interview performance.}) \\

\rowcolor{gray!8}
Counselor & \zh{你一方面松了一点，知道这不等于自己“有问题”；另一方面也担心这种焦虑会进一步削弱你的面试状态。} (\textit{On one hand you feel a bit relieved, knowing this does not mean there is something wrong with you. On the other hand, you worry this anxiety may further weaken your interview state.}) \\

Client & \zh{对，我就是特别担心这个。如果下次面试的时候，我因为太紧张导致表现失常，那不就真的应验了我最担心的“我不够好”了吗？这种恶性循环让我感觉很可怕。} (\textit{Yes, that is exactly what I worry about. If I become too nervous in the next interview and perform badly, would that not confirm my biggest fear that I am not good enough? This vicious cycle feels frightening.}) \\

\rowcolor{gray!8}
Counselor & \zh{你能把这个循环说出来很关键：担心失败——焦虑升高——害怕表现受影响——更担心失败。看见循环，才有可能在其中某一环轻轻打断它。} (\textit{It is important that you can name this cycle: fear of failure, rising anxiety, fear that performance will be affected, and then even more fear of failure. Seeing the cycle makes it possible to gently interrupt one part of it.}) \\

Client & \zh{那老师，我该怎么才能打断这个循环呢？每次站在面试室门口，我都感觉呼吸都变重了，根本没法静下心来。} (\textit{Then how can I interrupt this cycle? Every time I stand outside the interview room, my breathing feels heavy, and I cannot calm down at all.}) \\

\rowcolor{gray!8}
Counselor & \zh{如果我们先选一个最容易下手的地方，你觉得是面试前的准备、面试当下的紧张，还是等待结果时的反复想，哪一块最想先改善一点？} (\textit{If we first choose the easiest place to start, would you most want to improve preparation before the interview, nervousness during the interview, or repeated thoughts while waiting for results?}) \\

Client & \zh{我觉得最难受的还是等待结果的时候，那种时刻都在反复想的感觉，把我的精气神都磨没了。要是能把这个部分减弱一点，我觉得我在面试的时候可能也不会那么紧绷。} (\textit{I think the hardest part is waiting for results. The feeling of thinking about it constantly wears me down. If that part could be reduced a little, I might not be so tense during interviews either.}) \\

\rowcolor{gray!8}
Counselor & \zh{那我们先聚焦在等待结果这一段。你可以给自己设一个固定的“担心时间”，比如每天晚上用20分钟专门写下担心和能做的下一步，其他时间它冒出来时，就先记一句“晚上再处理”。这样不是压住焦虑，而是给它一个边界。} (\textit{Then let us focus on the waiting period first. You can set a fixed ``worry time,'' such as 20 minutes each evening to write down your worries and possible next steps. At other times, when the worry appears, note ``I will handle this tonight.'' This does not suppress anxiety; it gives it a boundary.}) \\

Client & \zh{这个方法听起来好像可行，确实我现在是随时随地都在担心，脑子里就像有个关不掉的广播。那如果到时候我写着写着，发现自己还是控制不住地想那些负面的结果，该怎么办呢？} (\textit{This sounds feasible. I really do worry anytime and anywhere now, like there is a radio in my mind that cannot be turned off. But what if, while writing, I still cannot stop thinking about negative outcomes?}) \\

\rowcolor{gray!8}
Counselor & \zh{当负面想法出现时，你愿不愿意先做一个很短的停顿，比如问自己：“这是事实，还是我在预测最坏结果？”不用马上说服自己乐观，只是先区分一下。} (\textit{When negative thoughts appear, would you be willing to pause briefly and ask yourself, ``Is this a fact, or am I predicting the worst outcome?'' You do not need to immediately persuade yourself to be optimistic; just distinguish them first.}) \\

Client & \zh{这听起来像是在帮我给大脑装个“红绿灯”，虽然不知道能不能一下子停住，但我愿意试试看。下次再冒出那些念头的时候，我就先问自己这一句，看看能不能把它和事实分开。} (\textit{That sounds like installing a traffic light in my brain. I do not know whether it can stop me immediately, but I am willing to try. Next time those thoughts come up, I will first ask myself that question and see whether I can separate them from facts.}) \\

\rowcolor{gray!8}
Counselor & \zh{很好。你可以把这句话写在备忘录里，尤其是晚上容易反复刷招聘软件的时候看一眼。目标不是让焦虑立刻消失，而是让你在焦虑中多一点点选择。} (\textit{Good. You can write this sentence in your notes, especially to look at when you tend to repeatedly check recruitment apps at night. The goal is not to make anxiety disappear immediately, but to give you a little more choice within it.}) \\

Client & \zh{我知道了，这听起来像是一个能让我喘口气的小练习。我会把它写在备忘录里的，今晚尝试一下这种方法，看看能不能稍微把那些念头和事实隔开一点。} (\textit{I understand. This sounds like a small exercise that can let me catch my breath. I will write it in my notes and try it tonight to see whether I can create a little distance between those thoughts and the facts.}) \\

\bottomrule
\caption{Career-development dialogue with a simulated client and CounselTrajLM counselor.}\label{tab:complete_dialogue_career_development}\\
\end{longtable}
\endgroup

%% file: custom.bib
@inproceedings{lee-etal-2024-cactus,
    title = "Cactus: Towards Psychological Counseling Conversations using Cognitive Behavioral Theory",
    author = "Lee, Suyeon  and
      Kim, Sunghwan  and
      Kim, Minju  and
      Kang, Dongjin  and
      Yang, Dongil  and
      Kim, Harim  and
      Kang, Minseok  and
      Jung, Dayi  and
      Kim, Min Hee  and
      Lee, Seungbeen  and
      Chung, Kyong-Mee  and
      Yu, Youngjae  and
      Lee, Dongha  and
      Yeo, Jinyoung",
    editor = "Al-Onaizan, Yaser  and
      Bansal, Mohit  and
      Chen, Yun-Nung",
    booktitle = "Findings of the Association for Computational Linguistics: EMNLP 2024",
    month = nov,
    year = "2024",
    address = "Miami, Florida, USA",
    publisher = "Association for Computational Linguistics",
    url = "https://aclanthology.org/2024.findings-emnlp.832",
    doi = "10.18653/v1/2024.findings-emnlp.832",
    pages = "14245--14274",
}

@inproceedings{qiu-etal-2024-smile,
    title = "{SMILE}: Single-turn to Multi-turn Inclusive Language Expansion via {C}hat{GPT} for Mental Health Support",
    author = "Qiu, Huachuan  and
      He, Hongliang  and
      Zhang, Shuai  and
      Li, Anqi  and
      Lan, Zhenzhong",
    editor = "Al-Onaizan, Yaser  and
      Bansal, Mohit  and
      Chen, Yun-Nung",
    booktitle = "Findings of the Association for Computational Linguistics: EMNLP 2024",
    month = nov,
    year = "2024",
    address = "Miami, Florida, USA",
    publisher = "Association for Computational Linguistics",
    url = "https://aclanthology.org/2024.findings-emnlp.34",
    doi = "10.18653/v1/2024.findings-emnlp.34",
    pages = "615--636",
}

@INPROCEEDINGS{AnnoMi,
  author={Wu, Zixiu and Balloccu, Simone and Kumar, Vivek and Helaoui, Rim and Reiter, Ehud and Reforgiato Recupero, Diego and Riboni, Daniele},
  booktitle={ICASSP 2022 - 2022 IEEE International Conference on Acoustics, Speech and Signal Processing (ICASSP)}, 
  title={{Anno-MI}: A Dataset of Expert-Annotated Counselling Dialogues}, 
  year={2022},
  volume={},
  number={},
  pages={6177--6181},
  doi={10.1109/ICASSP43922.2022.9746035}}

@inproceedings{qi-etal-2025-kokorochat,
    title = "{K}okoro{C}hat: A {J}apanese Psychological Counseling Dialogue Dataset Collected via Role-Playing by Trained Counselors",
    author = "Qi, Zhiyang  and
      Kaneko, Takumasa  and
      Takamizo, Keiko  and
      Ukiyo, Mariko  and
      Inaba, Michimasa",
    editor = "Che, Wanxiang  and
      Nabende, Joyce  and
      Shutova, Ekaterina  and
      Pilehvar, Mohammad Taher",
    booktitle = "Proceedings of the 63rd Annual Meeting of the Association for Computational Linguistics (Volume 1: Long Papers)",
    month = jul,
    year = "2025",
    address = "Vienna, Austria",
    publisher = "Association for Computational Linguistics",
    url = "https://aclanthology.org/2025.acl-long.608/",
    doi = "10.18653/v1/2025.acl-long.608",
    pages = "12424--12443",
    ISBN = "979-8-89176-251-0"
}

@inproceedings{zhang-etal-2024-cpsycoun,
    title = "{CP}sy{C}oun: A Report-based Multi-turn Dialogue Reconstruction and Evaluation Framework for {C}hinese Psychological Counseling",
    author = "Zhang, Chenhao  and
      Li, Renhao  and
      Tan, Minghuan  and
      Yang, Min  and
      Zhu, Jingwei  and
      Yang, Di  and
      Zhao, Jiahao  and
      Ye, Guancheng  and
      Li, Chengming  and
      Hu, Xiping",
    editor = "Ku, Lun-Wei  and
      Martins, Andre  and
      Srikumar, Vivek",
    booktitle = "Findings of the Association for Computational Linguistics: ACL 2024",
    month = aug,
    year = "2024",
    address = "Bangkok, Thailand",
    publisher = "Association for Computational Linguistics",
    url = "https://aclanthology.org/2024.findings-acl.830/",
    doi = "10.18653/v1/2024.findings-acl.830",
    pages = "13947--13966"
}

@inproceedings{xie-etal-2025-psydt,
    title = "{P}sy{DT}: Using {LLM}s to Construct the Digital Twin of Psychological Counselor with Personalized Counseling Style for Psychological Counseling",
    author = "Xie, Haojie  and
      Chen, Yirong  and
      Xing, Xiaofen  and
      Lin, Jingkai  and
      Xu, Xiangmin",
    editor = "Che, Wanxiang  and
      Nabende, Joyce  and
      Shutova, Ekaterina  and
      Pilehvar, Mohammad Taher",
    booktitle = "Proceedings of the 63rd Annual Meeting of the Association for Computational Linguistics (Volume 1: Long Papers)",
    month = jul,
    year = "2025",
    address = "Vienna, Austria",
    publisher = "Association for Computational Linguistics",
    url = "https://aclanthology.org/2025.acl-long.55/",
    doi = "10.18653/v1/2025.acl-long.55",
    pages = "1081--1115",
    ISBN = "979-8-89176-251-0"
}

@inproceedings{qiu-lan-2025-psydial,
    title = "{P}sy{D}ial: A Large-scale Long-term Conversational Dataset for Mental Health Support",
    author = "Qiu, Huachuan  and
      Lan, Zhenzhong",
    editor = "Che, Wanxiang  and
      Nabende, Joyce  and
      Shutova, Ekaterina  and
      Pilehvar, Mohammad Taher",
    booktitle = "Proceedings of the 63rd Annual Meeting of the Association for Computational Linguistics (Volume 1: Long Papers)",
    month = jul,
    year = "2025",
    address = "Vienna, Austria",
    publisher = "Association for Computational Linguistics",
    url = "https://aclanthology.org/2025.acl-long.1049/",
    doi = "10.18653/v1/2025.acl-long.1049",
    pages = "21624--21655",
    ISBN = "979-8-89176-251-0"
}

@inproceedings{chen-etal-2023-soulchat,
    title = "{S}oul{C}hat: Improving {LLM}s' Empathy, Listening, and Comfort Abilities through Fine-tuning with Multi-turn Empathy Conversations",
    author = "Chen, Yirong  and
      Xing, Xiaofen  and
      Lin, Jingkai  and
      Zheng, Huimin  and
      Wang, Zhenyu  and
      Liu, Qi  and
      Xu, Xiangmin",
    editor = "Bouamor, Houda  and
      Pino, Juan  and
      Bali, Kalika",
    booktitle = "Findings of the Association for Computational Linguistics: EMNLP 2023",
    month = dec,
    year = "2023",
    address = "Singapore",
    publisher = "Association for Computational Linguistics",
    url = "https://aclanthology.org/2023.findings-emnlp.83/",
    doi = "10.18653/v1/2023.findings-emnlp.83",
    pages = "1170--1183"
}

@article{rogers1957necessary,
  title   = {The Necessary and Sufficient Conditions of Therapeutic Personality Change},
  author  = {Rogers, Carl R.},
  journal = {Journal of Consulting Psychology},
  volume  = {21},
  number  = {2},
  pages   = {95--103},
  year    = {1957},
  publisher = {American Psychological Association}
}

@techreport{who2022worldmentalhealth,
  title        = {World Mental Health Report: Transforming Mental Health for All},
  author       = {WHO},
  institution  = {World Health Organization},
  year         = {2022},
  url          = {https://www.who.int/publications/i/item/9789240049338}
}

@article{hua2025large,
  title   = {A scoping review of large language models for generative tasks in mental health care},
  author  = {Hua, Yining and Na, Hongbin and Li, Zehan and Liu, Fenglin and Fang, Xiao and Clifton, David and Torous, John},
  journal = {npj Digital Medicine},
  year    = {2025},
  volume  = {8},
  number  = {1},
  pages   = {230},
  doi     = {10.1038/s41746-025-01611-4},
  url     = {https://doi.org/10.1038/s41746-025-01611-4}
}

@article{casu2024ai,
  title   = {{AI} Chatbots for Mental Health: A Scoping Review of Effectiveness, Feasibility, and Applications},
  author  = {Casu, Mirko and Triscari, Sergio and Battiato, Sebastiano and Guarnera, Luca and Caponnetto, Pasquale},
  journal = {Applied Sciences},
  volume  = {14},
  number  = {13},
  pages   = {5889},
  year    = {2024},
  month   = jul,
  doi     = {10.3390/app14135889},
  publisher = {MDPI}
}

@inproceedings{wu2026ecas,
  title     = {An {LLM}-based Simulation Framework for Embodied Conversational Agents in Psychological Counseling},
  author    = {Wu, Lixiu and Tang, Yuanrong and Pan, Qisen and Zhan, Xianyang and Han, Yuchen and Xiao, Lanxi and Wang, Tianhong and Zhong, Chen and Gong, Jiangtao},
  booktitle = {Proceedings of the AAAI Conference on Artificial Intelligence},
  volume    = {40},
  pages     = {29758--29766},
  year      = {2026},
  doi       = {10.1609/aaai.v40i35.40221}
}

@article{fu2024hmmcounselling,
  title   = {Using Hidden {Markov} Modelling to Reveal In-Session Stages in Text-Based Counselling},
  author  = {Fu, Ziru and Hsu, Yu Cheng and Chan, Christian S. and Liu, Joyce and Yip, Paul S. F.},
  journal = {npj Mental Health Research},
  volume  = {3},
  pages   = {56},
  year    = {2024},
  doi     = {10.1038/s44184-024-00103-9},
  url     = {https://www.nature.com/articles/s44184-024-00103-9}
}

@inproceedings{rashkin2019towards,
    title = "Towards Empathetic Open-domain Conversation Models: A New Benchmark and Dataset",
    author = "Rashkin, Hannah  and
      Smith, Eric Michael  and
      Li, Margaret  and
      Boureau, Y-Lan",
    editor = "Korhonen, Anna  and
      Traum, David  and
      M{\`a}rquez, Llu{\'i}s",
    booktitle = "Proceedings of the 57th Annual Meeting of the Association for Computational Linguistics",
    month = jul,
    year = "2019",
    address = "Florence, Italy",
    publisher = "Association for Computational Linguistics",
    url = "https://aclanthology.org/P19-1534/",
    doi = "10.18653/v1/P19-1534",
    pages = "5370--5381"
}

@inproceedings{sharma-etal-2020-computational,
    title = "A Computational Approach to Understanding Empathy Expressed in Text-Based Mental Health Support",
    author = "Sharma, Ashish  and
      Miner, Adam  and
      Atkins, David  and
      Althoff, Tim",
    editor = "Webber, Bonnie  and
      Cohn, Trevor  and
      He, Yulan  and
      Liu, Yang",
    booktitle = "Proceedings of the 2020 Conference on Empirical Methods in Natural Language Processing (EMNLP)",
    month = nov,
    year = "2020",
    address = "Online",
    publisher = "Association for Computational Linguistics",
    url = "https://aclanthology.org/2020.emnlp-main.425/",
    doi = "10.18653/v1/2020.emnlp-main.425",
    pages = "5263--5276"
}

@inproceedings{sharma2021facilitating,
author = {Sharma, Ashish and Lin, Inna W. and Miner, Adam S. and Atkins, David C. and Althoff, Tim},
title = {Towards Facilitating Empathic Conversations in Online Mental Health Support: A Reinforcement Learning Approach},
year = {2021},
isbn = {9781450383127},
publisher = {Association for Computing Machinery},
address = {New York, NY, USA},
url = {https://doi.org/10.1145/3442381.3450097},
doi = {10.1145/3442381.3450097},
booktitle = {Proceedings of the Web Conference 2021},
pages = {194--205},
numpages = {12},
location = {Ljubljana, Slovenia},
series = {WWW '21}
}

@article{sharma2023human,
  title   = {Human--{AI} Collaboration Enables More Empathic Conversations in Text-Based Peer-to-Peer Mental Health Support},
  author  = {Sharma, Ashish and Lin, Inna W. and Miner, Adam S. and Atkins, David C. and Althoff, Tim},
  journal = {Nature Machine Intelligence},
  volume  = {5},
  pages   = {46--57},
  year    = {2023},
  doi     = {10.1038/s42256-022-00593-2}
}

@inproceedings{liu2021towards,
    title = "Towards Emotional Support Dialog Systems",
    author = "Liu, Siyang  and
      Zheng, Chujie  and
      Demasi, Orianna  and
      Sabour, Sahand  and
      Li, Yu  and
      Yu, Zhou  and
      Jiang, Yong  and
      Huang, Minlie",
    editor = "Zong, Chengqing  and
      Xia, Fei  and
      Li, Wenjie  and
      Navigli, Roberto",
    booktitle = "Proceedings of the 59th Annual Meeting of the Association for Computational Linguistics and the 11th International Joint Conference on Natural Language Processing (Volume 1: Long Papers)",
    month = aug,
    year = "2021",
    address = "Online",
    publisher = "Association for Computational Linguistics",
    url = "https://aclanthology.org/2021.acl-long.269/",
    doi = "10.18653/v1/2021.acl-long.269",
    pages = "3469--3483"
}

@article{song2025typing,
author = {Song, Inhwa and Pendse, Sachin R and Kumar, Neha and De Choudhury, Munmun},
title = {The Typing Cure: Experiences with Large Language Model Chatbots for Mental Health Support},
year = {2025},
issue_date = {November 2025},
publisher = {Association for Computing Machinery},
address = {New York, NY, USA},
volume = {9},
number = {7},
url = {https://doi.org/10.1145/3757430},
doi = {10.1145/3757430},
journal = {Proc. ACM Hum.-Comput. Interact.},
month = oct,
articleno = {CSCW249},
eid = {CSCW249},
numpages = {29}
}

@article{wester2024chatbot,
author = {Wester, Joel and Pohl, Henning and Hosio, Simo and van Berkel, Niels},
title = {"This Chatbot Would Never...": Perceived Moral Agency of Mental Health Chatbots},
year = {2024},
issue_date = {April 2024},
publisher = {Association for Computing Machinery},
address = {New York, NY, USA},
volume = {8},
number = {CSCW1},
url = {https://doi.org/10.1145/3637410},
doi = {10.1145/3637410},
journal = {Proc. ACM Hum.-Comput. Interact.},
month = apr,
articleno = {133},
eid = {133},
numpages = {28}
}

@inproceedings{poulsen2025codesigning,
author = {Poulsen, Adam and Hickie, Ian B and de Haan, Zsofi and Capon, William and LaMonica, Haley M and Eyeson-Annan, Ebenezer and Radwan, Jalal and Barnabie, Nia and Scott, Elizabeth M and Iorfino, Frank},
title = {Co-designing a conversational generative artificial intelligence system for youth mental health},
year = {2025},
isbn = {9798400713958},
publisher = {Association for Computing Machinery},
address = {New York, NY, USA},
url = {https://doi.org/10.1145/3706599.3719696},
doi = {10.1145/3706599.3719696},
booktitle = {Proceedings of the Extended Abstracts of the CHI Conference on Human Factors in Computing Systems},
articleno = {138},
note = {Article 138, 7 pages},
numpages = {7},
location = {
},
series = {CHI EA '25}
}

@article{althoff2016large,
    title = "Large-scale Analysis of Counseling Conversations: An Application of Natural Language Processing to Mental Health",
    author = "Althoff, Tim  and
      Clark, Kevin  and
      Leskovec, Jure",
    editor = "Lee, Lillian  and
      Johnson, Mark  and
      Toutanova, Kristina",
    journal = "Transactions of the Association for Computational Linguistics",
    volume = "4",
    year = "2016",
    address = "Cambridge, MA",
    publisher = "MIT Press",
    url = "https://aclanthology.org/Q16-1033/",
    doi = "10.1162/tacl_a_00111",
    pages = "463--476"
}

@book{hill2009helping,
  title={Helping Skills: Facilitating Exploration, Insight, and Action},
  author={Hill, Clara E.},
  edition={3rd},
  year={2009},
  publisher={American Psychological Association}
}

@article{cohen1960coefficient,
  title   = {A Coefficient of Agreement for Nominal Scales},
  author  = {Cohen, Jacob},
  journal = {Educational and Psychological Measurement},
  volume  = {20},
  number  = {1},
  pages   = {37--46},
  year    = {1960},
  doi     = {10.1177/001316446002000104}
}

@inproceedings{qlora,
  title     = {{QLoRA}: Efficient Finetuning of Quantized {LLMs}},
  author    = {Dettmers, Tim and Pagnoni, Artidoro and Holtzman, Ari and Zettlemoyer, Luke},
  booktitle = {Advances in Neural Information Processing Systems},
  volume    = {36},
  pages     = {10088--10115},
  year      = {2023},
  publisher = {Curran Associates, Inc.},
  doi       = {10.52202/075280-0441}
}

@inproceedings{qiu-etal-2025-emoagent,
    title = "{E}mo{A}gent: Assessing and Safeguarding Human-{AI} Interaction for Mental Health Safety",
    author = "Qiu, Jiahao  and
      He, Yinghui  and
      Juan, Xinzhe  and
      Wang, Yimin  and
      Liu, Yuhan  and
      Yao, Zixin  and
      Wu, Yue  and
      Jiang, Xun  and
      Yang, Ling  and
      Wang, Mengdi",
    editor = "Christodoulopoulos, Christos  and
      Chakraborty, Tanmoy  and
      Rose, Carolyn  and
      Peng, Violet",
    booktitle = "Proceedings of the 2025 Conference on Empirical Methods in Natural Language Processing",
    month = nov,
    year = "2025",
    address = "Suzhou, China",
    publisher = "Association for Computational Linguistics",
    url = "https://aclanthology.org/2025.emnlp-main.594/",
    doi = "10.18653/v1/2025.emnlp-main.594",
    pages = "11741--11756",
    ISBN = "979-8-89176-332-6"
}

@inproceedings{qiu-etal-2025-deepwell,
    title = "{D}eep{W}ell-Adol: A Scalable Expert-Based Dialogue Corpus for Adolescent Positive Mental Health and Wellbeing Promotion",
    author = "Qiu, Wenyu  and
      Wang, Yuxiong  and
      Tan, Jiajun  and
      Hou, Hanchao  and
      Liu, Qinda  and
      Yao, Wei  and
      Ni, Shiguang",
    editor = "Christodoulopoulos, Christos  and
      Chakraborty, Tanmoy  and
      Rose, Carolyn  and
      Peng, Violet",
    booktitle = "Proceedings of the 2025 Conference on Empirical Methods in Natural Language Processing",
    month = nov,
    year = "2025",
    address = "Suzhou, China",
    publisher = "Association for Computational Linguistics",
    url = "https://aclanthology.org/2025.emnlp-main.646/",
    doi = "10.18653/v1/2025.emnlp-main.646",
    pages = "12786--12810",
    ISBN = "979-8-89176-332-6"
}

@inproceedings{kim-etal-2025-share-story,
    title = "Can You Share Your Story? Modeling Clients' Metacognition and Openness for {LLM} Therapist Evaluation",
    author = "Kim, Minju  and
      Yoo, Dongje  and
      Hwang, Yeonjun  and
      Kang, Minseok  and
      Kim, Namyoung  and
      Gwak, Minju  and
      Kwak, Beong-woo  and
      Chae, Hyungjoo  and
      Kim, Harim  and
      Lee, Yunjoong  and
      Kim, Min Hee  and
      Jung, Dayi  and
      Chung, Kyong-Mee  and
      Yeo, Jinyoung",
    editor = "Che, Wanxiang  and
      Nabende, Joyce  and
      Shutova, Ekaterina  and
      Pilehvar, Mohammad Taher",
    booktitle = "Findings of the Association for Computational Linguistics: ACL 2025",
    month = jul,
    year = "2025",
    address = "Vienna, Austria",
    publisher = "Association for Computational Linguistics",
    url = "https://aclanthology.org/2025.findings-acl.1332/",
    doi = "10.18653/v1/2025.findings-acl.1332",
    pages = "25943--25962",
    ISBN = "979-8-89176-256-5"
}

@inproceedings{hu2026agentmental,
  title     = {{AgentMental}: An Interactive Multi-Agent Framework for Explainable and Adaptive Mental Health Assessment},
  author    = {Hu, Jinpeng and Wang, Ao and Xie, Qianqian and Li, Zhuo and Ma, Hui and Guo, Dan},
  booktitle = {Proceedings of the AAAI Conference on Artificial Intelligence},
  pages     = {31050--31058},
  year      = {2026},
  doi       = {10.1609/aaai.v40i37.40365}
}

@inproceedings{qiu-etal-2026-client,
    title = "Client Simulation via Conversational Trajectory Modeling for Both Trainee Practice and Model Evaluation in Mental Health Counseling",
    author = "Qiu, Huachuan  and
      Chen, Zhaoming  and
      Chen, Yuqian  and
      Xie, Yuan  and
      Lu, Yu  and
      Lan, Zhenzhong",
    editor = "Lee, Hung-Yi  and
      Dev, Sunipa  and
      Li, Junyi Jessy  and
      Elliott, Desmond",
    booktitle = "Proceedings of the 2026 Conference on Empirical Methods in Natural Language Processing",
    month = oct,
    year = "2026",
    address = "Budapest, Hungary",
    publisher = "Association for Computational Linguistics",
    url = "https://aclanthology.org/2026.emnlp-main.45/",
    doi = "10.18653/2026.emnlp-main.45",
    pages = "975--1016",
    ISBN = "979-8-89176-459-0"
}

@inproceedings{qiu2024psychat,
  title     = {{PsyChat}: A Client-Centric Dialogue System for Mental Health Support},
  author    = {Huachuan Qiu and Anqi Li and Lizhi Ma and Zhenzhong Lan},
  booktitle = {2024 27th International Conference on Computer Supported Cooperative Work in Design (CSCWD)},
  pages     = {2979--2984},
  year      = {2024},
  publisher = {IEEE},
  doi       = {10.1109/CSCWD61410.2024.10580641}
}

@inproceedings{crisp,
    title = "Crisp: Cognitive Restructuring of Negative Thoughts through Multi-turn Supportive Dialogues",
    author = "Zhou, Jinfeng  and
      Chen, Yuxuan  and
      Yin, Jianing  and
      Huang, Yongkang  and
      Shi, Yihan  and
      Zhang, Xikun  and
      Peng, Libiao  and
      Zhang, Rongsheng  and
      Lv, Tangjie  and
      Hu, Zhipeng  and
      Wang, Hongning  and
      Huang, Minlie",
    editor = "Christodoulopoulos, Christos  and
      Chakraborty, Tanmoy  and
      Rose, Carolyn  and
      Peng, Violet",
    booktitle = "Proceedings of the 2025 Conference on Empirical Methods in Natural Language Processing",
    month = nov,
    year = "2025",
    address = "Suzhou, China",
    publisher = "Association for Computational Linguistics",
    url = "https://aclanthology.org/2025.emnlp-main.1652/",
    doi = "10.18653/v1/2025.emnlp-main.1652",
    pages = "32474--32503",
    ISBN = "979-8-89176-332-6"
}

@inproceedings{pan2026psycheval,
    title = "{P}sych{E}val: A Multi-Session and Multi-Therapy Benchmark for High-Realism {AI} Psychological Counselor",
    author = "Pan, Qianjun  and
      Wang, Junyi  and
      Zhou, Jie  and
      Yang, Yutao  and
      Li, Junsong  and
      Xu, Kaiyin  and
      Zhou, Yougen  and
      Li, Yihan  and
      Zhao, JingYuan  and
      Chen, Qin  and
      Zhou, Ningning  and
      Chen, Kai  and
      He, Liang",
    editor = "Liakata, Maria  and
      Moreira, Viviane P.  and
      Zhang, Jiajun  and
      Jurgens, David",
    booktitle = "Findings of the {A}ssociation for {C}omputational {L}inguistics: {ACL} 2026",
    month = jul,
    year = "2026",
    address = "San Diego, California, United States",
    publisher = "Association for Computational Linguistics",
    url = "https://aclanthology.org/2026.findings-acl.1115/",
    doi = "10.18653/v1/2026.findings-acl.1115",
    pages = "22192--22212",
    ISBN = "979-8-89176-395-1"
}

@article{horvath1989development,
  title={Development and Validation of the Working Alliance Inventory},
  author={Horvath, Adam O. and Greenberg, Leslie S.},
  journal={Journal of Counseling Psychology},
  volume={36},
  number={2},
  pages={223--233},
  year={1989},
  doi={10.1037/0022-0167.36.2.223}
}

@inproceedings{dai-etal-2026-psyche,
    title = "{Psyche-R1}: Towards Reliable Psychological {LLM}s through Unified Empathy, Expertise, and Reasoning",
    author = "Dai, Chongyuan  and
      Hu, Jinpeng  and
      Shi, Hongchang  and
      Li, Zhuo  and
      Guo, Dan  and
      Yang, Xun  and
      Wang, Meng",
    editor = "Liakata, Maria  and
      Moreira, Viviane P.  and
      Zhang, Jiajun  and
      Jurgens, David",
    booktitle = "Proceedings of the 64th Annual Meeting of the {A}ssociation for {C}omputational {L}inguistics (Volume 1: Long Papers)",
    month = jul,
    year = "2026",
    address = "San Diego, California, United States",
    publisher = "Association for Computational Linguistics",
    url = "https://aclanthology.org/2026.acl-long.1141/",
    doi = "10.18653/v1/2026.acl-long.1141",
    pages = "24889--24906",
    ISBN = "979-8-89176-390-6"
}

@inproceedings{qi-2026-less,
    title = "When Less Is More: An Empirical Study of Minimal Responses in Counseling Dialogues and the Behavior of {LLM}s",
    author = "Qi, Zhiyang",
    editor = "Lee, Hung-Yi  and
      Dev, Sunipa  and
      Li, Junyi Jessy  and
      Elliott, Desmond",
    booktitle = "Proceedings of the 2026 Conference on Empirical Methods in Natural Language Processing",
    month = oct,
    year = "2026",
    address = "Budapest, Hungary",
    publisher = "Association for Computational Linguistics",
    url = "https://aclanthology.org/2026.emnlp-main.527/",
    doi = "10.18653/2026.emnlp-main.527",
    pages = "12159--12173",
    ISBN = "979-8-89176-459-0"
}

@inproceedings{suzuki-etal-2026-multilingual,
    title = "Multilingual {K}okoro{C}hat: A Multi-{LLM} Ensemble Translation Method for Creating a Multilingual Counseling Dialogue Dataset",
    author = "Suzuki, Ryoma  and
      Qi, Zhiyang  and
      Inaba, Michimasa",
    editor = "Piperidis, Stelios  and
      Bel, N{\'u}ria  and
      van den Heuvel, Henk  and
      Ide, Nancy  and
      Krek, Simon  and
      Toral, Antonio",
    booktitle = "Proceedings of the Fifteenth Language Resources and Evaluation Conference",
    month = may,
    year = "2026",
    address = "Palma de Mallorca, Spain",
    publisher = "ELRA Language Resource Association",
    url = "https://aclanthology.org/2026.lrec-1.695/",
    doi = "10.63317/2d2qyq4zbps7",
    pages = "8838--8849"
}

@inproceedings{inaba-2026-beyond,
    title = "Beyond Reflection: Affirmation as a Promising Behavioral Marker Associated with Quality in Text-Based Counseling",
    author = "Inaba, Michimasa",
    editor = "Lee, Hung-Yi  and
      Dev, Sunipa  and
      Li, Jessy  and
      Elliott, Desmond",
    booktitle = "Findings of the {A}ssociation for {C}omputational {L}inguistics: {EMNLP} 2026",
    month = oct,
    year = "2026",
    address = "Budapest, Hungary",
    publisher = "Association for Computational Linguistics",
    url = "https://aclanthology.org/2026.findings-emnlp.782/",
    doi = "10.18653/2026.findings-emnlp.782",
    pages = "16557--16571",
    ISBN = "979-8-89176-462-0"
}

@misc{qi2025emostageframeworkaccurateempathetic,
      title={{EmoStage}: A Framework for Accurate Empathetic Response Generation via Perspective-Taking and Phase Recognition}, 
      author={Zhiyang Qi and Keiko Takamizo and Mariko Ukiyo and Michimasa Inaba},
      year={2025},
      eprint={2506.19279},
      archivePrefix={arXiv},
      primaryClass={cs.CL},
      url={https://arxiv.org/abs/2506.19279}, 
}

@article{hu2026patternrecognizers,
  title = {From Pattern Recognizers to Personalized Companions: A Survey of Large Language Models in Mental Health},
  author = {Hu, He and Zhou, Yucheng and Wang, Qianning and Zou, Yingjian and Ma, Chiyuan and Si, Juzheng and Liu, Jianzhuang and Yu, Zitong and Cui, Laizhong and Ma, Fei and Tian, Qi},
  journal = {IEEE Transactions on Affective Computing},
  year = {2026},
  volume = {17},
  number = {3},
  pages = {2525--2544},
  doi = {10.1109/TAFFC.2026.3689490},
  url = {https://ieeexplore.ieee.org/document/11503093/}
}
